\documentclass[twoside,11pt]{article}

\usepackage{jmlr2e}

\usepackage{algorithm}
\usepackage{algorithmic}
\usepackage{amsmath}

\usepackage{amsthm}
\usepackage{amssymb}
\usepackage{dsfont}
\usepackage{stmaryrd}
\usepackage{booktabs}
\usepackage{url}
\usepackage{graphicx}
\usepackage{xcolor}

\renewcommand{\Pr}{P}

\renewcommand{\vec}[1]{\boldsymbol{#1}}
\newcommand{\bx}{\vec{x}}
\newcommand{\by}{\vec{y}}
\newcommand{\bh}{\vec{h}}
\newcommand{\brY}{\vec{Y}}
\newcommand{\rY}{Y}
\newcommand{\given}{\, | \,}

\newcommand{\mx}[1]{\mathbf{\mathrm{#1}}}

\newcommand{\assert}[1]{\llbracket #1 \rrbracket}

\DeclareMathOperator*{\argmax}{\arg \max}
\DeclareMathOperator*{\argmin}{\arg \min}

\jmlrheading{1}{2000}{1-48}{4/00}{10/00}{Waegeman et al.}

\ShortHeadings{On the Bayes-optimality of F-measure Maximizers}{Waegeman et al.}
\firstpageno{1}

\begin{document}

\title{On the Bayes-optimality of F-measure Maximizers}


\author{\name Willem Waegeman \email willem.waegeman@ugent.be \\
\addr Department of Mathematical Modelling, Statistics and Bioinformatics \\ 
Ghent University, Ghent 9000 Belgium
\AND
\name Krzysztof Dembczy{\'n}ski \email krzysztof.dembczynski@cs.put.poznan.pl \\
\name Arkadiusz Jachnik \email arkadiusz.jachnik@cs.put.poznan.pl \\
\addr Institute of Computing Science \\
Poznan University of Technology, Poznan, 60-695 Poland
\AND
\name Weiwei Cheng \email roywwcheng@gmail.com \\
\addr Amazon Inc., Berlin 10707 Germany
\AND
\name Eyke H{\"u}llermeier \email eyke@informatik.uni-marburg.de  \\
\addr Department of Mathematics and Computer Science\\
Marburg University, Marburg, 35032 Germany \\}


\editor{}

\maketitle

\begin{abstract}%
The F-measure, which has originally been introduced in information retrieval, is nowadays routinely used as a performance metric for problems such as binary classification, multi-label classification, and structured output prediction. Optimizing this measure is a statistically and computationally challenging problem, since no closed-form solution exists. Adopting a decision-theoretic perspective, this article provides a formal and experimental analysis of different approaches for maximizing the F-measure. We start with a Bayes-risk analysis of related loss functions, such as Hamming loss and subset zero-one loss, showing that optimizing such losses as a surrogate of the F-measure leads to a high worst-case regret. Subsequently, we perform a similar type of analysis for F-measure maximizing algorithms, showing that such algorithms are approximate, while relying on additional assumptions regarding the statistical distribution of the binary response variables. Furthermore, we present a new algorithm which is not only computationally efficient but also Bayes-optimal, regardless of the underlying distribution. To this end, the algorithm requires only a quadratic (with respect to the number of binary responses) number of parameters of the joint distribution. We illustrate the practical performance of all analyzed methods by means of experiments with multi-label classification problems.   


\end{abstract}

\begin{keywords}
  F-measure, Bayes-optimal predictions, regret, statistical decision theory, multi-label classification, structured output prediction
\end{keywords}

\section{Introduction}

Being rooted in information retrieval \citep{rijsbergen74}, the so-called F-measure is nowadays routinely used as a performance metric for different types of prediction problems, including binary classification, multi-label classification (MLC), and certain applications of structured output prediction. Amongst others, examples of such applications include chunking or named entity recognition in natural language processing \citep{Sang2003}, image segmentation or edge detection in computer vision \citep{Martin2004} and detection of geographic coincidence in social networks \citep{Zhuang2012}.


Compared to measures like error rate in binary classification and Hamming loss in multi-label classification, the F-measure enforces a better balance between performance on the minority and the majority class, respectively. Therefore, it is more suitable in the case of imbalanced data, as it does not take the true negative rate into account. Given a prediction  $\vec{h} = (h_1, \ldots, h_m) \in \{0,1\}^m$ of an $m$-dimensional binary label vector $\vec{y} = (y_1, \ldots, y_m) $ (e.g., the class labels of a test set of size $m$ in binary classification or the label vector associated with a single instance in MLC or the binary vector indicating named entities in a text document in a structured output prediction task), the F-measure is defined as follows:
\begin{equation}
F(\vec{y},\vec{h}) = \frac{2 \sum_{i=1}^m y_i h_i}{\sum_{i=1}^m y_i + \sum_{i=1}^m h_i} \in [0,1] \enspace ,
\label{eq:f1}
\end{equation}
where $0/0 = 1$ by definition. 
This measure essentially corresponds to the harmonic mean of precision $prec$ and recall $recl$:
$$
prec(\vec{y},\vec{h}) = \frac{\sum_{i=1}^m y_i h_i}{\sum_{i=1}^m h_i} , \quad
recl(\vec{y},\vec{h}) = \frac{\sum_{i=1}^m y_i h_i}{\sum_{i=1}^m y_i} \enspace .
$$
One can generalize the F-measure to a weighted harmonic average of these two values, but for the sake of simplicity, we stick to the unweighted mean, which is often referred to as the F1-score or the F1-measure.

Despite its popularity in experimental settings, very few theoretical studies of the F-measure can be found. This paper intends to bridge this gap by analyzing existing methods and, moreover, by presenting a new algorithm that exhibits the desirable property of statistical consistency. To this end, we will adopt a decision-theoretic viewpoint. Modeling the ground-truth as a random variable $\brY = (\rY_1, \rY_2, \ldots, \rY_m)$, i.e., assuming an underlying probability distribution $\Pr$ over $\{0,1\}^m$, the prediction $\vec{h}$ that maximizes the expected F-measure is given by 
\begin{equation}
\vec{h}_F = \argmax_{\vec{h} \in \lbrace 0,1 \rbrace^m} \mathbb{E} \left[F(\brY,\vec{h})\right]
          = \argmax_{\vec{h} \in \lbrace 0,1 \rbrace^m} \sum_{\vec{y} \in \lbrace 0,1 \rbrace^m}\Pr(\vec{y})\,F(\vec{y},\vec{h})  .
\label{eq:argmax_f1}
\end{equation}
The corresponding optimization problem is non-trivial and cannot be solved in closed form. Moreover, a brute-force search is infeasible, as it would require checking all $2^m$ combinations of prediction vector $\vec{h}$ and summing over an exponential number of terms in each combination. As a result, many researchers who report the F-measure in experimental studies rely on optimizing a surrogate loss as an approximation of (\ref{eq:argmax_f1}). For problems such as multi-label classification and structured output prediction, the Hamming loss and the subset zero-one loss are immediate candidates for such surrogates. However, as will be shown in Section~3, these surrogates do not yield a statistically consistent model and, more importantly, they manifest a high regret. As an intermezzo, we present results for the Jaccard index, which has recently gained an increased popularity in areas such as multi-label classification. This measure is closely related to the F-measure, and its optimization appears to be even more difficult.    

Apart from optimizing surrogates, a few more specialized approaches for finding the F-measure maximizer (\ref{eq:argmax_f1}) have been presented in the last decades \citep{lewis95, Chai2005, jansche07,Ye2012,Quevedo2012}. These algorithms will be revisited in Section~4. They typically require the assumption of independence of the $Y_i$, i.e., 
\begin{equation}
\label{eq:label_independence}
\Pr(\brY = \vec{y}) = \prod_{i=1}^m p_i^{y_i}(1 - p_i)^{1-y_i} \,, 
\end{equation}
with $p_i = \Pr(Y_i = 1)$.
While being natural for problems like binary classification, this assumption is indeed not tenable in domains like MLC and structured output prediction. We will show in Section~4 that algorithms based on independence assumptions or marginal probabilities are not statistically consistent when arbitrary probability distributions $\Pr$ are considered. Moreover, we also show that the worst-case regret of these algorithms is very high. 

Looking at (\ref{eq:argmax_f1}), it seems that information about the entire joint distribution $\Pr$ is needed to maximize the F-measure. Yet, as will be shown in this paper, the problem can be solved more efficiently. In Section~5, we present a general algorithm that requires only a quadratic instead of an exponential (with respect to $m$) number of parameters of the joint distribution. If these parameters are given, then, depending on their form, the exact solution can be obtained in 
quadratic or cubic time.
This result holds regardless of the underlying distribution. In particular, unlike algorithms such as \citep{Chai2005,jansche07,Ye2012} and \citep{Quevedo2012}, we do not require independence of the binary response variables (labels).  

Our theoretical results are specifically relevant for applications in multi-label classification and structured output prediction. In these application domains, three different aggregation schemes of the F-measure can be distinguished, namely instance-wise, micro- and macro-averaging. One should carefully distinguish these versions, since algorithms optimized with a given objective are usually performing suboptimally for other (target) evaluation measures---see e.g.\ \citep{Luaces2012}. In Section~\ref{sec:application_to_mlc}, we present extensive experimental results to illustrate the practical usefulness of our findings. More specifically, all examined methods are compared for a series of multi-label classification problems. One particular dataset originates from a recent data mining competition, in which we obtained the second place using some of the algorithms presented in this article. Let us anticipate that our experimental results will not determine a clear winner. This is not at all surprising: while enjoying the advantage of consistency, our algorithm requires the estimation of more parameters than existing approximate algorithms. As a consequence, exact optimization is not necessarily superior to approximate optimization. Instead, the relative performance of exact and approximate optimization depends on several factors, such as the sample size, the length of $\brY$, the shape of the distribution $\Pr$, etc.   

As mentioned above, we adopt a decision-theoretic point of view: assuming a probabilistic model to be given, the problem of F-measure maximization is interpreted as an inference problem. Before going into technical details, we like to stress that this is only one way of looking at F-measure maximization. A second, somewhat orthogonal approach is to optimize the F-measure during the training phase. This is sometimes referred to as empirical utility maximization. In general, optimality in this framework is different from our definition of optimality, but connections between the two paradigms have  recently been discussed by \citet{Ye2012}. These authors establish asymptotic equivalence results under the assumption of independence and infinitely large vectors $\brY$. They focus on binary classification problems, for which such assumptions are realistic, because the vector $\brY$ then represents an entire test set of i.i.d.\ observations. The same assumptions are made in another recent work that provides an interesting theoretical analysis for binary classification problems \citep{Zhao2013}. However, in structured output prediction and multi-label classification, independence does not hold and the length of $\brY$ might be small, especially if the instance-wise F-measure needs to be optimized. 

Algorithms that optimize the F-measure during training will hence not be discussed further in this article. Nevertheless, we briefly mention some of them here for the sake of completeness. In binary classification, such algorithms are extensions of support vector machines~\citep{Musicant_et_al_2003,Joachims_2005}, logistic regression~\citep{Jansche_2005} or boosting~\citep{Kokkinos2010}. However, the most popular methods, including \citep{Keerthi_et_al_2007}, rely on explicit threshold adjustment. A few specific algorithms have also been proposed for certain applications in structured output  prediction~\citep{Tsochantaridis_et_al_2005,Suzuki_et_al_2006,Daume_et_al_2009} and multi-label classification \citep{fan07,zhang10,Petterson_Caetano_2010,Petterson_Caetano_2011}. During training, some of these methods, especially those based on structured SVMs, need to solve an inference problem that is closely related but not identical to (\ref{eq:argmax_f1}). In a recent paper, we have presented a theoretical and experimental comparison of approaches that optimize the F-measure during training or inference, respectively, in the context of multi-label classification \citep{Dembczynski_et_al_2013}.  Since we focus on the decision-theoretic point of view in this work, we do not discuss the theoretical results obtained in that article, but for completeness we report some experimental results. We also do not discuss existing extensions of F-measure maximization for more specific learning settings, such as active learning \citep{Sawade2010}.

Parts of this article have already been published in previous conference papers \citep{Dembczynski_et_al_2011,Dembczynski_et_al_2013}. 
Here, we summarize the results of these papers in a unifying framework, provide a much more detailed theoretical analysis, and complement them by additional formal results as well as novel experimental studies. 
The rest of the article is organized as follows. Section~2 gives a formal definition of the notion of regret, which will serve as a key element for showing that most existing algorithms are suboptimal. Section~3 contains a regret analysis of algorithms that optimize other loss functions than the F-measure. In Section~4, we perform a similar analysis for F-measure inference algorithms that have been proposed in the literature. Subsequently, our own algorithm is presented in Section~5. In Section~\ref{sec:simulations}, we test the inference algorithms on synthetic data, while 
practical considerations, applications and experimental results on benchmark data are further discussed in Section~\ref{sec:application_to_mlc}. The proofs of all theorems in Sections~3 and 4 can be found in an appendix.

\section{Formal Presentation of our Mathematical Framework}

In order to show formally that many existing algorithms are sub-optimal w.r.t. optimising the F-measure, we will use a mathematical framework that is closely related to frameworks encountered in classical machine learning papers. However, our analysis will slightly differ from the analysis performed in such papers, as we are investigating inference algorithms instead of training algorithms. Let us start with a formal definition of what we will call the regret of an inference algorithm. \\

\noindent {\bf Definition 2.1} \textit{Given a probability distribution $\Pr$, the regret of a vector of predictions $\vec{h}$ w.r.t.\ the F-measure is formally defined as 
\begin{eqnarray*}
R_F(\vec{h}) = \mathbb{E} \big[ F(\brY, \vec{h}_F) -  F(\brY, \vec{h}) \big] = \sum_{\vec{y} \in \{0,1\}^m} \big[ F(\vec{y}, \vec{h}_F) -  F(\vec{y}, \vec{h}) \big] \Pr(\vec{y}) \,,
\end{eqnarray*}
with $\vec{h}_F$ the F-measure maximizer in (\ref{eq:argmax_f1}).} \\

The above definition of regret (aka \emph{excess risk} in the statistical literature) can be considered as a classical tool in the framework of Bayes-risk consistency  \citep{Devroye1997,Bartlett2006}. However, let us emphasize that the theoretical analysis presented below differs from traditional techniques for investigating the consistency of machine learning algorithms. Typically, a training algorithm is considered consistent if its risk converges in probability to the Bayes risk as the training sample grows to infinity. Since many training algorithms optimize a convex and differentiable surrogate of the target loss of interest, such an analysis is often performed by bounding the risk of the target loss as a function of the surrogate \emph{$\phi$-risk} of the surrogate loss ---see e.g.\ \citep{Breiman2000b,Steinwart2001,Zhang2004,Bartlett2006,Tewari2007,Duchi2010,Gao2011}. 

In this article, we start from a different perspective, since we are analyzing the Bayes-optimality of inference algorithms. As such, we call a given loss a surrogate loss for the F-measure if an inference algorithm optimises this loss instead of the F-measure. This is different from, for example, the above papers, which analyse the consistency of surrogate losses during the training phase of a machine learning algorithm, using the surrogate loss as the internal training loss, i.e., as a convex and differentiable approximation of the target loss that is optimised during training. Furthermore, a second notable difference to other papers is that sample size convergence is less important in our analysis, as we are starting from a trained probabilistic model that is assumed to deliver consistent estimates. 

We will consider the regret of various types of loss functions and algorithms under any arbitrary probability distribution $\Pr$. By searching for probability distributions that maximize the regret, we are mainly considering the worst-case scenario. In the case of surrogate losses, we restrict this search to probability distributions that deliver unique risk minimizers; the reasons for this restriction are of technical nature. Similar to the F-measure maximizer, let us introduce the risk minimizer of a loss $L: \{0,1\}^m \times \{0,1\}^m \rightarrow \mathbb{R}_+$ as
\begin{equation}
\vec{h}_L = \argmin_{\vec{h} \in \lbrace 0,1 \rbrace^m} \mathbb{E} \left[L(\brY,\vec{h})\right]
          = \argmin_{\vec{h} \in \lbrace 0,1 \rbrace^m} \sum_{\vec{y} \in \lbrace 0,1 \rbrace^m}\Pr(\vec{y})\,L(\vec{y},\vec{h})  .
\label{eq:argmin_L}
\end{equation}
This allows us to introduce the worst-case regret formally. \\

\noindent {\bf Definition 2.2} 
\textit{Let $L: \{0,1\}^m \times \{0,1\}^m \rightarrow \mathbb{R}_+$ be a loss, let $\mathcal{P}$ be the set of all probability distributions over $\{0,1\}^m$, and let $\mathcal{P}^u_L$ 
be the subset of $\mathcal{P}$ that delivers unique solutions to 
(\ref{eq:argmin_L}). Then, the worst-case regret is defined as 
\begin{equation}\label{eqn:wcr}
\sup_{\Pr \in \mathcal{P}^u_L} \, \mathbb{E} \big[ F(\brY, \vec{h}_F) -  F(\brY, \vec{h}_L) \big] \enspace ,
\end{equation}
with $\vec{h}_F$ and $\vec{h}_L$ defined by (\ref{eq:argmax_f1}) and (\ref{eq:argmin_L}), respectively.} \\

Note that, in the above definition, we restrict the worst-case analysis to probability distributions with a unique risk minimizer for $L$. Technically, the problem would otherwise become more difficult, as it would require the comparison of the F-measure maximizer with a \emph{set} of risk minimizers $\vec{H}_L$ instead of a unique minimizer $\vec{h}_L$. This could be done in different ways, for example, by looking at the most favorable case for $L$, leading to
$$
\sup_{\Pr \in \mathcal{P}} \, \mathbb{E} \Big[ \min \big\{
 F(\brY, \vec{h}_F) -  F(\brY, \vec{h}_L) \, \vert \, 
 \vec{h}_L \in \vec{H}_L  \big\}
 \Big] \enspace ,
$$
or the least favorable one, leading to 
$$
\sup_{\Pr \in \mathcal{P}} \, \mathbb{E} \Big[ \max \big\{
 F(\brY, \vec{h}_F) -  F(\brY, \vec{h}_L) \, \vert \, 
 \vec{h}_L \in \vec{H}_L  \big\}
 \Big] \enspace .
$$
To avoid a more or less arbitrary decision, we prefer to exclude these cases from the beginning. In any case, it is clear that the regret (\ref{eqn:wcr}) provides a lower bound for any other definition of regret that maximizes over the entire set $\mathcal{P}$ of distributions. Furthermore, remark that non-uniqueness of the F-measure maximizer is unproblematic, since for the definition of the regret, only the \emph{value} of the F-measure is important, not the maximizer itself.   


Using classical notions such as Fisher consistency \citep{Wu2010}, one can say that a sufficient condition for inconsistency is encountered if (\ref{eqn:wcr}) does not evaluate to zero. However, since exact solutions or lower bounds will be derived, we are able to give much more precise information on the degree of incorrectness.

\section{The F-measure and Related Loss Functions}
Given the difficulty of maximizing the F-measure, we start our analysis by investigating a few related loss functions that have been used as surrogates in some multi-label classification and structured output prediction papers. We will analyze the Hamming loss, the subset zero-one loss and the Jaccard index. For the former two loss functions, we perform a regret analysis to show that optimizing these loss functions is not consistent if the F-measure is our performance metric of interest. For the Jaccard index, we derive a simple upper bound on the regret when optimizing the F-measure instead. 

\subsection{The Hamming Loss}
The Hamming loss can be considered as the most standard loss for multi-label classification problems -- see e.g.\ \citep{Schapire2000,Tsoumakas_and_Katalos_2007,Hariharan_et_al_2010}. It is also widely used in many structured output prediction methods -- see e.g.\ \citep{Taskar_et_al_2004,Daume_et_al_2009,Finley_Joachims_2008}. Using the general notation that was introduced above, the Hamming loss simply corresponds to the error rate in binary classification, and it can be formally defined as follows:
\begin{equation}
\label{eqn:Hamming_loss}
L_H(\vec{y}, \bh) = \frac{1}{m} \sum_{i=1}^m  \left\{ \begin{array}{cl} 1  & \quad
\textrm{if $y_i \neq  h_i(\vec{x})$} \\
0  & \quad \textrm{if $y_i =  h_i(\vec{x}) \,.$}
\end{array} \right.
\end{equation}
For the Hamming loss, 
the risk minimizer is
\begin{equation}
\vec{h}_H = \argmin_{\vec{h} \in \lbrace 0,1 \rbrace^m} \mathbb{E} \left[L_H(\brY,\vec{h})\right]
          = \argmin_{\vec{h} \in \lbrace 0,1 \rbrace^m} \sum_{\vec{y} \in \lbrace 0,1 \rbrace^m}\Pr(\vec{y})\,L_H(\vec{y},\vec{h})  
\end{equation}
is obtained by $\bh_H(\vec{x}) = (h_{H,1}, \ldots, h_{H,m})$,
where
\begin{equation}\label{eq:hm}
h_{H,i}(\vec{x}) = \argmax_{b \in \{0,1\}}  \Pr(Y_i = b) \qquad \forall i \in \{1,\ldots, m\}.
\end{equation}
Thus, in order to optimize the Hamming loss, one should select the marginal modes of $\Pr$. The following theorem presents our main result for the Hamming loss.\\

\noindent {\bf Theorem 3.1}
{\it Let $\vec{h}_H$ be a vector of predictions obtained by minimizing the Hamming loss, 
Then for $m > 2$ the worst-case regret is given by:
\begin{eqnarray*}
\sup_{\Pr \in \mathcal{P}^u_{L_H}} \, (\mathbb{E} \big[ F(\brY, \vec{h}_F) -  F(\brY, \vec{h}_H) \big]) = 0.5 \,, \\
\end{eqnarray*}
where the supremum is taken over all possible distributions $\Pr$ that result in a unique F-measure maximizer and Hamming loss minimizer.}\\

In other words, the theorem indicates that optimizing the Hamming loss as a surrogate for the F-measure results in a prediction that is far from optimal. This claim will be further confirmed by experimental results in Sections 6 and 7.  

\subsection{The subset zero-one loss}
The next multi-label loss function we analyze is the subset 0/1 loss, which generalizes the well-known 0/1 loss from the conventional to the multi-label setting:
\begin{equation}
\label{eqn:zero-one_loss}
L_s(\vec{y}, \bh) = \left\{ \begin{array}{cl} 1  & \quad
\textrm{if $\vec{y} \ne \bh$} \\
0  & \quad \textrm{if $\vec{y} = \bh $}
\end{array} \right.
\end{equation}
Admittedly, this loss function may appear overly stringent, especially in the case of many labels. Moreover, since making a mistake on a single label is punished as hardly as a mistake on all labels, it does not discriminate well between ``almost correct'' and ``completely wrong'' predictions. Still, a lot of existing frameworks for multi-label classification and structured output prediction optimize a convex upper bound on this loss in a direct or approximate manner. For example, conditional random fields optimize the log-loss as a surrogate for the subset zero-one loss \citep{Lafferty_et_al_2001}, structured support vector machines consider the structured hinge loss as a surrogate of the subset 0/1 loss when no margin/slack rescaling is performed \citep{Tsochantaridis_et_al_2005} and probabilistic classifier chains with logistic base classifiers optimize the log-loss approximately by means of pseudo-likelihood maximization \citep{Dembczynski_et_al_2010a, Demb2012b}. Moreover, maximum a posteriori (MAP) estimation techniques in Bayesian statistics and graphical models are also known to minimize the subset 0/1 loss.  

As for any other 0/1 loss, the risk-minimizing prediction for (\ref{eqn:zero-one_loss}) is simply given by the mode of the distribution:
\begin{equation}\label{eqn:subset}
\bh_s = \argmax_{\vec{y} \in \{0,1\}^m} \Pr(\vec{y})
\end{equation}
Thus, unlike the Hamming loss, looking at marginal probabilities does not suffice to minimize the subset 0/1 loss. When the independence assumption is violated, information about the joint distribution over labels is needed, similar as for the F-measure. Our interest in the subset 0/1 loss is primarily fueled by this connection. Summarized in the following theorem, we perform a similar type of regret analysis as for the Hamming loss. \\

\noindent {\bf Theorem 3.2}
{\it Let $\vec{h}_s$ be a vector of predictions obtained by minimizing the subset 0/1 loss, then for $m > 2$ the worst-case regret is given by:
\begin{eqnarray}
\label{eq:regret_subset01}
\sup_{\Pr \in \mathcal{P}^u_{L_s}} \, (\mathbb{E} \big[ F(\brY, \vec{h}_F) -  F(\brY, \vec{h}_s) \big]) = \frac{ (-m-2+2m^2)m}{(2m-1)(4+m+m^2)} \,, 
\end{eqnarray}
where the supremum is taken over all possible distributions $\Pr$.}\\

\begin{figure}
\begin{center}
\includegraphics[scale=0.4]{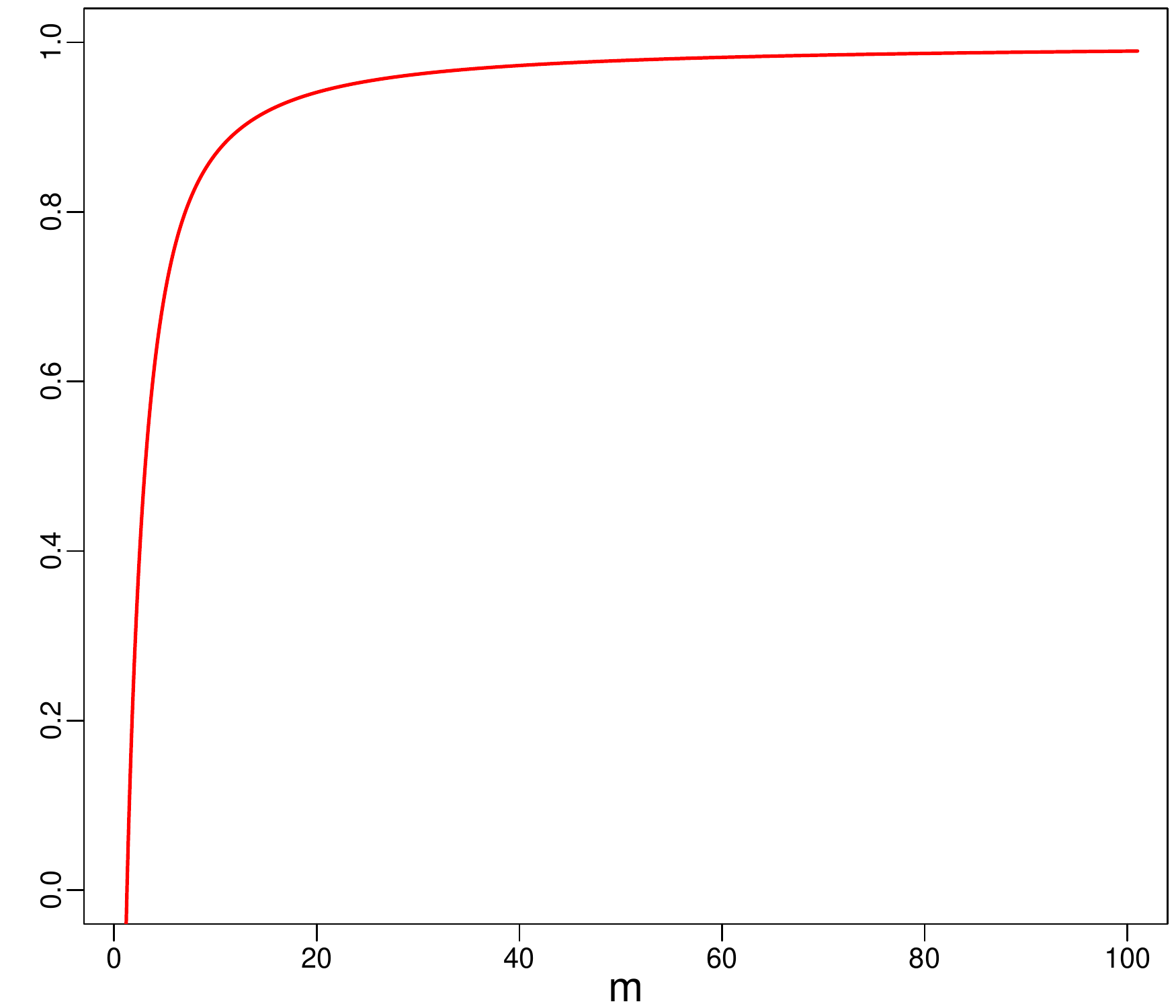}
\end{center}
\caption{Plot of the worst-case regret for the subset zero-one loss (\ref{eq:regret_subset01}) as a function of the number of labels $m$.}
\label{fig:regret}
\end{figure}

Let us remark that the worst-case regret converges rapidly to one as a function of the number of labels $m$, as illustrated in Figure~\ref{fig:regret}.
Similar to the result for the Hamming loss, the above theorem confirms that using the subset zero-one loss as an alternative for the F-measure can potentially yield a high regret. Optimizing the subset zero-one loss might hence be not a valid alternative. Our experimental results in Sections 6 and 7 will indeed make clear that such an approach performs suboptimal for several datasets. 

\subsection{The Jaccard index}
The F-measure was originally defined by set operators, as a measure for expressing the similarity of sets. In this literature, it is known as the dice coefficient  \citep{Dice1945}. Another well-known measure for expressing similarity of sets is the Jaccard index. The two measures are very related, since both belong to a more general parametric family of similarity measures for sets \citep{DeBaets2009}. The Jaccard index computes the ratio of intersection to union:
\begin{equation}
J(\vec{y}, \bh) = \frac{ |\{i \mid y_i = 1 \wedge h_i = 1, i=1,\ldots,m  \}|}{| \{i \mid y_i = 1 \vee h_i = 1, i=1,\ldots,m \}|} 
\end{equation}
Owing to a simple transformation, it can also be written as follows:\footnote{Similar as the F-measure, note that the denominator is 0 if $y_i = h_i = 0$ for all $i=1,\ldots , m$. In this case, the utility is 0 by definition.}
\begin{equation}
\label{eqn:Jaccard_distance_loss}
J(\vec{y}, \bh) = \frac{\sum_{i=1}^m y_i h_i}{\sum_{i=1}^m y_i + \sum_{i=1}^m h_i - \sum_{i=1}^m y_i h_i} 
\end{equation}
In recent years, the Jaccard index has gained popularity in the machine learning community. In the context of kernel methods, it is often used as an alternative to the linear kernel for binary feature vectors, such as fingerprints of molecules in cheminformatics and bioinformatics. In these application domains, one often speaks of the Tanimoto kernel \citep{Swamidass2005}. 

As a utility function the Jaccard index is often considered in multi-label classification. It remains an open question whether or not a closed-form solution for the risk minimizer of the Jaccard similarity exists, but the maximization is far from straightforward. Under the assumption of label independence, which allows one to transform many loss functions to a contingency table,  \citet{Quevedo2012} have recently proposed an exact algorithm for maximizing the instance-wise Jaccard similarity, as well as other loss functions that can be computed from such a contingency table. However, without this assumption, one commonly believes that exact optimization is intractable \citep{Chierichetti_et_al_2010}. As a result, the analysis that we report for the Jaccard index differs from the one reported for the Hamming loss and the subset 0/1 loss. Given that the maximization of the Jaccard index is believed to be much harder, it does not make sense to use this measure as a surrogate for the F-measure during optimization. In contrast, one might think of maximizing the F-measure as a surrogate for the Jaccard index. The following result characterizes what we can lose with such a strategy.   \\

\noindent {\bf Theorem 3.3}
{\it Let $\vec{h}_J$ and $\vec{h}_F$ be vectors of predictions obtained by maximizing the Jaccard index and the F-measure, respectively. Let the utility of the F-measure maximizer be given by
\begin{equation*}
\delta(\Pr) = \max_{\vec{h} \in \lbrace 0,1 \rbrace^m} \mathbb{E} \left[F(\brY,\vec{h})\right]
          = \max_{\vec{h} \in \lbrace 0,1 \rbrace^m} \sum_{\vec{y} \in \lbrace 0,1 \rbrace^m}\Pr(\vec{y})\,F(\vec{y},\vec{h})  .
\end{equation*}
The regret of the F-measure maximizer with respect to the Jaccard index is then upper bounded by 
\begin{eqnarray*}
 \mathbb{E} \big[J(\brY, \vec{h}_J) -  J(\brY, \vec{h}_F) \big] \leq 1 - \delta(\Pr) / 2  \, \\
\end{eqnarray*}
for all possible distributions $\Pr$}.\\

Notwithstanding that the above upper bound on the regret remains rather loose, the observation is interesting because the upper bound on the regret decreases as a function of the predictive performance, which gives a good indication of the signal strength in the data. In fact the utility of the F-measure maximizer $\delta(\Pr)$ might represent the predictive performance in a practical situation, as measured for a hypothetical machine learning method and dataset.  Due to the relationship between Jaccard index and F-measure, a high utility for the F-measure implies that optimizing this measure as a surrogate for the Jacccard similarity might be a reasonable thing to do. Practically, this will be mainly the case for datasets with relatively few noise.

\section{Existing Algorithms for F-Measure Maximization}
\label{sec:F-measure}

The previous section revealed that optimizing more conventional loss functions as surrogates for the F-measure might result in a poor predictive performance. In this section, we perform a similar type of analysis for more specialized algorithms that intend to solve~(\ref{eq:argmax_f1}). These algorithms make different types of assumptions to simplify the problem. First of all, the algorithms operate on a constrained hypothesis space, sometimes justified by theoretical arguments. Secondly, they only guarantee optimality for specific distributions $\Pr$. 


\subsection{Algorithms Based on Label Independence}


By assuming independence of the random variables $Y_1,...,Y_m$, optimization problem (\ref{eq:argmax_f1}) can be substantially simplified.
It has been shown independently in \citep{lewis95} and \citep{jansche07} that the optimal solution then always contains the labels with the highest marginal probabilities, or no labels at all. 
As a consequence, only a few hypotheses $\vec{h}$ ($m\!+\!1$ instead of $2^m$) need to be examined, and the computation of the expected F-measure can be performed in an efficient way. \\

\noindent {\bf Theorem 4.1} [\citep{lewis95}]
{\it Let $Y_1, Y_2, \ldots, Y_m$ be independent Bernoulli variables with parameters $p_1, p_2, \ldots, p_m$ respectively. 
Then, for all $j,k \in \{1, \ldots, m\}$, $h_{F,j} = 1$ and $h_{F,k} = 0$ implies $p_j \ge p_k$.}\\

In addition, \citet{lewis95} showed that the expected F-measure can be approximated by the following expression under the assumption of independence
$$
\mathbb{E} \left[F(\brY,\vec{h})\right] \simeq \left \{
\begin{array}{ll}
\prod_{i=1}^m (1 - p_i), & \textrm{if~~} \vec{h} = \vec{0}_m \\
\frac{2 \sum_{i=1}^m p_i h_i}{\sum_{i=1}^m p_i + \sum_{i=1}^m h_i}, & \textrm{if~~} \vec{h} \ne \vec{0}_m
\end{array} \right .
$$
This approximation is exact for $\vec{h} = \vec{0}$, while for $\vec{h} \ne \vec{0}$, an upper bound of the error can easily be determined~\citep{lewis95}.

However, \citet{Chai2005}, \citet{jansche07} and \citet{Quevedo2012} have independently proposed exact procedures for computing the F-maximizer. To this end, independence is assumed and marginal probabilities $p_1, p_2, \ldots, p_m$ serve as input for the algorithms. The method of Jansche runs in $\mathcal{O}(m^4)$, while the other two approaches solve the same problem more efficiently in $\mathcal{O}(m^3)$. 

As a starting point for explaining the three algorithms, notice that (\ref{eq:argmax_f1}) can be solved via outer and inner maximization. Namely, (\ref{eq:argmax_f1}) can be transformed into an inner maximization
\begin{equation}\label{eq:inner}
\vec{h}^{(k)} = \argmax_{\vec{h} \in H_k} \mathbb{E} \left[ F(\brY, \vec{h}) \right],
\end{equation}
where $H_k = \lbrace \vec{h} \in \lbrace 0,1 \rbrace^m \, \vert \, \sum_{i=1}^m h_i = k \rbrace$, followed by an outer maximization
\begin{equation}\label{eq:outer}
\vec{h}_F = \argmax_{\vec{h} \in \lbrace \vec{h}^{(0)}, \ldots, \vec{h}^{(m)} \rbrace} \mathbb{E} \left[ F(\brY, \vec{h}) \right].
\end{equation}
The outer maximization (\ref{eq:outer}) can be done by simply checking all $m+1$ possibilities. The main effort is then devoted to solving the inner maximization~(\ref{eq:inner}).
According to Lewis' theorem, to solve~(\ref{eq:inner}), one needs to check only one vector $\vec{h}$ for a given $k$, in which $h_i = 1$ for the $k$ labels with highest marginal probabilities $p_i$. The remaining problem is the computation of the expected F-measure in (\ref{eq:inner}).
This expectation cannot be computed naively, as the sum is over exponentially many terms. But the F-measure is a function of integer counts that are bounded, so it can normally only assume a much smaller number of distinct values. It has been further shown that the expectation has a domain with a cardinality exponential in $m$; but since the cardinality of its range is polynomial in $m$, it can be computed in polynomial time. As a result, \citet{jansche07} obtains an algorithm that is cubic in $m$ for computing~(\ref{eq:inner}), resulting in an overall $\mathcal{O}(m^4)$ time complexity. 
He also presents an approximate version of this procedure, reducing the complexity from cubic to quadratic. This approximation leads to an overall complexity of $\mathcal{O}(m^3)$, but it does no longer guarantee optimality of the prediction. 

As a more efficient alternative, the procedure of \citet{Chai2005} is based on ordering the labels according to the marginal probabilities. For $\vec{h}^{(k)} \in H_k$, thus $h_i = 1$ for $k$ labels with the greatest marginal probabilities, he derives the following expression:
$$
\mathbb{E} \left[F(\brY,\vec{h}^{(k)})\right] = 2 \prod_{i=1}^{m} (1 - p_i) I_1(m)\; ,
$$
where $I_1(m)$ is given by the following recurrence equations and boundary conditions:
\begin{eqnarray*}
I_t(a) & = & I_{t+1}(a) + r_t I_{t+1}(a + 1) + r_t J_{t+1}(a + 1) \\
J_t(a) & = & J_{t+1}(a) + r_t J_{t+1}(a + 1) \\
 & & I_{k+1}(a) = 0 \quad J_{m + 1}(a) = a^{-1} 
\end{eqnarray*}
with $r_i = p_i/(1 - p_i)$.
These equations suggest a dynamic programming algorithm of space $\mathcal{O}(m)$ and time $\mathcal{O}(m^2)$ in computing the expected F-measure for given~$k$. This yields an overall time complexity of $\mathcal{O}(m^3)$. 

In a more recent follow-up paper, \citet{Ye2012} further improved the dynamic programming algorithm to an $\mathcal{O}(m^2)$ complexity by additional sharing of internal representations. The old and the new version of the algorithm both rely on Lewis' theorem and a factorization of the probability mass for constructing recurrence equations, hence leaving few hope for extending the algorithm to situations where label independence cannot be assumed. In another recent paper, \citet{Quevedo2012} propose a general inference procedure that utilizes similar recurrence equations and dynamic programming techniques. In contrast to \citep{Chai2005}, \citep{jansche07} and \citep{Ye2012}, the authors primarily address multi-label classification problems, focusing on a wide range of loss functions that can be computed from a contingency table in an instance-wise manner. As a result, the instance-wise F-measure is maximized as a special case, while assuming label independence.

If the independence assumption is violated, none of the above methods is able to guarantee optimality. In the most general case, the F-maximizer needs to be computed by analyzing the joint distribution. The above methods rely on modeling or ordering marginal probabilities, which is not sufficient to compute the F-maximizer for many distributions. This is illustrated by the following example, in which two joint distributions with identical marginal probabilities have different F-measure maximizers:
\begin{center}
\begin{tabular}{@{}cc@{}}
$\vec{y}$ & $\Pr(\vec{y})$ \\
\hline
0001 & 0.1 \\
0010 & 0.2 \\
0100 & 0.2 \\
1000 & 0.5 \\
\end{tabular}
$\quad$
\begin{tabular}{@{}cc@{}}
$\vec{y}$ & $\Pr(\vec{y})$ \\
\hline
0000 & 0.5 \\
1001 & 0.1 \\
1010 & 0.2 \\
1100 & 0.2 \\
\end{tabular}
\end{center}
The non-specified configurations have zero probability mass. For both distributions, we have $p_1 = \Pr(\rY_1 = 1) = 0.5$, $p_2 = \Pr(\rY_2 = 1) = 0.2$, $p_3 = \Pr(\rY_3 = 1) = 0.2$, $p_4 = \Pr(\rY_4 = 1) = 0.1$, but one can easily check that the F-measure maximizers are $\vec{h} = (1000)$ and $\vec{h} = (0000)$, respectively. The regret is small for this simple example, but methods that assume independence may produce predictions being far away from the optimal one. The following result shows this concretely. \\

\noindent
{\bf Theorem 4.2}
{\it Let $\vec{h}_I$ be a vector of predictions obtained by assuming label independence as defined in (\ref{eq:label_independence}), then the worst-case regret is lower-bounded by:
\begin{eqnarray*}
\sup_{\Pr} \, (\mathbb{E} \big[ F(\brY, \vec{h}_F) -  F(\brY, \vec{h}_I) \big]) \geq 2q -1 , \\
\end{eqnarray*}
for all $q \in [1/2,1]$ satisfying
$\sum_{s=1}^m  \big( \frac{2m!}{(m-s)! (s-1)! (m + s)} q^{m-s} (1-q)^s \big)  - q^m > 0 $ and the supremum taken over all possible distributions $\Pr$.} \\

For increasing $m$, the condition is satisfied for $q$ close to one (see the appendix for details).   In such a scenario, the worst-case regret is lower bounded by $R_q = 2 q - 1$, so that $\lim_{q \rightarrow 1, m \rightarrow \infty} R_q = 1$.
As a consequence, the lower bound becomes tight in the limit of $m$ going to infinity, as summarized in the following corollary. \\

\noindent
{\bf Corollary 4.3}
{\it Let $\vec{h}_I$ be a vector of predictions obtained by assuming independence, then the worst-case regret converges to $1$ in the limit of $m$, i.e.,
\begin{eqnarray*}
\lim_{m \rightarrow \infty} \sup_{\Pr} \, (\mathbb{E}  \big[ F(\brY, \vec{h}_F) -  F(\brY, \vec{h}_I) \big]) = 1 , \\
\end{eqnarray*}
where the supremum is taken over all possible distributions $\Pr$.} \\

Again we will show by means of experiments in Sections 6 and 7 that algorithms based on label independence can be suboptimal on real-world datasets. 

\subsection{Algorithms Based on the Categorical Distribution}

Solving (\ref{eq:argmax_f1}) becomes straightforward in the case of a specific distribution in which the probability mass is distributed over vectors $\vec{y}$ containing only a single positive label, i.e., $\sum_{i = 1}^m y_i = 1$, corresponding to the categorical distribution. This was studied in~\citep{delcoz_et_al_2009} in the setting of so-called \emph{non-deterministic classification}. \\

\noindent {\bf Theorem 4.4} [\citep{delcoz_et_al_2009}]
{\it Denote by $\vec{y}(i)$ a vector for which $y_i = 1$ and all the other entries are zeros. Assume that $\Pr$ is a joint distribution such that $\Pr(\brY = \vec{y}(i)) = p_i$. The maximizer $\vec{h}$ of (\ref{eq:argmax_f1}) consists of the $k$ labels with the highest marginal probabilities, where $k$ is the first integer for which
$$
\sum_{j=1}^k p_j  \ge (1 + k) p_{k+1};
$$
if there is no such integer, then $\vec{h} = \vec{1}$, where $\vec{1}$ is a vector of all ones.}\\

The categorical distribution reflects the case of multi-class classification problems, as a special case of multi-label classification problems. The above approach is only applicable to such problems.  

\subsection{Algorithms that use both the Marginal and the Joint Distribution}\label{sec:threshold}

Since all the methods so far rely on the fact that the optimal solution contains ones for the labels with the highest marginal probabilities (or consists of a vector of zeros), one may expect that thresholding on the marginal probabilities ($h_i(\theta) = 1$ for $p_i \ge \theta$, and $h_i(\theta) = 0$ otherwise) will provide a solution to (\ref{eq:argmax_f1}) in general. Practically, despite using marginal probabilities for the thresholds, such a scenario does not assume label independence anymore, because also the joint probability distribution $\Pr$ must be provided.\footnote{In fact most thresholding methods that optimize the F-measure during training do not use the joint distribution, and define a threshold based on marginals only. However, that is in practice the same as assuming independence, and resembles the same conclusions as in Section 4.1. In contrast, the regret will be lower when also the joint distribution is used to define the threshold. When the F-measure is optimized in an inference phase, starting from a trained probabilistic model, access to the joint distribution is of course needed.} When the labels are ordered according to the marginal probabilities, thus $p_i \geq p_{i+1}$ for all $i \in \{1,...,m-1\}$, this approach resembles the following optimization problem: 
\begin{equation*}
\vec{h}_T = \argmax_{\theta \in \lbrace 1, p_1, \ldots, p_m \rbrace} \mathbb{E} \left[ F(\brY, \vec{h}(\theta)) \right].
\end{equation*}
Thus, to find an optimal threshold $\theta$, access to the entire joint distribution is needed. However, this is not the main problem here, since in the next section, we will show that only a polynomial number of parameters of the joint distribution is needed.
What is more interesting is the observation that the F-maximizer is in general not consistent with the order of marginal label probabilities. In fact, the regret can be substantial, as shown by the following result.\\


\noindent {\bf Theorem 4.5}
{\it Let $\vec{h}_T$ be a vector of predictions obtained by putting a threshold on sorted marginal probabilities, then the worst-case regret is lower bounded by
\begin{eqnarray*}
\sup_{\Pr} \, (\mathbb{E} \big[ F(\brY, \vec{h}_F) -  F(\brY, \vec{h}_T) \big]) \geq \max \left(0,\frac{1}{6} - \frac{2}{m+4} \right) , 
\end{eqnarray*}
where the supremum is taken over all possible distributions $\Pr$.}\\

Finding the exact value of the supremum in the worst case is for the above formulation an interesting open question. The statement is a surprising result in light of the existence of many algorithms that rely on finding a threshold for maximizing the F-measure~\citep{Keerthi_et_al_2007,fan07,zhang10} -- remark that those methods rather seek for a threshold on socring functions instead of marginal probabilities. While being justified by Theorems~4.1, 4.2 and~4.4 for specific applications, thresholding does not yield optimal predictions in general. Let us illustrate this with an example for which $m=12$: 

\begin{center}
\begin{tabular}{@{}cc@{}}
$\vec{y}$ & $\Pr(\vec{y})$ \\
\hline
000000000000 & 0.21 \\
100000000000 & 0.39 \\
011111100000 & 0.2 \\
010000011111 & 0.2 \\
\end{tabular}
\end{center}
The non-specified configurations have zero probability mass. The F-measure maximizer is given by $(1000000000000)$; yet, not the first label but the second one exhibits the highest marginal probability. The regret remains rather low in this case, but higher values can be easily obtained by constructing more complicated examples from (\ref{eq:prob_dist}) -- see the appendix.


\section{An Exact Algorithm for F-Measure Maximization}
\label{sec:an_efficient_algorithm}

We now introduce an exact and efficient algorithm for computing the F-maximizer without using any additional assumption on the probability distribution $\Pr$.
%
%
While adopting the idea of decomposing the problem into an outer and an inner maximization, our algorithm differs 
in the way the inner maximization is solved.\footnote{The description of the method slightly differs from the previous paper~\citep{Dembczynski_et_al_2011}, and it is concordant with~\citep{Dembczynski_et_al_2013}.}
%
%
For convenience, let us introduce the following quantities:
$$
s_{\vec{y}} = \sum_{i=1}^m y_i, \quad \delta^u_{ik} = \sum_{\vec{y}:y_i=u} \frac{2\Pr(\vec{y})}{s_{\vec{y}} + k}. 
$$
The first quantity gives the number of ones in the label vector $\vec{y}$, while $\Delta^u_{ik}$ is a specific marginal value for $i$-th label, which for $u=1$ corresponds to weighted true positives. 
Using these quantities, we show that only $m^2 + 1$ parameters of the joint distribution $\Pr(\vec{y})$ are needed to compute the F-maximizer. \\

\noindent {\bf Theorem 5.1}
{\it The solution of (\ref{eq:argmax_f1}) can be computed by solely using $\Pr(\vec{y} = \vec{0})$ and the values of $\Delta^u_{ik}$, for $i,k \in  
\{1,\ldots,m\}$, 
that constitute an $m \times m$ matrix $\mx{\Delta}$.}

\begin{proof}
The inner optimization problem (\ref{eq:inner}) can be formulated as follows:
$$
\vec{h}^{(k)} =  \argmax_{\vec{h} \in H_k} \mathbb{E} \left[ F(\brY, \vec{h}) \right] = \argmax_{\vec{h} \in \mathcal{H}_k} \sum_{\vec{y} \in \{0,1\}^m} \Pr(\vec{y}) \frac{2 \sum_{i=1}^m y_ih_i}{s_{\vec{y}} + k} \enspace .
$$
The sums in $\argmax$ can be swapped, resulting in
$$
\sum_{\vec{y} \in \{0,1\}^m} \Pr(\vec{y}) \frac{2 \sum_{i=1}^m y_ih_i}{s_{\vec{y}} + k} = \sum_{i=1}^m  h_i \sum_{\vec{y} \in \{0,1\}^m} \frac{2\Pr(\vec{y})y_i}{s_{\vec{y}} + k} = 
\sum_{i=1}^m  h_i \sum_{\vec{y}: y_i = 1 } \frac{2\Pr(\vec{y})}{s_{\vec{y}} + k}  \,.
$$
Finally, we obtain
\begin{equation}
\vec{h}^{(k)} =  \argmax_{\vec{h} \in H_k}  \sum_{i=1}^m  h_i \Delta^1_{ik} \,.
\label{eqn:inner_final}
\end{equation}

As a result, one does not need the whole distribution to find the maximizer of the F-measure, but the values of $\Delta_{ik}$, which can be given in the form of an $m \times m$ matrix $\mx{\Delta}$. For the special case of $k = 0$, we have $\vec{h}^{(k)} = \vec{0}$ and $\mathbb{E}_{\vec{y} \sim \Pr({\vec{y}})} \left[ F(\vec{y}, \vec{0}) \right] = \Pr(\vec{y} = 0)$.
\end{proof}


If the matrix $\mx{\Delta}$ is given, the solution of the F-measure maximization~(\ref{eq:argmax_f1}) is straight-forward, since for each inner maximization the problem boils down to selecting the $k$ labels with the highest $\Delta^1_{ik}$. The resulting algorithm, referred to as General F-measure Maximizer (GFM), is summarized in Algorithm~1 and its time complexity is analyzed in the following theorem.\\

\begin{algorithm}[t]
\caption{General F-measure Maximizer}
\label{alg:greedy}
\begin{algorithmic}
\STATE {\bf INPUT}: matrix $\mx{\Delta}$ and probability $\Pr(\vec{y} = 0)$
\vskip2pt
\FOR{$k = 1$ to $m$}
\vskip2pt
\STATE \textbf{solve} the inner optimization problem~(\ref{eq:inner}):
 $$
	\vec{h}^{(k)} = \argmax_{\vec{h} \in H_i} \; \sum_{i=1}^m h_i \Delta^1_{ik} \,
 $$
by setting $h_i = 1$ to $k$ labels with the highest $\Delta^1_{ik}$ (in case of ties take any $k$ top labels), and $h_i = 0$ for the rest;
\vskip2pt
\STATE \textbf{store} a value of
$$
	\mathbb{E} \left[ F(\brY, \vec{h}^{(k)}) \right] =  \sum_{i=1}^m h^{(k)}_i \Delta^1_{ik};
$$
\ENDFOR

\STATE \textbf{define}  $\vec{h}^{(0)} = \vec{0}$, and $\mathbb{E} \left[F(\brY, \vec{0}) \right] = \Pr(\vec{y} = 0)$;
\vskip2pt
\STATE \textbf{solve} the outer optimization problem~(\ref{eq:outer}):
$$
\vec{h}_F = \argmax_{\vec{h} \in \lbrace \vec{h}^{(0)}, \ldots, \vec{h}^{(m)} \rbrace} \mathbb{E} \left[ F(\brY, \vec{h}) \right];
$$
\vskip2pt
\STATE \textbf{return} $\vec{h}$ and $\mathbb{E} \left[ F(\brY, \vec{h}) \right]$;
\end{algorithmic}
\end{algorithm}

\noindent {\bf Theorem 5.2}
{\it Algorithm~1 solves problem~(\ref{eq:argmax_f1}) in time $\mathcal{O}(m^{2})$ assuming that the matrix $\mx{\Delta}$ of $m^2$ parameters and $\Pr(\vec{y} = 0)$ are given.}

\begin{proof}
To solve (\ref{eqn:inner_final}), it is enough to find the top $k$ elements (i.e., the elements with the highest values) in the $k$-th column of matrix $\mx{\Delta}$, which can be carried
out in linear time~\citep{Cormen_et_al_2001}. This step has to be repeated for all $k$. Therefore, the overall complexity of the inner maximization is quadratic. The solution of the outer optimization problem (\ref{eq:outer}) is then straight-forward and requires linear time.
\end{proof}

In light of combining the inference algorithm with particular training algorithms, like multinomial regression as we discuss it in Section~\ref{sec:parametric_models},  it could be reasonable to redefine the formulation in the following way. Consider the probabilities 
\begin{equation}
p_{is} = \Pr(y_i = 1, s_{\vec{y}} = s), \quad i,s \in \{1,\ldots,m\} \,,
\label{eq:matrix_P}
\end{equation}
that constitute an $m \times m$ matrix $\mx{P}$. Let us also introduce an $m \times m$ matrix $\mx{W}$  with elements
\begin{equation}
w_{sk} = \frac{2}{(s + k)} \enspace , \qquad s,k \in \{1, \ldots, m\} \,.
\label{eq:elements_of_W}
\end{equation}
It can be easily shown that 
\begin{equation}
\mx{\Delta} = \mx{P} \mx{W},
\label{eqn:matrix_multiplication}
\end{equation}
since 
$$
\Delta^1_{ik} = \sum_{\vec{y}: y_i = 1 } \frac{2\Pr(\vec{y})}{s_{\vec{y}} + k} = \sum_{s =1}^m \frac{2p_{is}}{s + k}.
$$

If the matrix $\mx{P}$ is taken as an input by the algorithm, then its complexity is dominated by the matrix multiplication (\ref{eqn:matrix_multiplication}) that is solved naively in $\mathcal{O}(m^{3})$, but faster algorithms working in $\mathcal{O}(m^{2.376})$ are known~\citep{Coppersmith_Winograd_1990}.\footnote{The complexity of the Coppersmith-Winograd algorithm \citep{Coppersmith_Winograd_1990} is more of theoretical significance, since practically this algorithm outperforms the na\"ive method only for huge matrices.}

Interestingly, the above results clearly suggest that the F-measure maximizer is more affected by the number of 1s in the $\vec{y}$-vectors than by the interdependence between particular labels. In other words, modeling of pairwise or higher degree dependences between labels is not necessary to obtain an optimal solution, but a proper estimation of marginal quantities ($\Delta^1_{ik}$, or $p_{is}$) that take the number of co-occurring labels into account.

In the reminder of this section, we discuss the properties of the GFM algorithm in comparison to the other algorithms discussed in Section~4. The methods presented in \citep{Chai2005,jansche07} and \citep{Ye2012} all assume label independence and produce exactly the same result, apart from small numerical instabilities that might always occur. These methods, contrary to GFM, will not deliver an exact F-maximizer if the assumption of independence is violated. On the other hand, the disadvantage of GFM is the quadratic number of parameters it requires as input, while the other methods only need $m$ parameters. Since the estimation of a larger number of parameters is statistically more difficult, it is a~priori unclear which method performs better in practice. We are facing here a common trade-off between an approximate method on better estimates (we need to estimate a smaller number of parameters from a given sample) and an exact method on potentially weaker estimates. Nonetheless, if the joint distribution is concentrated on a small number of different label combinations $\vec{y}$, the estimates of $\mx{\Delta}$ or $\mx{P}$ can be as good as the estimates of the marginal probabilities $p_i$.

From the computational perspective, Jansche's method is characterized by a much higher time complexity, being respectively $\mathcal{O}(m^{4})$ and $\mathcal{O}(m^{3})$ for the exact and the approximate versions. The method of Chai has a cubic complexity, and the enhanced version presented in \citep{Ye2012} is more efficient, since it solves the problem in ${O}(m^{2})$ time. The GFM algorithm is quite competitive, as its complexity is of ${O}(m^{2})$ or ${O}(m^{3})$, depending on the setting. Moreover, the cubic complexity of GFM, which follows from the matrix multiplication, can be further decreased if the number of distinct values of $s_{\vec{y}}$ with non-zero probability mass is smaller than $m$.

\section{Simulations}
\label{sec:simulations}

In the previous sections, we gave theoretical results concerning the performance of different inference methods in the worst case scenarios. Here, we verify the methods empirically on synthetic data to check the difference in average performance on two large classes of distributions. The first class assumes independence of labels, while the second class uses a model with strong label dependences. 

We test four inference methods optimal for different performance measures. The first one is suited for Hamming loss. It estimates the marginal probabilities by simple counting from a given sample and gives empirical marginal modes  as output. We denote this method MM, since it estimates marginal modes. The second one is tailored for subset 0/1 loss. It seeks for the joint mode by checking the frequencies of label combinations appearing in the sample. We refer to this method as JM, since it estimates the joint mode. The two remaining methods are suited for F-measure maximization. We use the dynamic programming method of~\citet{Ye2012} that assumes label independence, denoted by FM, and the exact GFM method described in the previous section, which performs exact inference. All parameters required by these algorithms are also estimated from the sample by simple counting. We verify the performance of the inference methods by using Hamming loss, subset 0/1 loss, the Jaccard index, and the F-measure. 

We run these simulations, as well as the other experiments described later in this paper, on a Debian virtual machine with 8-core x64 processor and 5GB RAM.

\subsection{Label independence}

\begin{figure}[ht!]
\begin{tabular}{@{}cc@{}}
\multicolumn{2}{c}{Hamming loss} \\
\includegraphics[width=.45\textwidth]{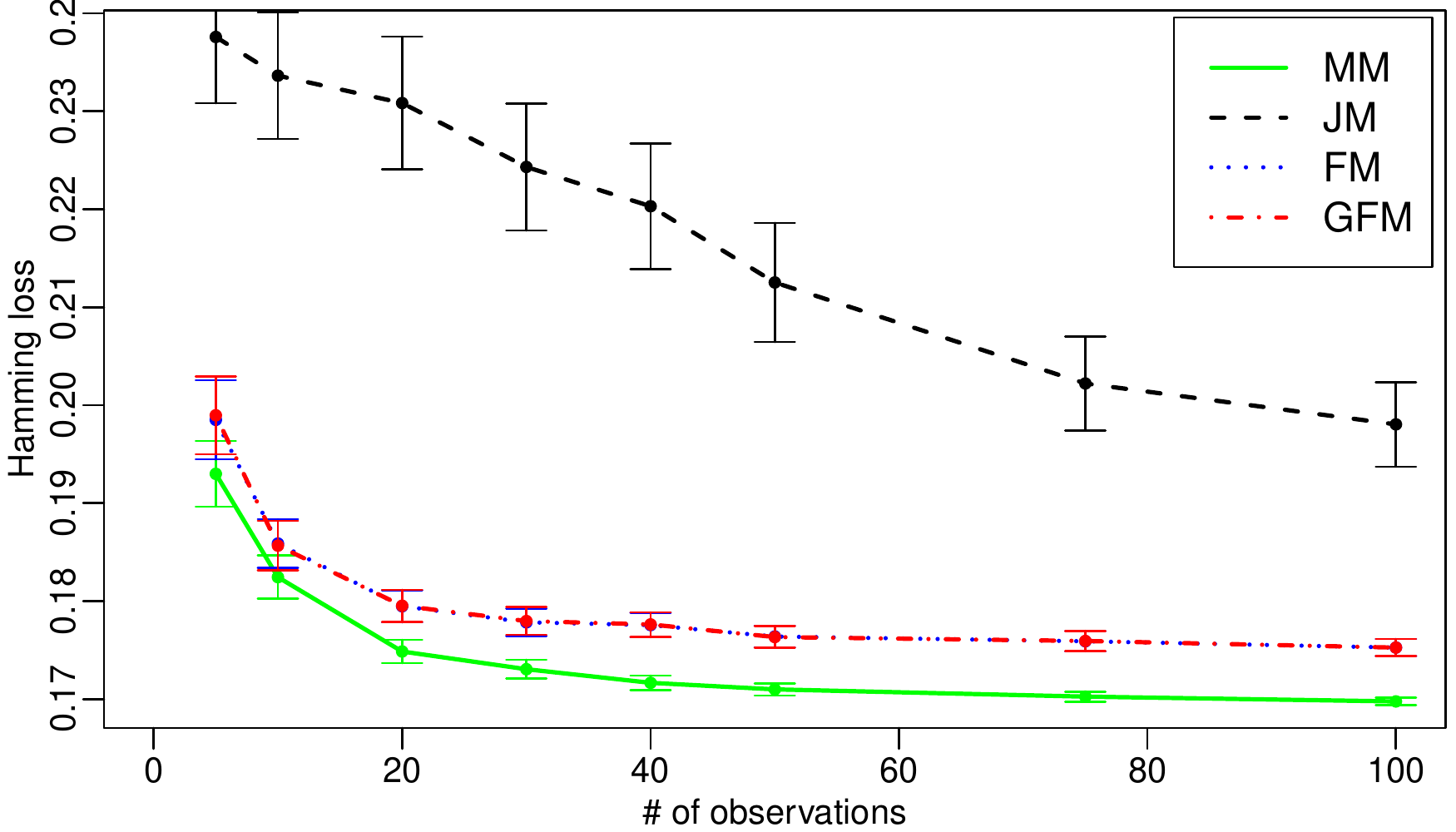} & \includegraphics[width=.45\textwidth]{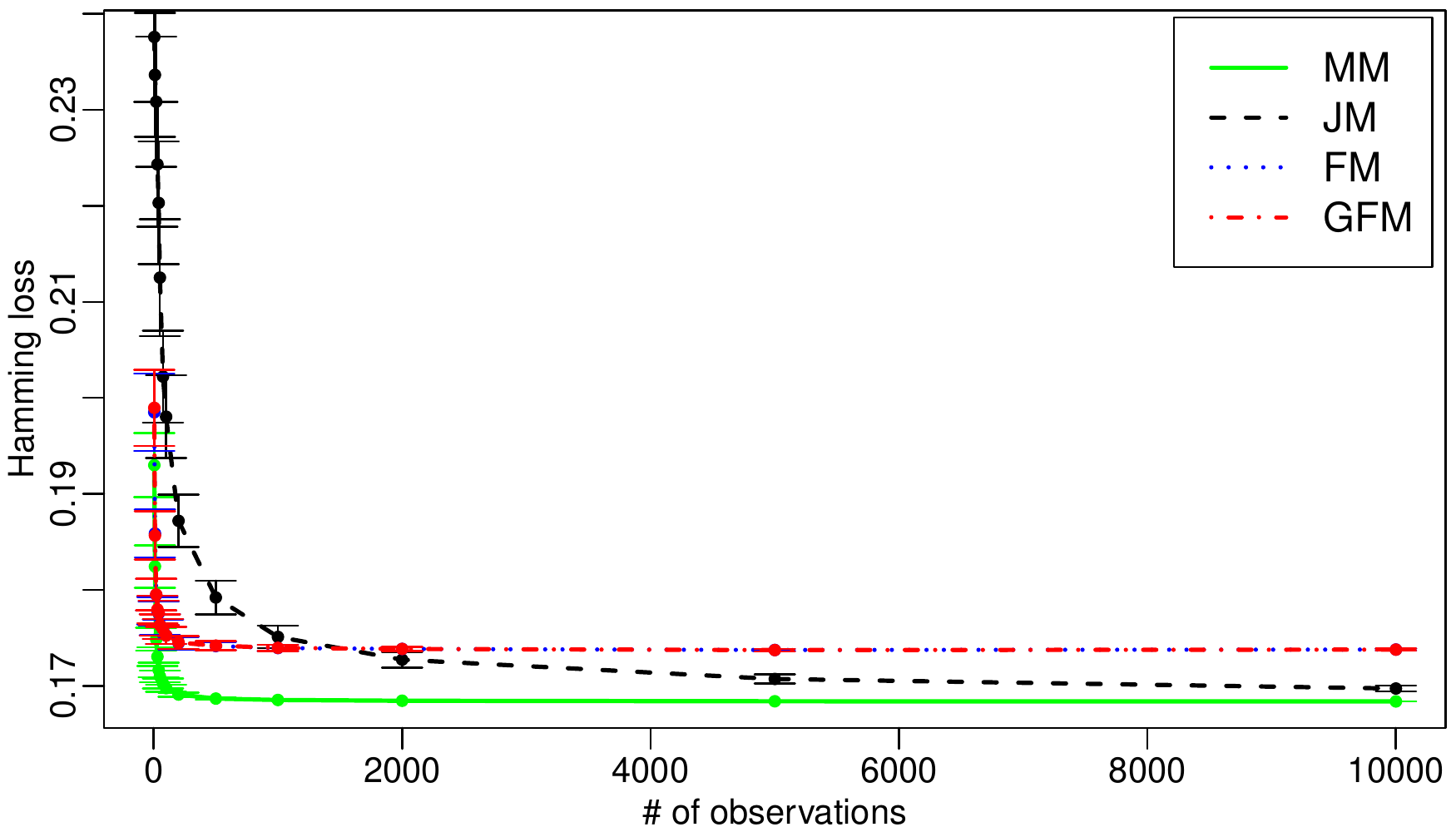}\\
\multicolumn{2}{c}{subset 0/1 loss} \\
\includegraphics[width=.45\textwidth]{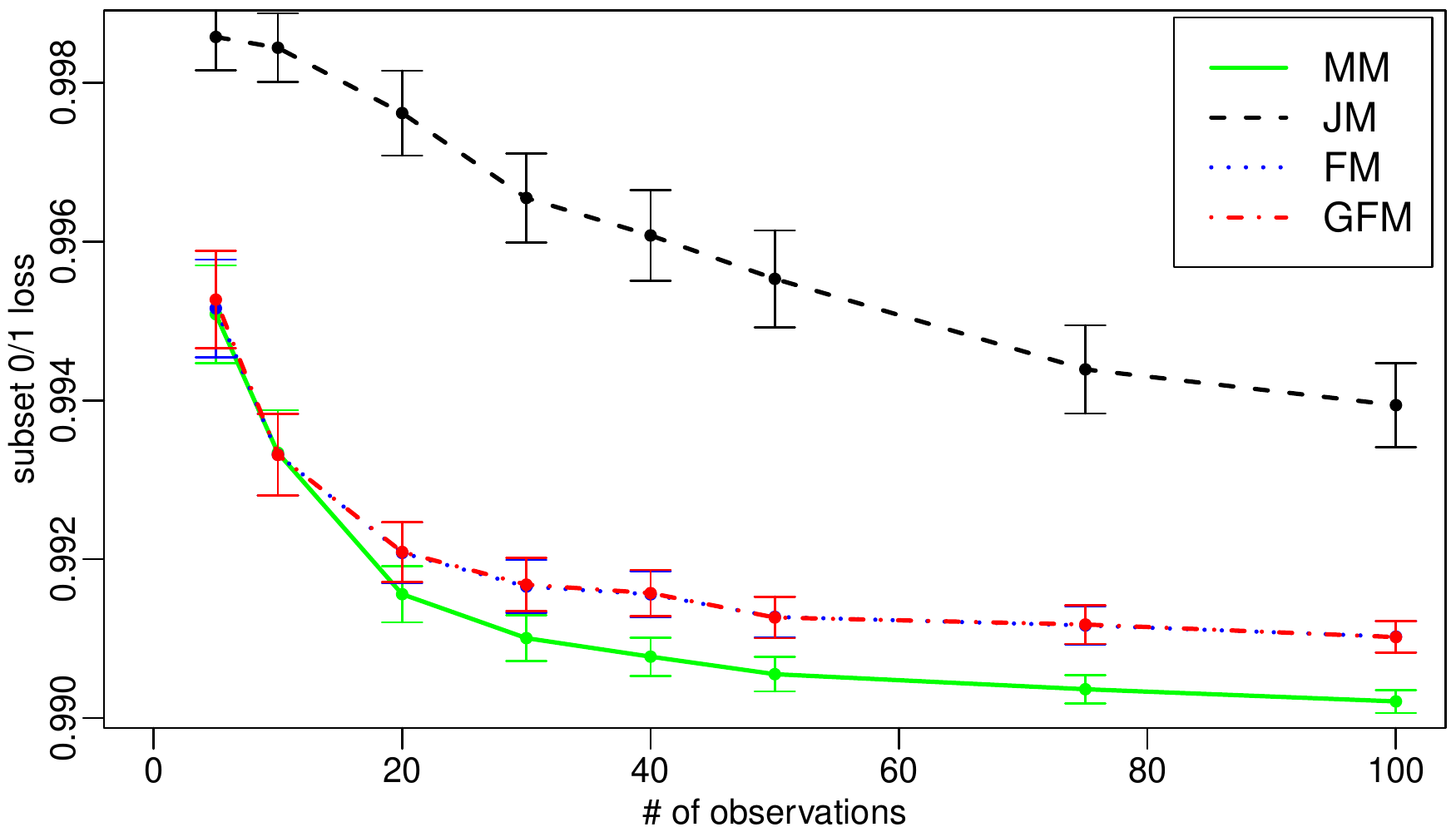} & \includegraphics[width=.45\textwidth]{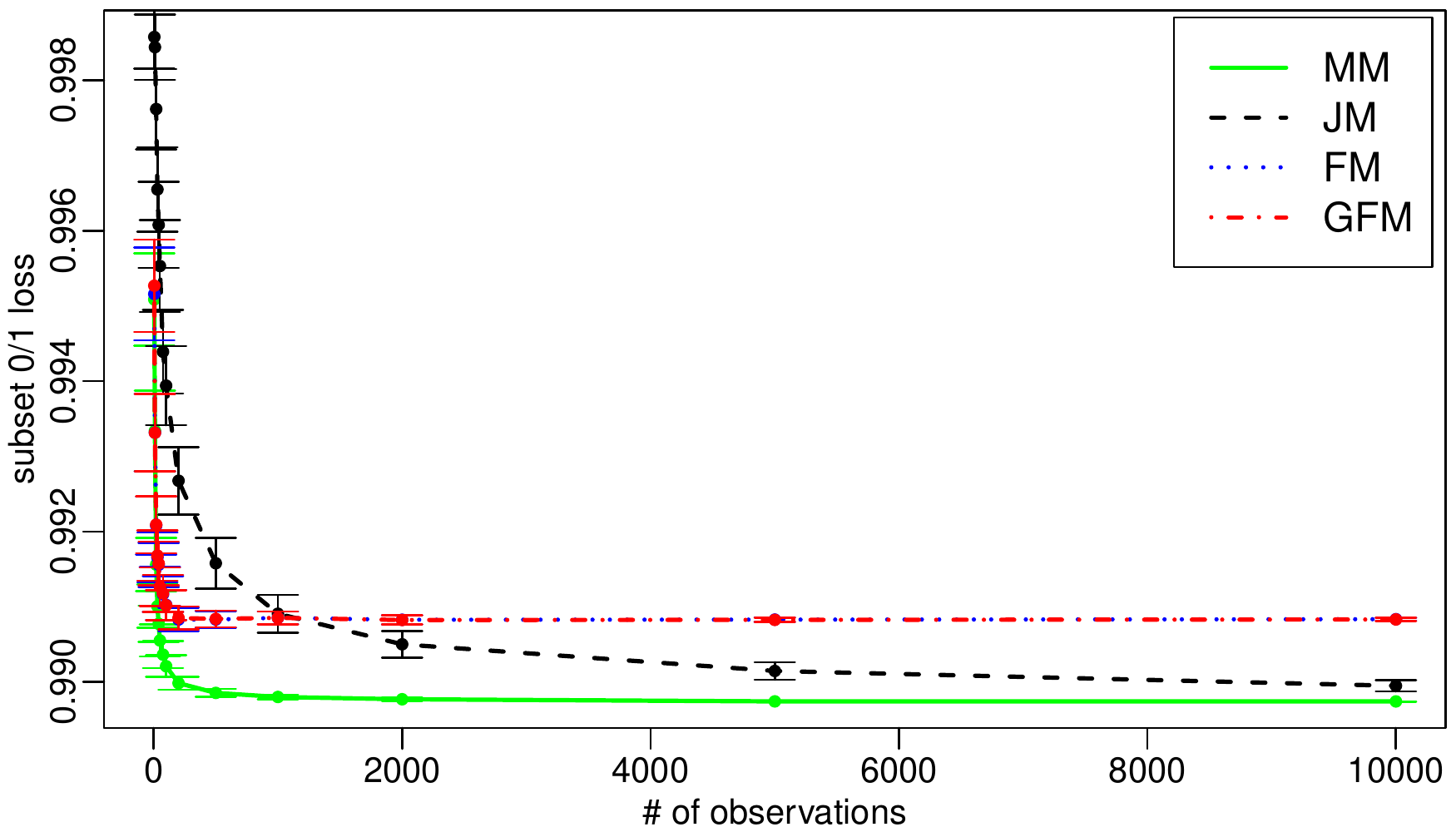}\\
\multicolumn{2}{c}{F-measure} \\
\includegraphics[width=.45\textwidth]{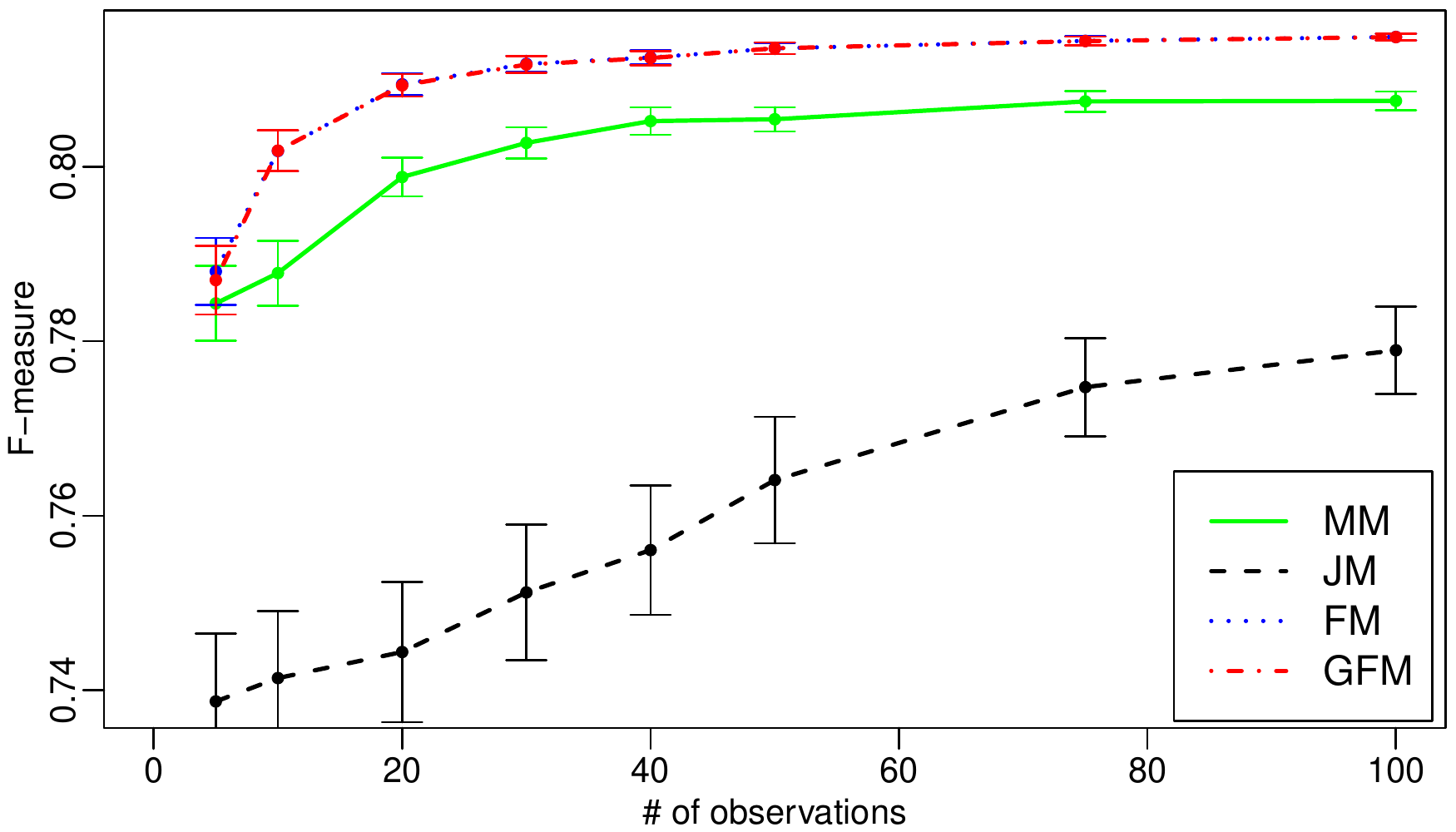} & \includegraphics[width=.45\textwidth]{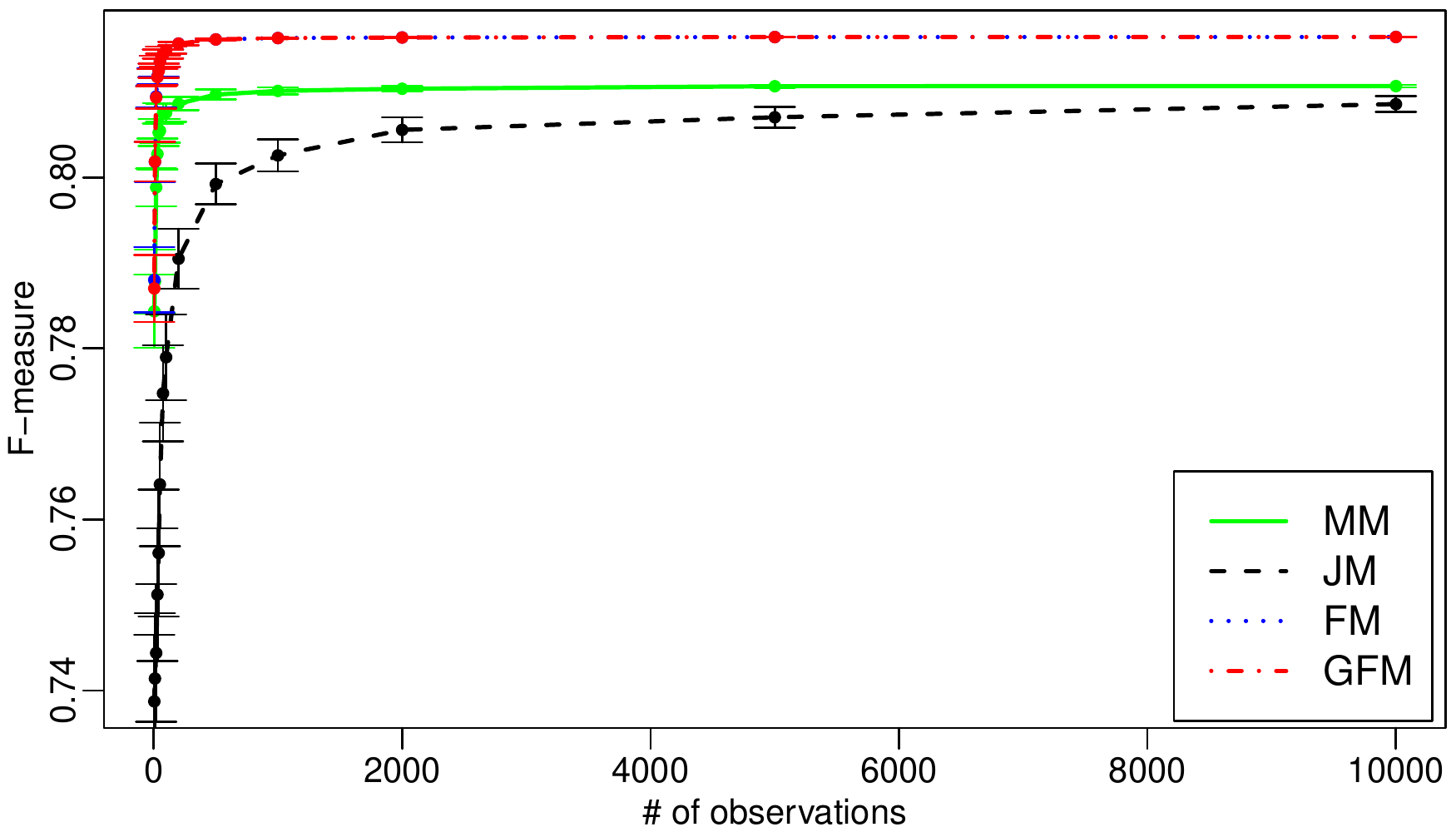}\\
\multicolumn{2}{c}{Jaccard index} \\
\includegraphics[width=.45\textwidth]{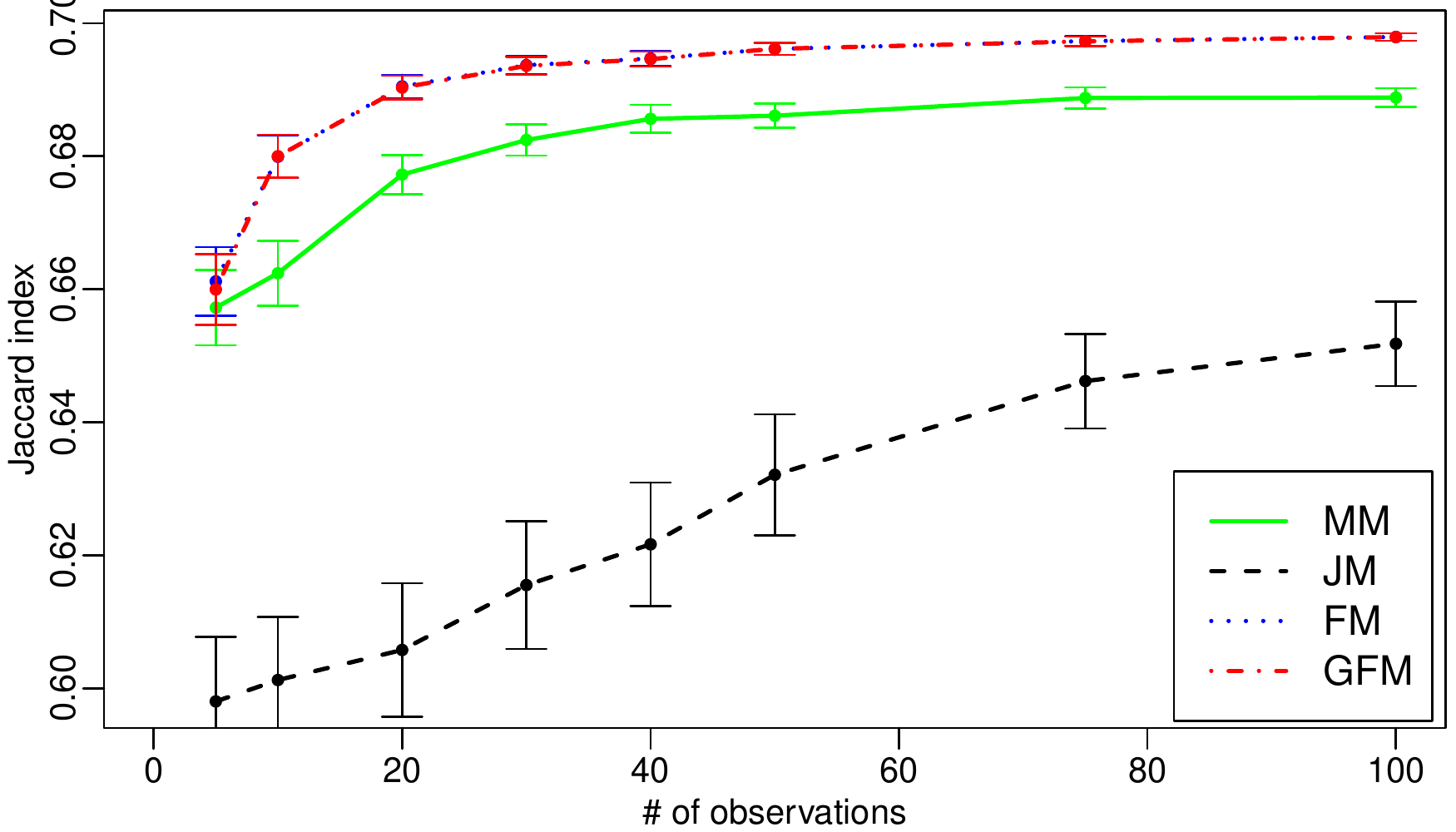} & \includegraphics[width=.45\textwidth]{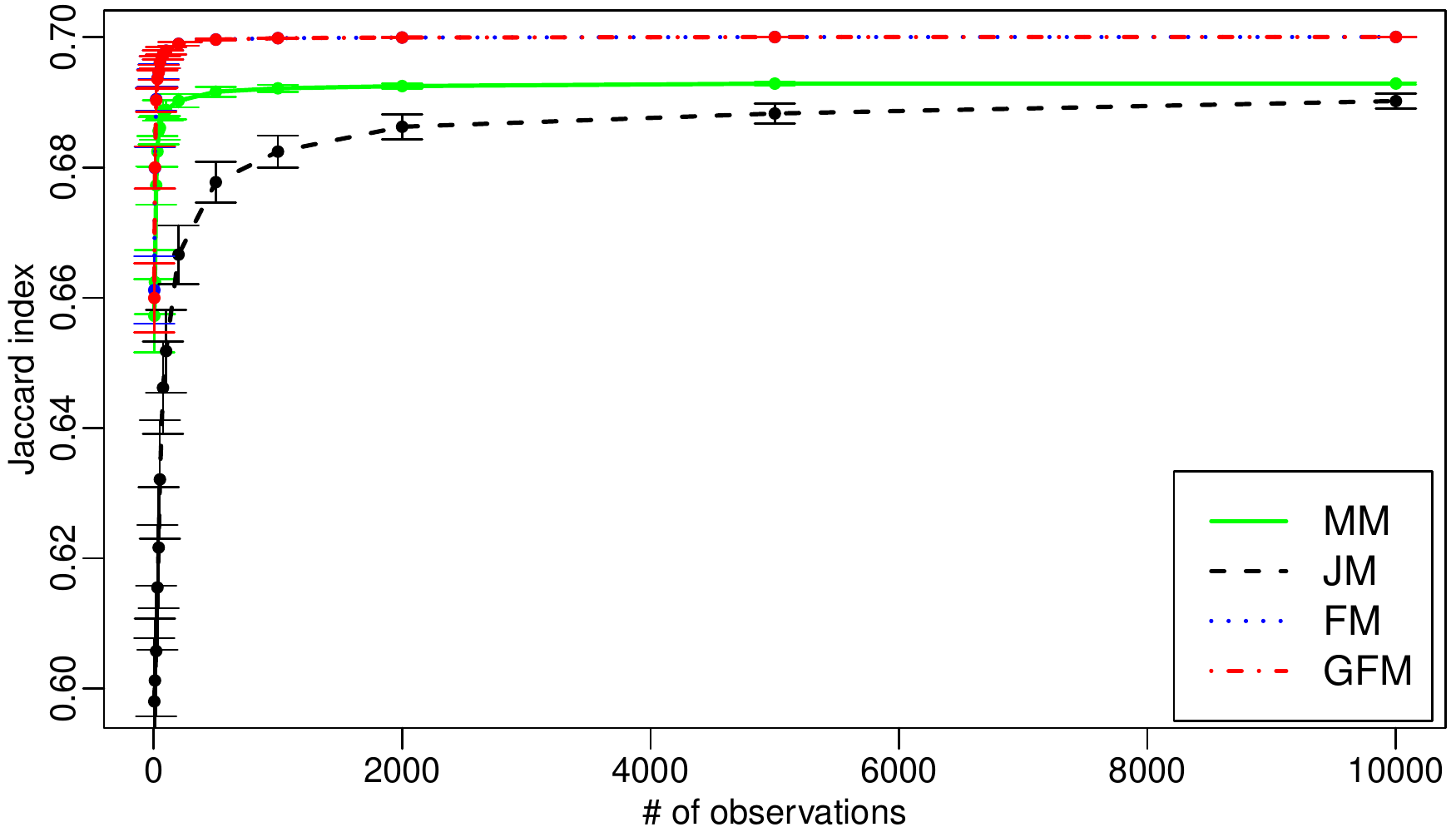}\\
\end{tabular}
\caption{Performance of inference methods in case of label independence as a function of  the number of training observations. Left: the performance up to 100 training observations. Right: the performance up to 10000 training observations.  The error bars show the standard error of the measured quantities.}
\label{fig:independent_data}
\end{figure}

The independent data are generated according to:
$$
\Pr(\vec{y}) = \prod_{i=1}^m \Pr(y_i)\,,
$$
where the probabilities $\Pr(y_i)$ are given by the logistic model:
$$
\Pr(y_i = 1) = \frac{1}{1 + \exp(- w_i)}, \qquad \textrm{where~}  w_i \sim N(0,3) \,.
$$
In experiments we set the number of labels to 25 and vary the number of observations using the following values $\{5, 10, 20, 30, 40, 50, 75, 100, 200, 500, 1000, 2000, 5000, 10000\}$. We repeat the experiment for 30 different models, i.e., sets of values $w_i$.  For each model, we use 50 different training sets of the same size, but to reduce the variance of the results we use for testing one set of 100,000 observations. 
The results are given in Figure~\ref{fig:independent_data}. The right column presents the performance with respect to different measures as a function of the number of training observations. The left column gives the same results, but zoomed to the range from 5 to 100 training observations.
We see from the plots that MM and JM get the best results for Hamming loss and subset 0/1 loss. Since the labels are independent, the marginal modes and joint mode are the same. Therefore, for large enough training samples, these two algorithms converge to the same prediction that should be optimal for Hamming and subset 0/1. However, JM coverges much slower, since it directly estimates the joint mode, by checking the frequencies of label combinations in the training set. FM and GFM perform very similarly for each performance measure. Since the labels are independent, FM and GFM should converge to the same prediction, being optimal for the F-measure. GFM, however, may get slightly worse results for small sample sizes, since it needs to estimate a larger number of parameters than FM. We also see that algorithms maximizing the F-measure perform the best for Jaccard index.   

\subsection{Strong label dependence}

\begin{figure}[ht!]
\begin{tabular}{cc@{}}
\multicolumn{2}{c}{Hamming loss} \\
\includegraphics[width=.45\textwidth]{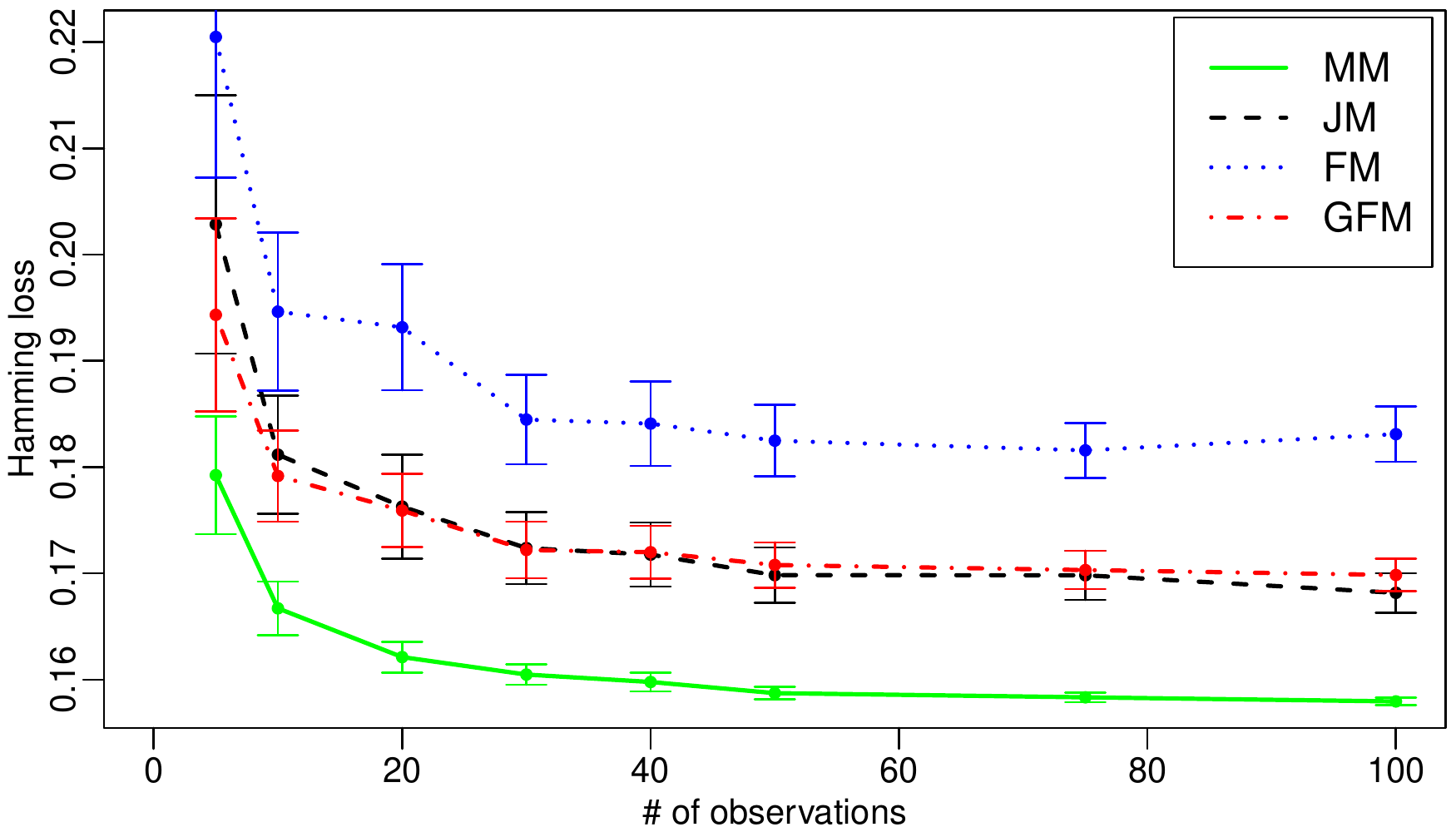} & \includegraphics[width=.45\textwidth]{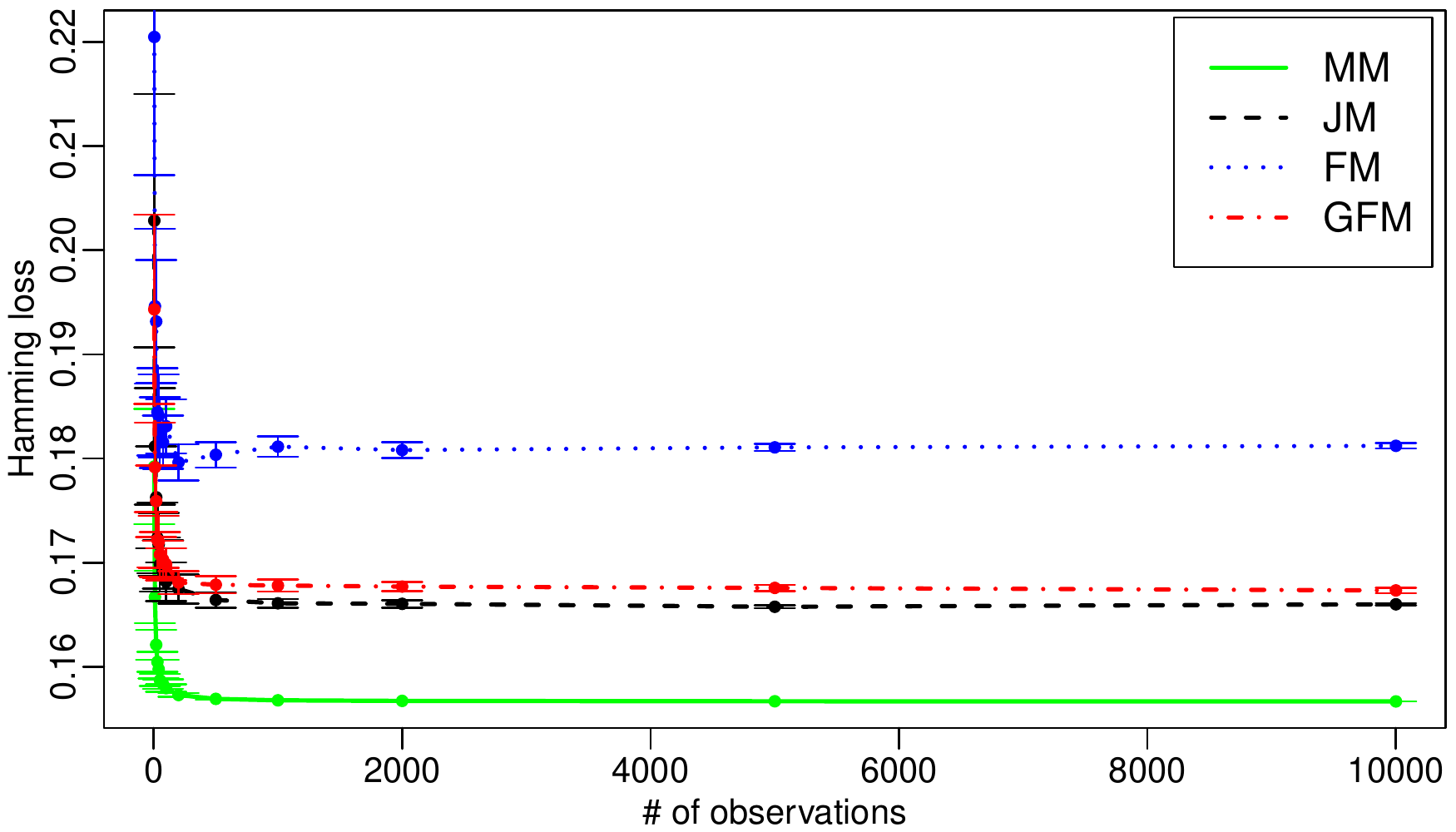}\\
\multicolumn{2}{c}{subset 0/1 loss} \\
\includegraphics[width=.45\textwidth]{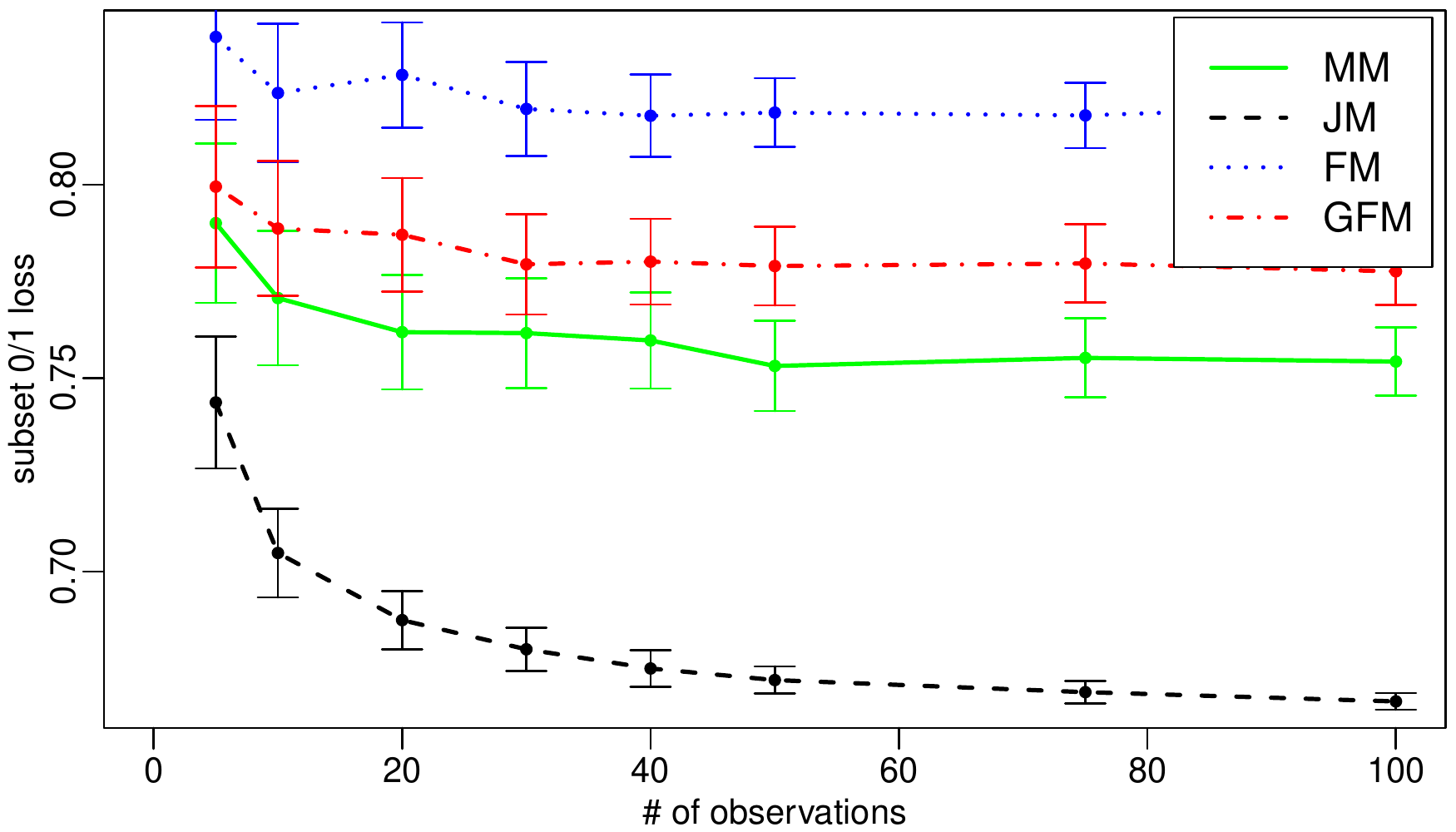} & \includegraphics[width=.45\textwidth]{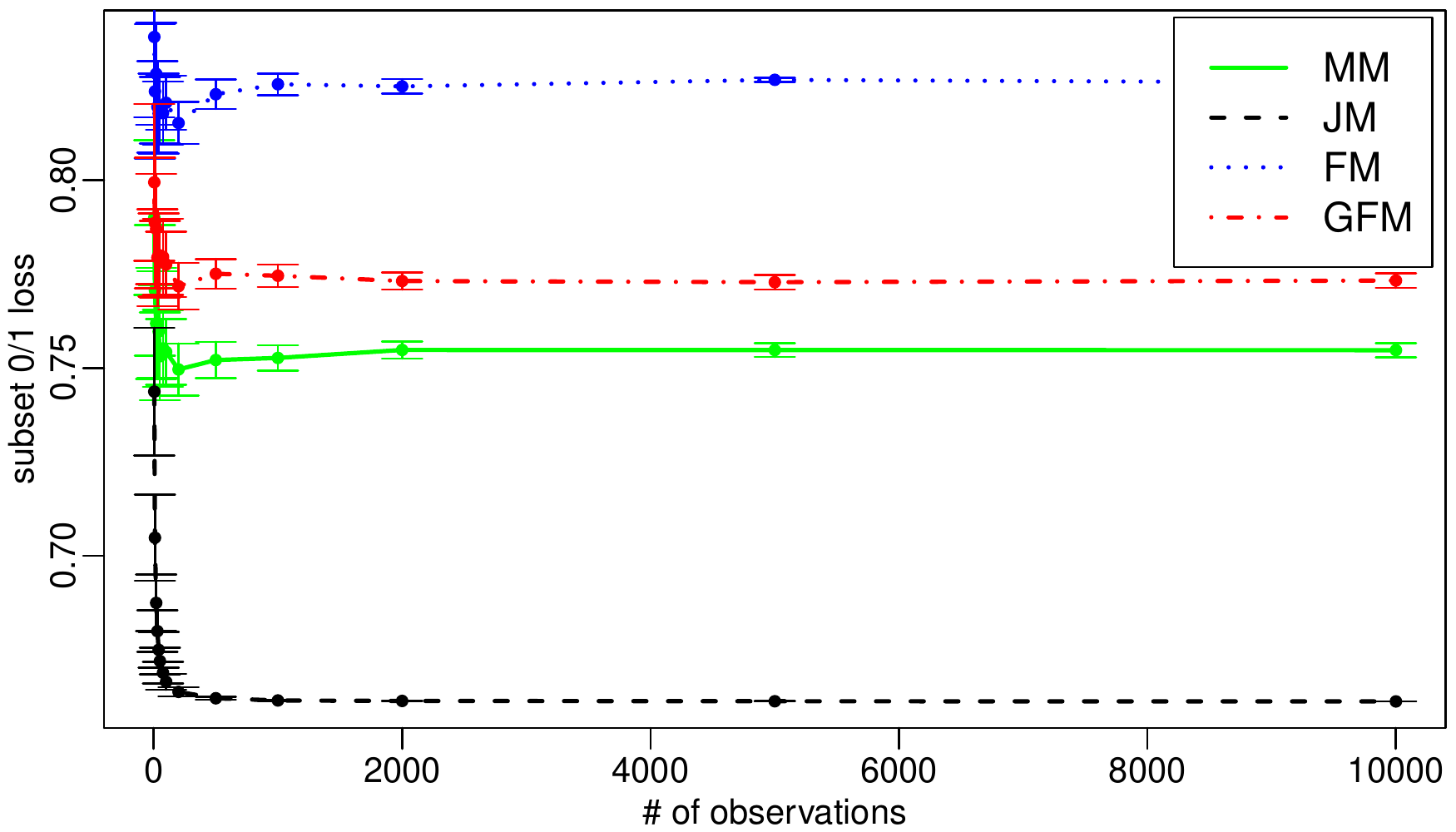}\\
\multicolumn{2}{c}{F-measure} \\
\includegraphics[width=.45\textwidth]{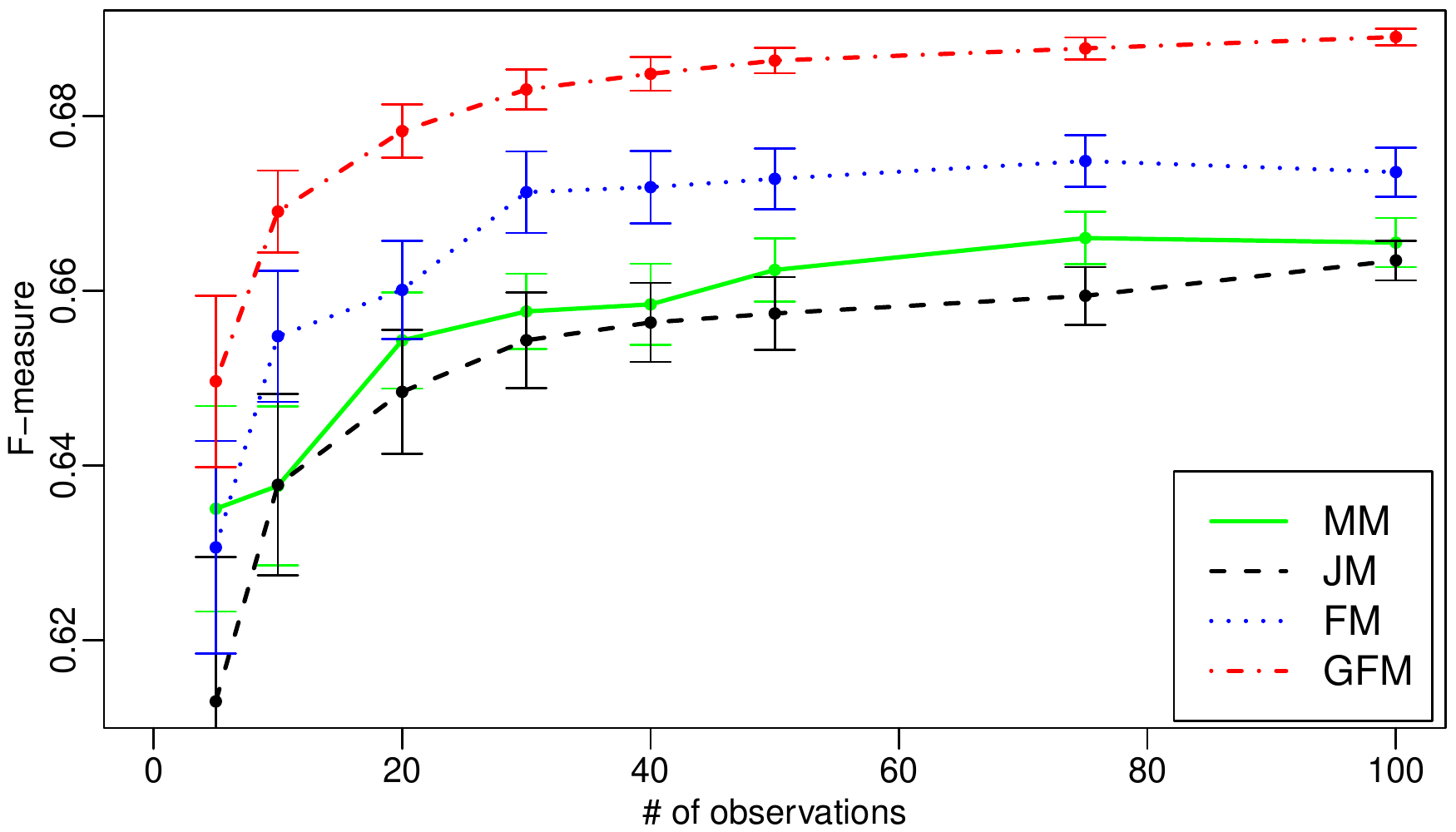} & \includegraphics[width=.45\textwidth]{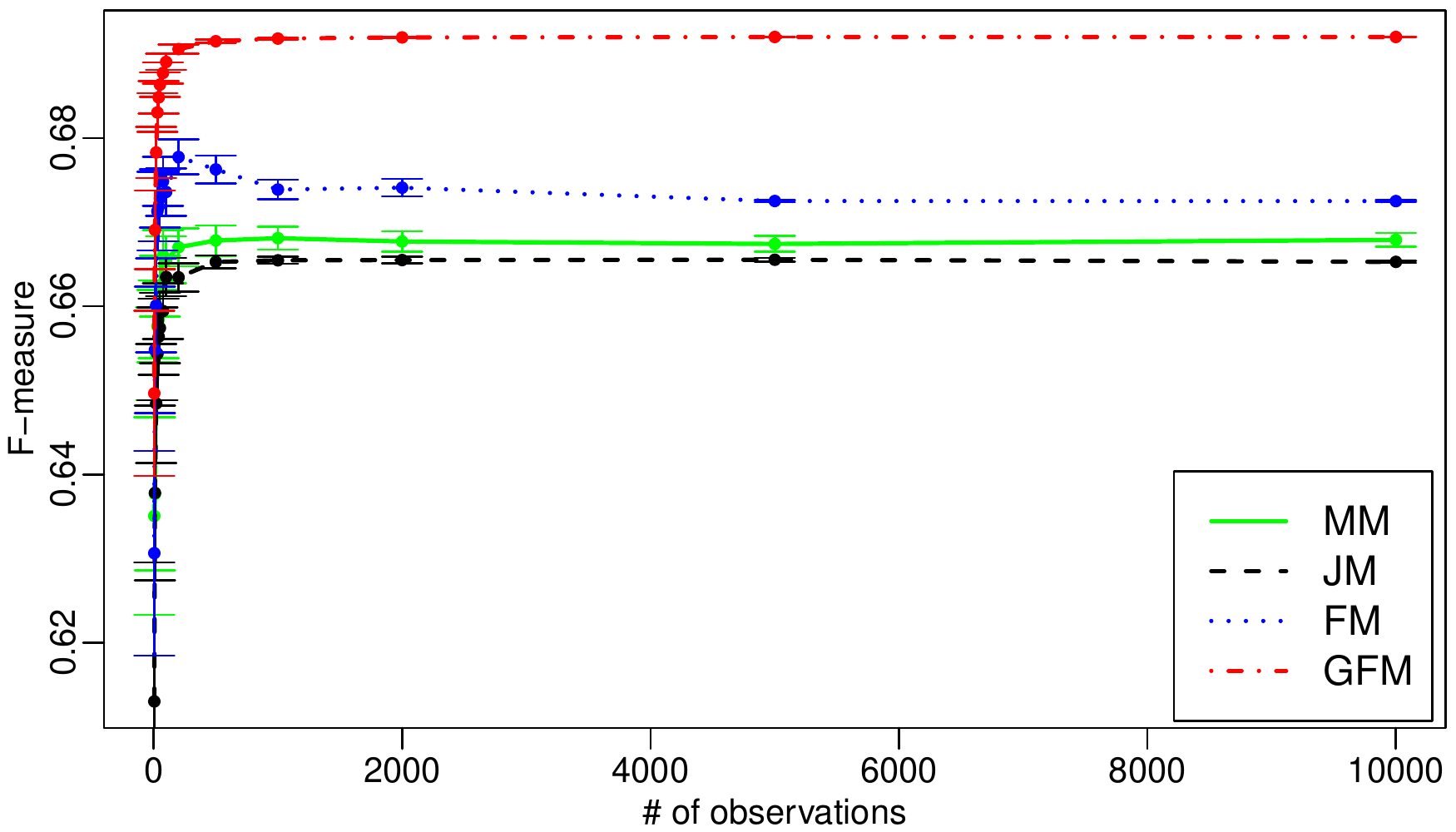}\\
\multicolumn{2}{c}{Jaccard index} \\
\includegraphics[width=.45\textwidth]{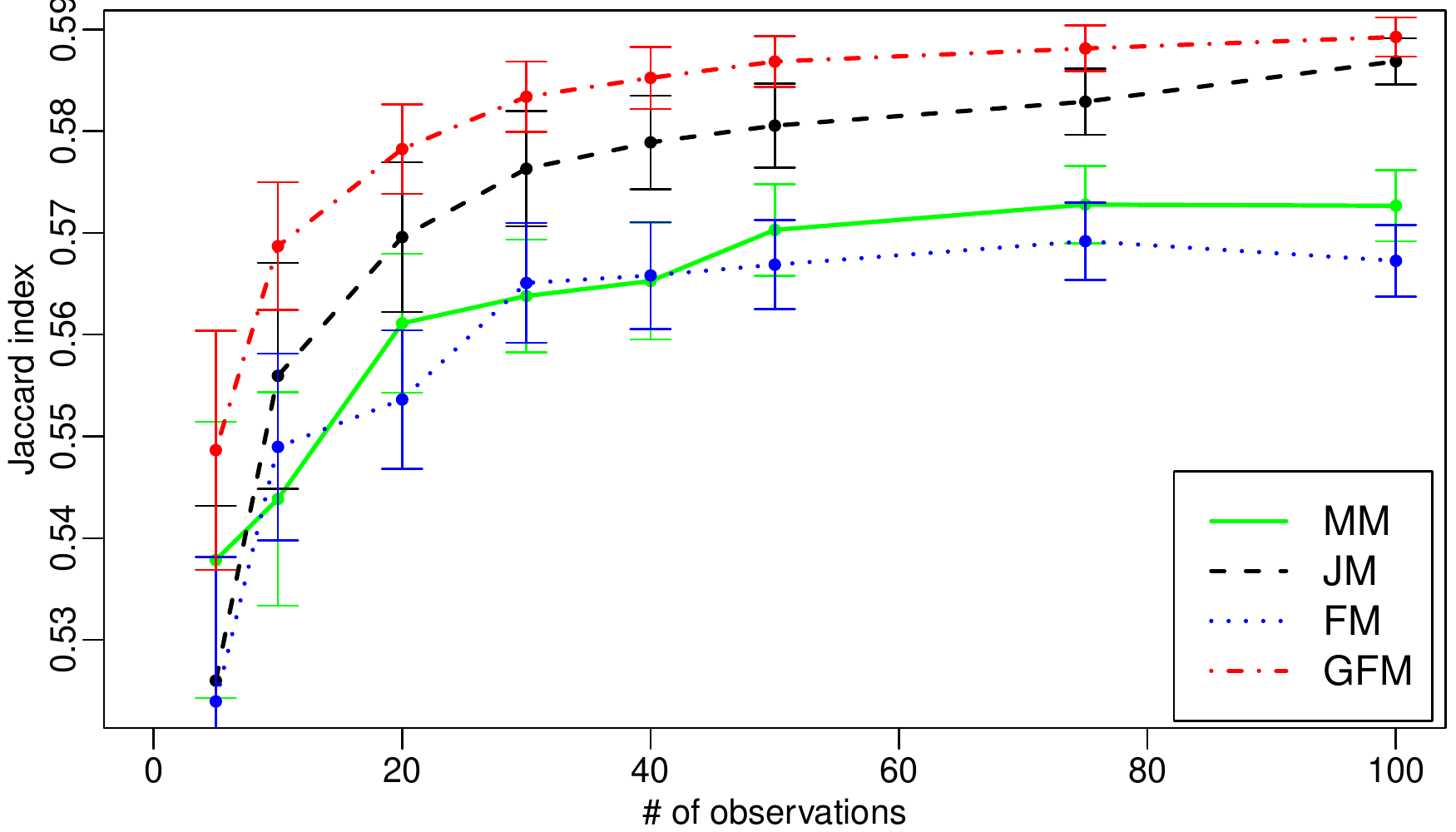} & \includegraphics[width=.45\textwidth]{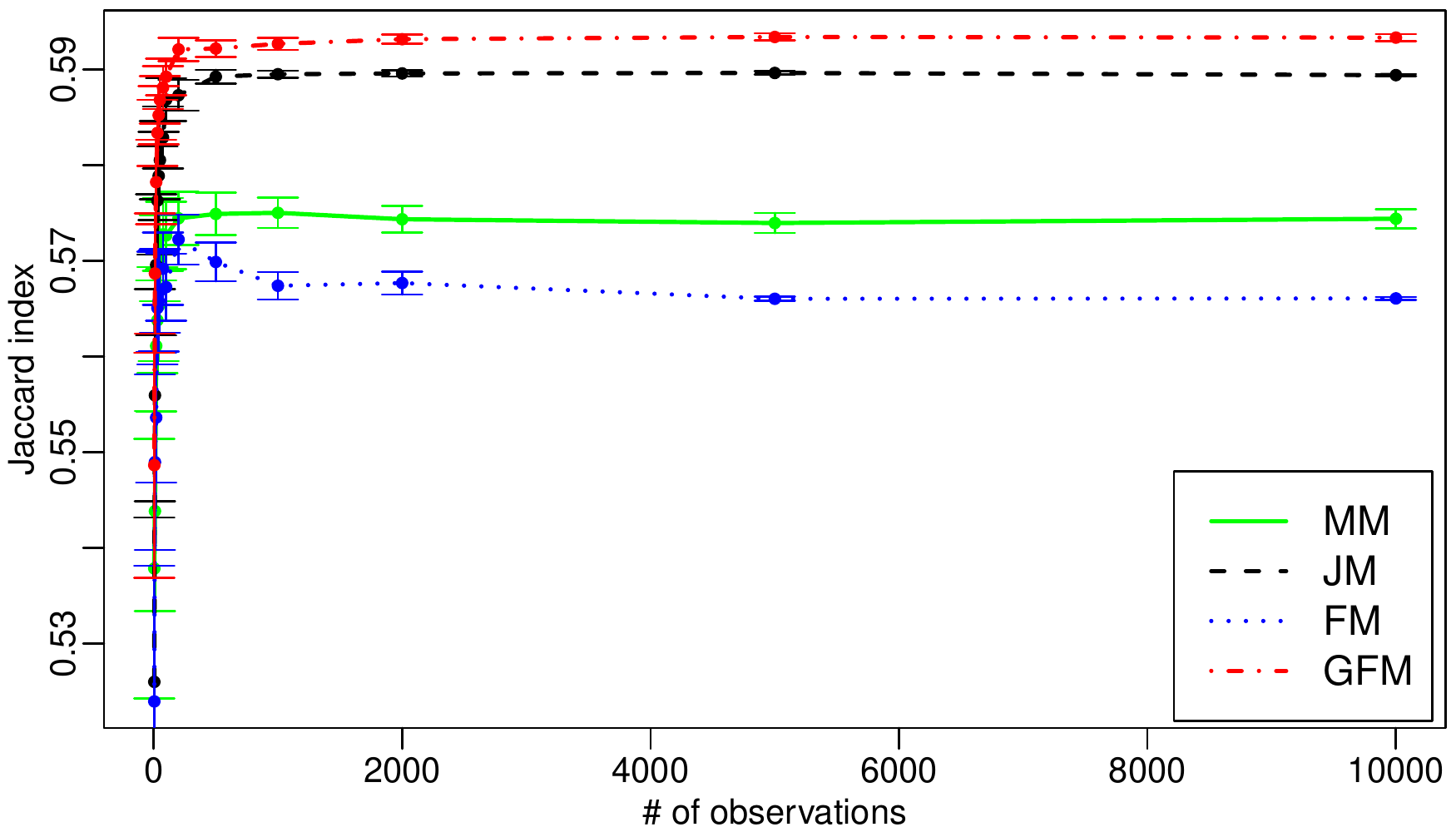}\\
\end{tabular}
\caption{Performance of inference methods in case of label dependence as a function of the number of training observations. Left: the performance up to 100 training observations. Right: the performance up to 10000 training observations. The error bars show the standard error of the measured quantities.}
\label{fig:dependent_data}
\end{figure}

\begin{figure}[ht!]
\begin{tabular}{cc@{}}
\multicolumn{2}{c}{Estimation and prediction time} \\
\includegraphics[width=.45\textwidth]{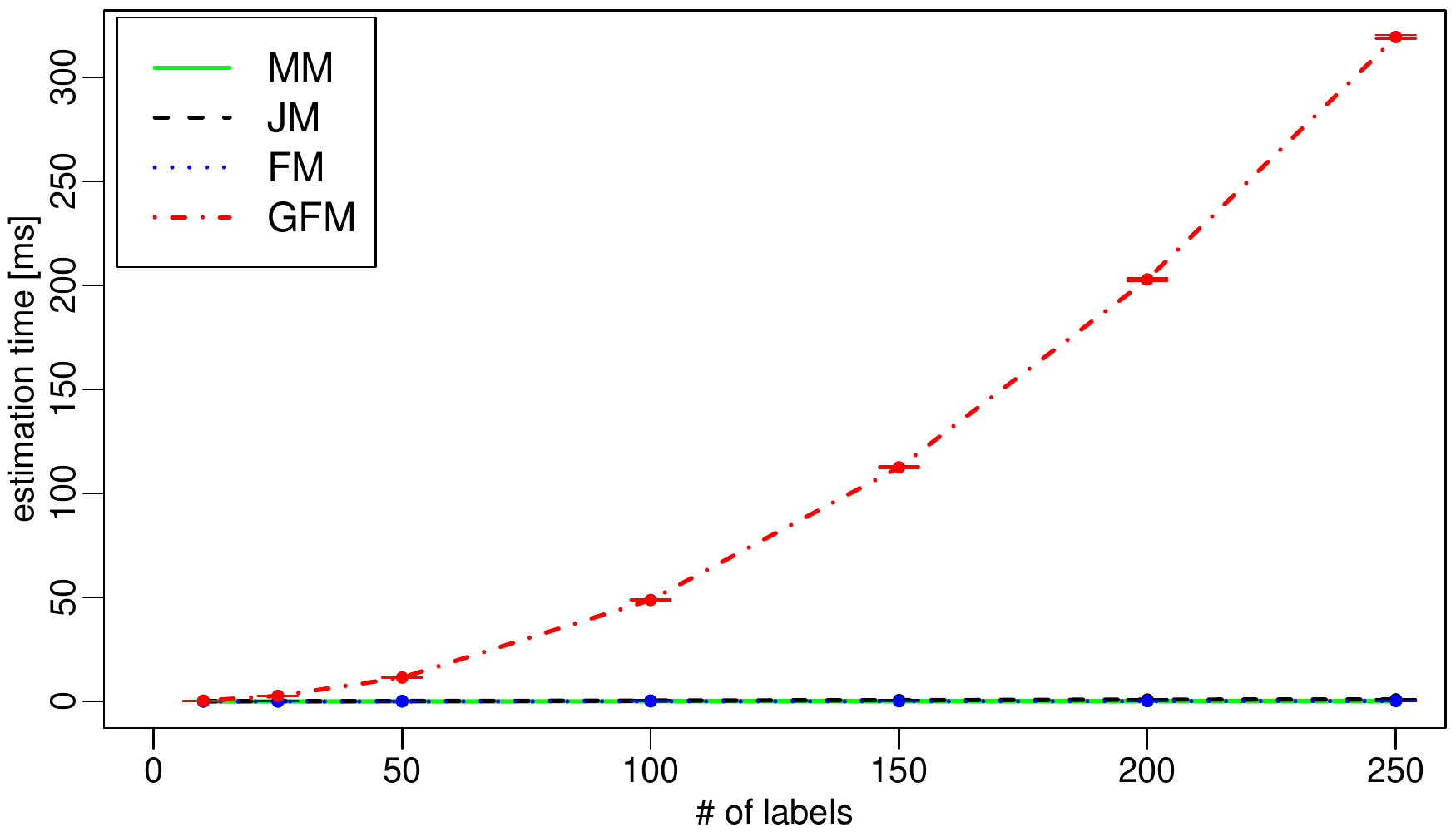} & \includegraphics[width=.45\textwidth]{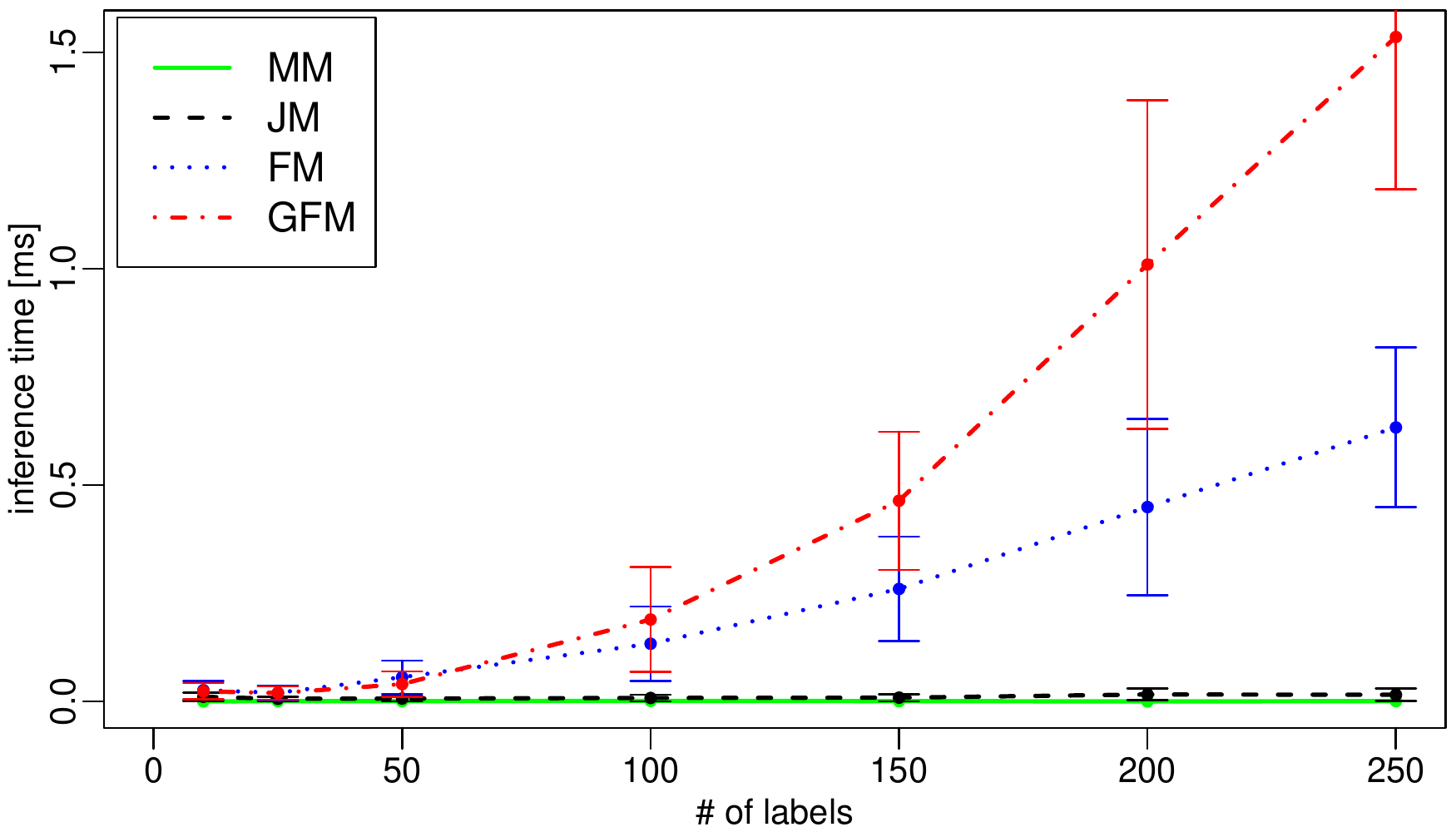}\\
\end{tabular}
\caption{Running times in milliseconds of the inference methods for label dependent data as a function of the number of labels. Left plot shows estimation time of parameters required by the method. Right plot shows the inference time.  The error bars show the standard error of the measured quantities.}
\label{fig:running_times}
\end{figure}

We perform a similar experiment for a data model with strong label dependences. The data are generated by the chain rule of probability, i.e.,
$$
\Pr(\vec{y}) = \prod_{i=1}^m \Pr(y_i \given y_1,\ldots, y_{i-1}),
$$
where the probabilities $\Pr(y_i \given y_1,\ldots, y_{i-1})$ are coming from a logistic model of the form:
$$
\Pr(y_i \given y_1,\ldots, y_{i-1}) = \frac{1}{1 + \exp(- \sum_{j=1}^{i-1} 2 w_{ij} (y_j - \frac{1}{2})  -  w_{i0})},
$$
with all $w_{ij}\sim N(1, 3)$ and $w_{i0} \sim N(1, 3)$. This model tends to produce for a given label $y_i$ a value that appeared more often on previous labels. The results are presented in Figure~\ref{fig:dependent_data}. In this case, the marginal modes and the joint mode are not the same. Therefore MM performs the best for Hamming loss and JM for subset 0/1 loss.  More importantly,  we can see that in this case FM performs suboptimally for F-measure, and a clear winner in this case is GFM. Also GFM performs the best for the Jaccard index, followed by the JM. The results confirm our theoretical analysis and show the benefits of the GFM inference method. Of course, there is a price we have to pay. Figure~\ref{fig:running_times} presents running times of parameter estimation and inference  of the algorithms as a function of the number of labels. GFM is the slowest method. The running times increase quadratically with the number of labels. We also see that the inference time of FM grows quadratically (but with a lower rate), however, this algorithm only needs to estimate marginal probabilities, therefore its estimation time is exactly the same as for the MM method. 

\section{Application to Multi-Label Classification Problems}
\label{sec:application_to_mlc}

The inference methods for F-measure maximization can be used whenever an estimation of required parameters is possible. 
%
%
In this section, we focus on the application of the inference methods in the multi-label setting. Thus, we consider a task of predicting a vector $\vec{y} = (y_1, y_2, \ldots, y_m) \in \{0,1 \}^m$ given another vector $\vec{x} = (x_1, x_2,\ldots,x_n) \in \mathbb{R}^n$ as input attributes. To this end, we use 
a training set $\{(\vec{x}_i,\vec{y}_i)\}_{i=1}^N$ to estimate the required parameters and perform inference for a given test vector $\vec{x}$ so as to deliver an optimal prediction under the F-measure~(\ref{eq:f1}). Thus, we optimize the performance for each instance individually (instance-wise F-measure), in contrast to macro- and micro-averaging of the F-measure.

\subsection{Learning Algorithms}
\label{sec:paradis}

The inference methods for F-measure maximization can be combined with many conventional learning approaches. In the following, we mainly focus on algorithms in which the final decision is made based on an empirical distribution, like in nearest-neighbors or decision trees. 

In these algorithms all parameters required by F-measure maximization methods are estimated from the empirical distribution. 
At the end of this section we also discuss another approach in which the paremeters are obtained from a parametric model, for example, from logistic regression.  

We present the algorithms with the GFM method for F-measure maximization; however, it should be clear from the description how to obtain corresponding variants of the algorithms for the methods that assume label independence. 


\subsubsection{Instance-based Learning}
\label{sec:ibl}

Several instance-based methods for multi-label classification have been proposed in the past, see e.g.\ \citep{zhang07,Cheng_Hullermeier_2009}, but none of these methods is tailored for optimizing the F-measure during an inference phase. 
Consider a query instance $\vec{x} \in \mathcal{X}$ and let $\{\vec{y}_j,  \vec{x}_j\}_{j=1}^l$ denote the $l$ nearest neighbors of $\vec{x}$ with respect to a distance measure on $\mathcal{X}$ in the training set. The number $l$ is a fixed parameter of the method. Instance-based learning can be extended for maximizing the F-measure in a straight-forward way, namely by replacing the distribution $\Pr(\vec{y})$ with the empirical distribution in the neighborhood of the query. Correspondingly, the values $\mx{\Delta}$ and $\Pr(\vec{y} = \vec{0})$ are estimated through simple counting:
$$
\hat{\Delta}^1_{ik} = \frac{1}{l}\sum_{j=1}^l  \frac{\llbracket y_{ik} = 1 \rrbracket}{s_{\vec{y}_j} + k}, \quad  \hat{\Pr}(\vec{y} = \vec{0}) =  \frac{1}{l}\sum_{j=1}^l \llbracket \vec{y}_{j} = \vec{0} \rrbracket.
$$
By using these estimates in the GFM algorithm, we obtain an estimate of the F-measure maximizer. 

\subsubsection{Decision Trees}
\label{sec:decision_trees}


Decision tree methods have been extensively studied in standard classification and regression settings, and have also been generalized to multi-output problems like MLC and multivariate regression, see e.g.\ \citep{zhang98,lee06}. 
An adaptation of decision trees for maximizing the F-measure is slightly more complicated than for instance-based learning. The method we present here resembles the ideas  used in predictive clustering trees~\citep{Vens_et_al_2008}. For simplicity, we only consider binary trees. Moreover, since decision tree induction is well-known in the machine learning field, we restrict our discussion to the main differences to conventional (classification or regression) tree learning. In a decision tree, each leaf node represents a (typically rectangular) part of the instance space $\mathcal{X}$ and is labeled with a local model. Typically, a local model consists of a single prediction, namely the prediction that minimizes the average loss among the associated training examples (e.g., the mean value in regression and the most frequent label in classification).
Applying the same principle in the case of the F-measure in MLC comes down to computing the maximizer of this measure over the instances in a leaf node. This is similar to the case of instance-based learning, except that the examples used for estimating $\mx{\Delta}$ and $\Pr(\vec{y} = \vec{0})$ originate from a rectangular region of the instance space $\mathcal{X}$ and not from the neighborhood of the query instance $\vec{x}$.

The more demanding part is the induction of the tree, i.e., finding optimal splits with respect to the F-measure.
For a given node of the tree, we search over all attributes and possible split points, just like in regular decision tree algorithms. Let us denote by $\mathcal{N}$ a set of training examples $\{(\vec{y}_j,  \vec{x}_j)\}_{j=1}^l$ in a node. The task is then to find a split of $\mathcal{N}$ into two subsets, $\mathcal{N}_L$ and $\mathcal{N}_R$, that maximize some purity criterion. Our approach is analogous to the one of conventional decision trees, but based on the F-measure:
$$
 Q = \frac{\#(\mathcal{N}_L)}{\#(\mathcal{N})} F(\mathcal{N_L}) + \frac{\#(\mathcal{N}_R)}{\#(\mathcal{N})} F(\mathcal{N_R})
$$
where $\#(\mathcal{A})$ is a cardinality of a set $\mathcal{A}$, and
$$
F(\mathcal{A}) = \max_{\vec{h}} \frac{1}{\#(\mathcal{A})} \sum_{\vec{y}_i \in \mathcal{A}} F(\vec{y}_i,\vec{h}).
$$
In order to speed up computations of $F(\cdot)$, we notice that searching a split is usually performed in an example-by-example manner, which means that we can easily update our estimates of $\mx{\Delta}$ and $\Pr(\vec{y} = \vec{0})$. Moreover, assuming that a given training example has a lower number of relevant labels, we do not have to recompute the whole matrix $\mx{\Delta}$, but only update some of its rows and columns. Finally, the search for the top $k$ elements in each column of $\mx{\Delta}$ can be made faster by checking local changes in the current rankings of labels. We repeat the above step recursively until a stop condition is reached, for example the F-measure becomes maximal or the number of examples in the leaf node falls below a threshold. Of course, more sophisticated approaches are conceivable.

Let us also mention that it is possible to generalize bagging~\citep{Breiman1996} with these decision trees. Then, GFM can easily be applied over the bootstrap sample of the weak hypotheses returned by the trees. 

\subsubsection{Probabilistic Classifier Chains}

Probabilistic classifier chains (PCCs)~\citep{Dembczynski_et_al_2010a} is an approach similar to maximum entropy Markov models~\citep{McCallum_et_al_2000} and to conditional random fields~(CRFs)~\citep{Lafferty_et_al_2001,Ghamrawi_McCallum_2005}. All these approaches estimate the joint conditional distribution $\Pr(\vec{y} \given  \vec{x})$. PCC has an additional advantage that one can easily sample from the estimated distribution. 
The underlying idea is to repeatedly apply the product rule of probability to the joint distribution of the labels:
\begin{equation}
\label{eqn:prp}
\Pr(\vec{y} \given \vec{x}) = \prod_{i=1}^{m} \Pr(y_i \given \vec{x}, y_1, \ldots,y_{i-1})
\end{equation}

Learning in this framework can be considered as a simple procedure. According to (\ref{eqn:prp}), we decompose the joint distribution into a sequence of marginal distributions that depend on a subset of the labels. These marginal distributions can be learned by $m$ functions 
$f_i : \,   \mathcal{X} \times \{0,1\}^{i-1} \rightarrow  [ 0, 1 ] $
on an augmented input space $\mathcal{X} \times \{0,1\}^{i-1}$, taking $y_1, \ldots , y_{i-1}$ as additional input attributes:
\begin{equation}
f_i: \,   
 (\vec{x}, y_1, \ldots , y_{i-1})  \mapsto  \Pr(y_i = 1 \given \vec{x}, y_1, \ldots , y_{i-1})
\end{equation}
By plugging the log-linear model into (\ref{eqn:prp}), it can be shown that pairwise dependencies between labels $y_i$ and $y_j$ are modeled (see also \citep{Kumar2013}). 

The algorithm is mainly suitable for subset 0/1 loss. Exploration of the structure of the chain in the inference phase boils down to search for the most probable label combination in a resulting probabilistic binary tree. A greedy algorithm follows only one path choosing always only the most probable label in each position in the chain~\citep{Read_et_al_2009}. This algorithm, however, may lead to suboptimal results. It has been shown \citep{Demb2012b}, however, that an exact method based on a variant of uniform-cost search with a cut-off list finds the joint mode in a linear time of $1/p_{max}$, where $p_{max}$ is the probability of the joint mode. For reasonable values of $p_{max}$, this method works very fast.

To optimize the response of PCC for other loss functions, we need to obtain a sample of observations from the conditional joint distribution $\Pr(\vec{y} \given  \vec{x})$. To get a single observation, we can follow the chain and pick the value of label $y_i$ by tossing a biased coin with probabilities given by the $i$-th classifier. Such a procedure is sometimes referred to as ancestral sampling \citep[Chapter 8]{Bishop_2006}. From the sample of such observations, we can estimate all the parameters required by inference methods like GFM, similarly as in the case of nearest neighbors and decision trees. More precisely, let  $\{\vec{y}_j\}_{j=1}^n$ denote a set of sampled observations for a given test example $\bx$. Then, the values $\mx{\Delta}$ and $\Pr(\vec{y} = \vec{0})$ can be estimated through simple counting:
$$
\hat{\Delta}^1_{ik} = \frac{1}{n}\sum_{j=1}^n  \frac{\llbracket y_{ik} = 1 \rrbracket}{s_{\vec{y}_j} + k}, \quad  \hat{\Pr}(\vec{y} = \vec{0}) =  \frac{1}{n}\sum_{j=1}^n \llbracket \vec{y}_{j} = \vec{0} \rrbracket.
$$
By plugging these estimates into the GFM algorithm, we obtain for a given $\bx$ a prediction  optimized for the F-measure.

\subsubsection{Parametric Models}
\label{sec:parametric_models}

Alternatively to the approaches described above, which estimate the parameters required by the F-measure maximization methods on empirical distributions, we discuss here an approach in which the parameters are efficiently obtained from a parametric model~\citep{Dembczynski_et_al_2013}.  Unfortunately, there is no easy way to estimate directly the matrix $\Delta$, since the elements of this matrix do not correspond to a proper probability distribution, i.e., $\Delta^0_{ik}$ and $\Delta^1_{ik}$ do not sum up to 1 or to any known constant; moreover, there is also no other subset of the elements of $\Delta$ that may posses a property of that kind. However, we can estimate matrix $\mx{P}$, defined in (\ref{eq:matrix_P}), which elements are probabilities:
$$
p_{is} = \Pr(y_i = 1, s_{\vec{y}} = s), \quad i,s \in \{1,\ldots,m\} \,.
$$
 Multiplying the matrix $\mx{P}$ by a weight matrix $\mx{W}$ with elements (\ref{eq:elements_of_W}) results in the estimate of $\Delta$, as shown in (\ref{eqn:matrix_multiplication}). 

To estimate the elements of $\mx{P}$ we can use a simple reduction to $m$ multi-class probability estimation problems (e.g., multinomial regression) with at most $m+1$ classes. We obtain one multinomial regression model for each row of matrix $\mx{P}$. Since 
$$
\sum_{y \in \{0, \ldots, m\}}  \Pr(y = \assert{y_i = 1} \cdot s_{\by} \given \bx) = 1,
$$
we can define the $i$-th problem as:
\begin{equation}
f_i: \,   \vec{x} \mapsto \Pr(y = \assert{y_i = 1} \cdot s_{\by} \given \bx), \mathrm{~for~} y \in \{0, \ldots, m\}.
 \end{equation}
%

In a similar way, we can estimate $\Pr(\by = \vec{0} \given \bx)$ by performing an additional reduction to binary probability estimation:
$$
f_0: \bx \mapsto \Pr(y = \assert{\by = \vec{0}} \given \bx ) \,,
$$
and solving it via logistic regression.

The decomposition of the original problem into independent multinomial regression tasks has computational advantages. Moreover, since the number of distinct values of $s_{\by}$ is usually small, the number of classes in a single multinomial regression task is much smaller than $m+1$; only in the worst case, we end up with a quadratic complexity in the number of labels $m$. 

Let us, however, remark that the elements of matrix $\mx{P}$ estimated across different tasks are not fully independent of each other (e.g., $p_{im}$ is the same for all $i$). Consequently, learning on a finite training set may lead to conflicting estimates that are not in agreement with any valid distribution. To avoid such conflicts, one may include additional constraints in the learning problem or calibrate the estimates afterwards. However, \citet{Dembczynski_et_al_2013} have shown that with the sample size growing to infinity this approach is statistically consistent.

To summarize this approach we notice that learning of the probabilistic model has a time complexity that is at most quadratic in $m$. In the inference phase for a test instance $\bx$, we first get estimates of $\Pr(\vec{0} \given \bx)$ and $\mx{P}$ from the probabilistic model, again in at most quadratic time. Then, we need to multiply matrices $\mx{P}$ and $\mx{W}$ to get $\mx{\Delta}$. Finally, all the parameters are plugged into the GFM method. 
This approach has in the original paper~\citep{Dembczynski_et_al_2013} been referred to as Exact-F-measure-Plug-in classifier (EFP).

Under the assumption of label independence, one can simplify this approach. Since we only need marginal probabilities $p_i$, it is enough to reduce the problem to $m$ binary probability estimation tasks that can be solved, for example, by logistic regression. Then, for each test instance $\bx$, we obtain a vector of marginal probabilities $p_i$, to which one might apply, for example, the inference method of \citet{Ye2012}. We refer to this approach as Label-independence F-measure-Plug-in classifier (LFP), similarly as in~\citep{Dembczynski_et_al_2013}.


%

\subsection{Experimental Results}

We test some of the algorithms described above on four commonly used multi-label benchmark datasets with known training and test sets. We take these datasets from the MULAN\footnote{\url{http://mulan.sourceforge.net/datasets.html}} and LibSVM\footnote{\url{http://www.csie.ntu.edu.tw/~cjlin/libsvmtools/datasets/multilabel.html}} repositories. Table~\ref{tbl:datasets} contains basic statistics of these datasets.
We also relate the obtained results to the results of a variant of structured SVMs that moves the effort of maximizing the F-measure to the training phase.

We run the experiments on the machine that was also used for the simulations described earlier, i.e., on a Debian virtual machine with 8-core x64 processor and 5GB RAM.

\begin{table}[h]
\caption{Datasets and their properties. The number of training and test  observations is denoted by \#train and \#test, respectively, $m$ is the number of labels, $d$ is the number of features.}
\label{tbl:datasets}
\begin{center}
\begin{sc}
\begin{tabular}{lrrrr}
\toprule 
Dataset & \#train & \#test & $m$ & $d$  \\
\midrule
Scene       & 1211  & 1196  & 6   & 294 \\
Yeast       & 1500  & 917   & 14  & 103 \\
Enron       & 1123  & 579   & 53  & 1001 \\
Mediamill   & 30993 & 12914 & 101 & 120 \\
\bottomrule 
\end{tabular}
\end{sc}
\end{center}
\end{table}

\subsubsection{Instance-based Learning}
We first present results of instance-based learning.  We use a different number of nearest neighbors, $l \in \{10, 20, 50, 100\}$. For each test example, we seek for its nearest neighbors and apply different inference methods. We use exactly the same methods we applied in our experiments on synthetic data given in Section~\ref{sec:simulations}. The first method, MM, estimates the marginal modes, JM estimates the joint mode, FM approximates the F-measure by assuming label independence, and the introduced GFM performs exact inference for the F-measure. The nearest-neighbor search is performed by using the Weka~\citep{Hall_et_al_2009} and Mulan~\citep{Tsoumakas_et_al_2011} implementation of instance-based learning.

The results are given in Table~\ref{tbl:jnn-results}. Similarly as in Section~\ref{sec:simulations}, we report the performance in terms of Hamming loss, subset 0/1 loss, F-measure and Jaccard index.
\begin{table}
\caption{Empirical results on 4 benchmark datasets. Instance-based methods are used with a different number of neighbors ($l \in \{10, 20, 50, 100\}$) and with different inference methods: MM -- estimates marginal modes, JM -- the joint mode, FM -- approximates the F-measure maximizer by assuming independence of labels, and GFM -- computes the exact F-measure maximizer over the nearest neighbors. Results are reported for Hamming loss, subset 0/1 loss, F-measure and Jaccard index. The best results for a given number of neighbors and a performance measure are marked in bold.}
\label{tbl:jnn-results}
\begin{center}
{\scriptsize
\begin{sc}
\begin{tabular}{@{}l@{~~}r@{~}r@{~}r@{~}r@{~~}r@{~}r@{~}r@{~}r@{~~}r@{~}r@{~}r@{~}r@{~~}r@{~}r@{~}r@{~}r@{}}
\toprule 
          & \multicolumn{4}{c}{Hamming Loss [\%]} & \multicolumn{4}{c}{Subset 0/1 Loss [\%]} & \multicolumn{4}{c}{F-measure [\%]} & \multicolumn{4}{c}{Jaccard [\%]} \\
    & $10$ & $20$ & $50$ & $100$ & $10$ & $20$ & $50$ & $100$ & $10$ & $20$ & $50$ & $100$ & $10$ & $20$ & $50$ & $100$\\
\midrule
          & \multicolumn{16}{c}{Scene} \\
\midrule
MM  &\bf 10.28 &\bf 10.62 &\bf 11.52 &\bf 13.03 &    40.47 &    45.15 &     54.01 &    66.05 &    65.80 &    59.53 &    49.39 &    36.40 &    64.23 &    58.36 &    48.54 &    35.79\\
JM  &    11.19 &    10.99 &    11.97 &    12.28 &\bf 36.54 &\bf 36.29 & \bf 39.13 &\bf 40.05 &    68.26 &    68.73 &    65.89 &\bf 64.97 &\bf 67.06 &\bf 67.47 &\bf 64.63 &\bf 63.71\\
FM  &    13.80 &    14.40 &    16.74 &    19.83 &    53.09 &    56.10 &     64.55 &    75.17 &    70.80 &    70.12 &    67.20 &    62.65 &    64.59 &    63.24 &    58.76 &    52.48\\
GFM &    13.03 &    14.09 &    16.15 &    19.12 &    49.75 &    54.68 &     63.13 &    72.91 &\bf 71.42 &\bf 70.29 &\bf 67.85 &    64.31 &    65.95 &	   63.80 &    59.73 &    54.41\\
\midrule
          & \multicolumn{16}{c}{Yeast} \\
\midrule
MM  &\bf 20.72 &\bf 20.00 &\bf 20.02 &\bf 20.38 &    81.57 &    81.03 &    82.77 &    85.71 &    62.88 &    61.93 &    60.31 &    58.26 &    52.31 &    51.49 &    49.74 &    47.47\\
JM  &    23.09 &    21.78 &    21.84 &    22.04 &\bf 76.66 &\bf 76.23 &\bf 77.54 &\bf 78.30 &    59.14 &    60.53 &    60.97 &    60.68 &    49.75 &    51.06 &    51.20 &    50.72\\
FM  &    22.94 &    23.26 &    22.83 &    23.21 &    83.21 &    83.21 &    85.93 &    88.11 &    65.29 &    65.06 &    65.23 &    64.85 &    54.14 &    53.74 &    53.71 &\bf 53.10\\
GFM &    22.94 &    23.17 &    22.87 &    23.61 &    82.99 &    83.97 &    86.37 &    88.66 &\bf 65.49 &\bf 65.47 &\bf 65.75 &\bf 64.98 &\bf 54.31 &\bf 54.09 &\bf 54.11 &    53.08\\
\midrule
          & \multicolumn{16}{c}{Enron} \\
\midrule
MM  &\bf  5.73 &\bf 5.94 &\bf  6.28 &\bf   6.46 &    88.08 &    88.60 &    89.46 &    89.64 &    35.56 &    27.91 &    23.23 &    23.71 &    28.70 &    22.97 &    19.36 &    19.60\\
JM  &     6.56 &    6.51 &     6.70 &      6.73 &\bf 86.18 &    87.39 &    89.29 &    89.46 &    33.38 &    29.52 &    25.52 &    24.68 &    27.68 &    24.51 &    21.05 &    20.38\\
FM  &     6.51 &    6.23 &     6.34 &      6.48 &    87.56 &    88.60 &\bf 87.22 &\bf 88.26 &\bf 47.50 &\bf 47.01 &\bf 42.10 &\bf 37.86 &\bf 37.31 &\bf 36.82 &\bf 33.65 &\bf 30.17\\
GFM &     6.61 &    6.28 &     6.54 &      6.63 &    88.08 &\bf 86.87 &    88.43 &    89.29 &    44.43 &    42.45 &    34.76 &    29.27 &    35.12 &    34.10 &    27.93 &    23.70\\
\midrule
          & \multicolumn{16}{c}{Mediamill} \\
\midrule
MM  &\bf  3.29 &\bf 3.18 &\bf  3.16 &\bf   3.19 &    89.17 &    89.66 &    90.30 &    91.13 &    54.98 &    53.92 &    52.68 &    51.81 &    43.32 &    43.34 &    43.10 &    42.74\\
JM  &     4.17 &    4.03 &     3.93 &      3.91 &\bf 88.93 &\bf 88.82 &\bf 89.24 &\bf 89.05 &    48.87 &    48.58 &    47.32 &    47.07 &    43.55 &\bf 43.66 &\bf 43.30 &\bf 42.80\\
FM  &     3.97 &    3.81 &     3.72 &      3.68 &    91.30 &    91.99 &    92.64 &    93.39 &    55.47 &    55.65 &    55.51 &\bf 55.24 &\bf 43.60 &    42.64 &    41.49 &    40.58\\
GFM &     3.87 &    3.74 &     3.66 &      3.65 &    90.48 &    91.15 &    91.99 &    92.81 &\bf 55.52 &\bf 55.80 &\bf 55.54 &    55.16 &    38.56 &    38.40 &    37.39 &    37.28\\
\bottomrule 
\end{tabular}
\end{sc}
}
\end{center}
\end{table}

We can generally confirm our previous results on synthetic data: an inference method tailored for a given performance measure obtains the best results. This is clear for Hamming loss, for which MM has the smallest error throughout. JM performs the best for subset 0/1 loss on all datasets with some exceptions on \textsc{Enron}. Both methods tailored for F-measure maximization, FM and GFM, substantially outperform MM and JM on this performance criterion. FM seems to beat GFM on \textsc{Enron}, while  the latter method gets better results on the other datasets. There are, however, no clear results for the Jaccard index. 

Let us underline that in the case of nearest neighbor methods, we are dealing with a specific trade-off between the size of the neighborhood and its volume. In general, increasing the sample size in the inference methods should improve the results. But in this case, by increasing the number of neighbors, we simultaneously increase the volume of the space that contains the neighbors. In other words, some of the neighbors can be far away from the test example, which usually deteriorates the quality of the estimates. This is usually the case for high-dimensional problems. This may partially explain the results on the \textsc{Enron dataset. The performance under the F-measure decreases substantially with the number of neighbors. Also when comparing the instance-based methods with other methods, presented later in this section, we see that the overall performance on this data set is much worse for the former methods. }

In Table~\ref{tbl:jnn-times} we present the computation time of the instance-based methods, including both searching and inference time. We can easily observe that the searching time dominates inference time and there is only a small difference between the inference methods for a given dataset and $l$. We do not present the inference times separately here, since their characteristics are exactly the same as presented in Section~\ref{sec:simulations}. 
\begin{table}
\caption{Computation time in seconds for the instance-based methods for different $l = \{10, 20, 50, 100\}$ and inference methods: MM, JM, FM, and GFM. Computation time includes searching time and inference time}
\label{tbl:jnn-times}
\begin{center}
\begin{sc}
{\scriptsize
\begin{tabular}{l r r r r r r r r r}
\toprule
& \multicolumn{4}{c}{Inference time [s]} &
& \multicolumn{4}{c}{Inference time [s]} \\
  & $10$ & $20$ & $50$ & $100$  & &  $10$ & $20$ & $50$ & $100$\\
\midrule
 & \multicolumn{4}{c}{Scene} &
 & \multicolumn{4}{c}{Yeast} \\
\midrule
MM & 7.622	& 8.110 &	 9.078 &	10.064 &
MM & 3.662	& 3.907 &	 4.352 &	4.793 \\
JM   & 7.519	& 8.052 &	 9.093 &	10.010 &
JM &  3.687	& 3.958 & 4.362 &	4.812 \\
FM  & 7.488	& 8.023 & 9.106 &	10.126 &
FM & 3.728	& 3.920 & 4.404 &	4.788 \\
GFM & 8.311	& 8.171 &	 9.169 &	10.475 &
GFM & 3.634	& 3.944 &	 4.467 &	 4.875 \\
\midrule
 & \multicolumn{4}{c}{Enron} &
 & \multicolumn{4}{c}{Mediamill} \\ 
\midrule
MM & 1.601	& 1.618 &	 1.978 &	2.172 &
MM & 336.550	 & 374.207 & 435.171 &	499.977 \\ 
JM & 1.477	& 1.607 &	 1.929 &	2.236 &
JM & 338.855 	 & 373.297 & 431.354 &	492.079 \\
FM & 1.483	& 1.645 &	 1.931 &	2.145 &
FM & 339.704	 & 375.230 & 433.660 &	493.424 \\
GFM & 1.518	& 1.681 &	 2.022 &	2.228 &
GFM & 347.402 & 382.663 & 442.356 &	502.060 \\
\bottomrule
\end{tabular}
}
\end{sc}
\end{center}
\end{table}

\subsubsection{Probabilistic Classifier Chains}

In the next experiments, we use PCC. We train PCC by using linear regularized logistic regression. We use the implementation of logistic regression from Mallet~\citep{McCallum_2002}. 
We tune the regularization parameter for each base classifier independently by minimizing the logistic loss, hoping to thereby produce better probability estimates. We use 5-fold cross-validation and choose the regularization parameter from the following set of
possible values $\{10^{-4}, 10^{-3}, \ldots, 10^3\}$. Similarly as in the previous experiments, we use four different inference mechanisms. For MM, FM, and GFM methods, we obtain the estimates of the required parameters by performing ancestral sampling from the conditional joint distribution of each test example $\vec{x}$. The JM method, instead of estimating the joint mode from the sample, applies the efficient  exact search method~\citep{Demb2012b}. 

\begin{figure}
\begin{center}
{\scriptsize
\begin{sc}
\begin{tabular}{@{}cc@{}}
\multicolumn{2}{c}{Scene} \\
\includegraphics[width=.45\textwidth]{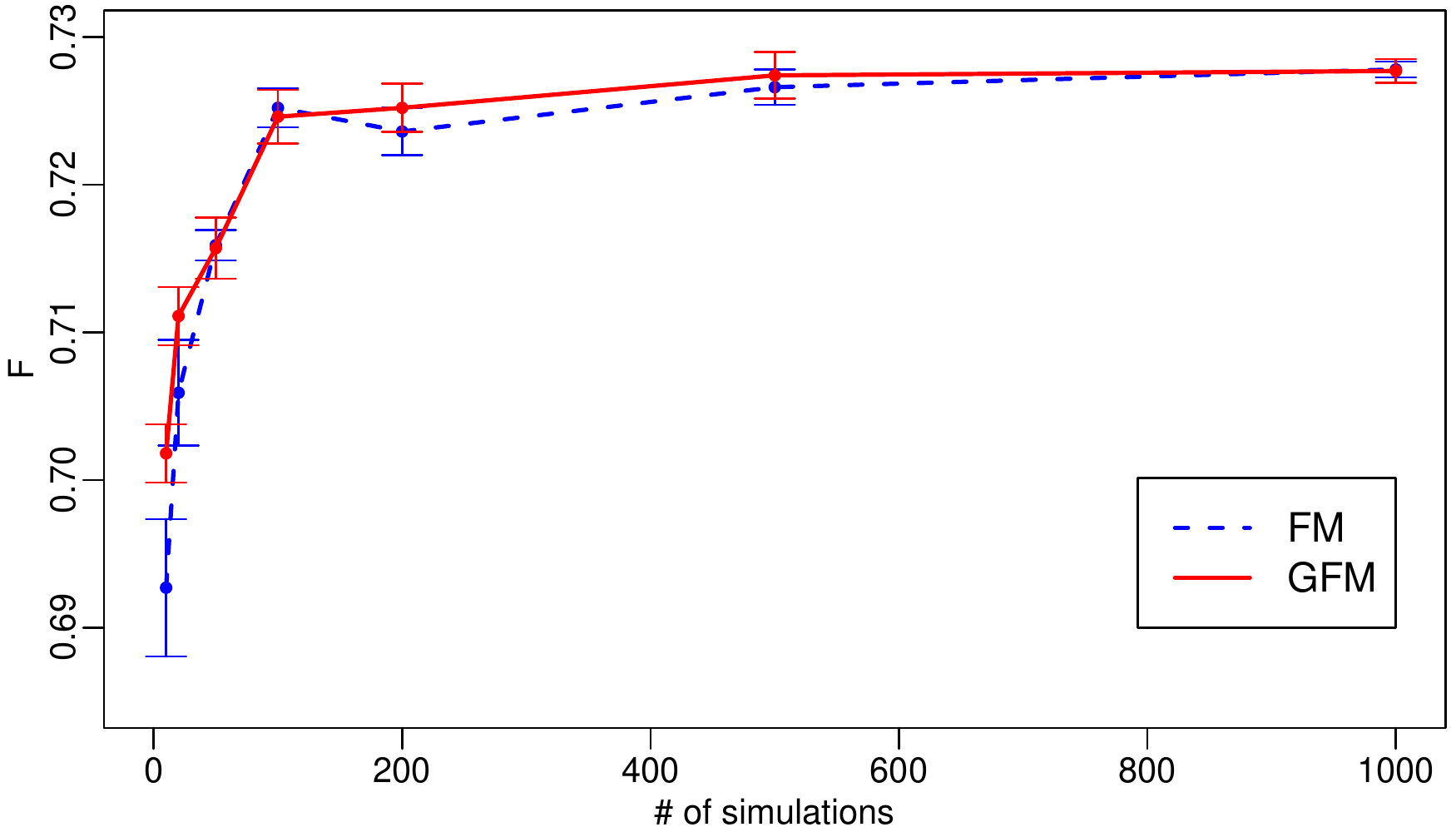} & \includegraphics[width=.45\textwidth]{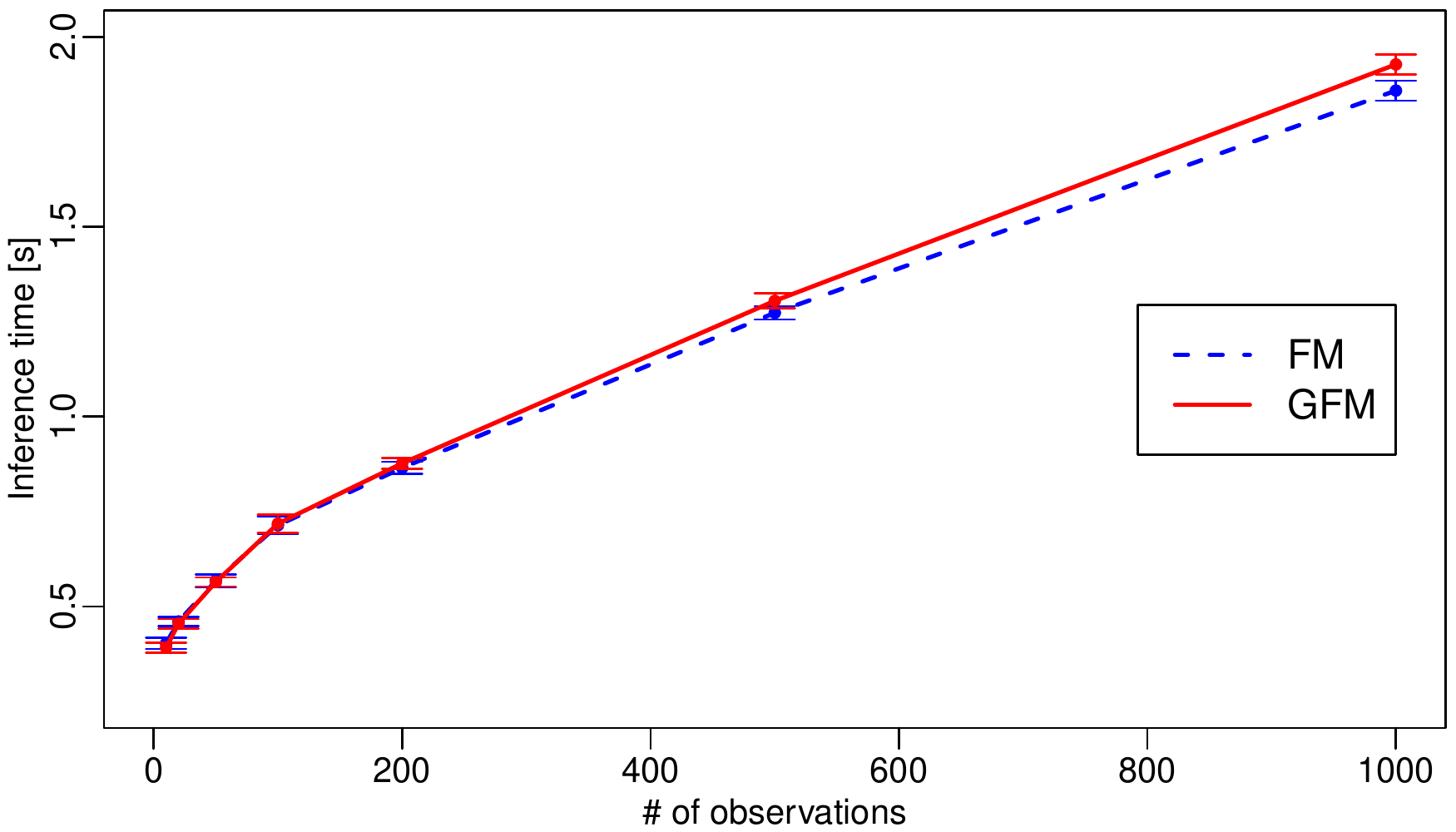}\\
\multicolumn{2}{c}{Yeast} \\
\includegraphics[width=.45\textwidth]{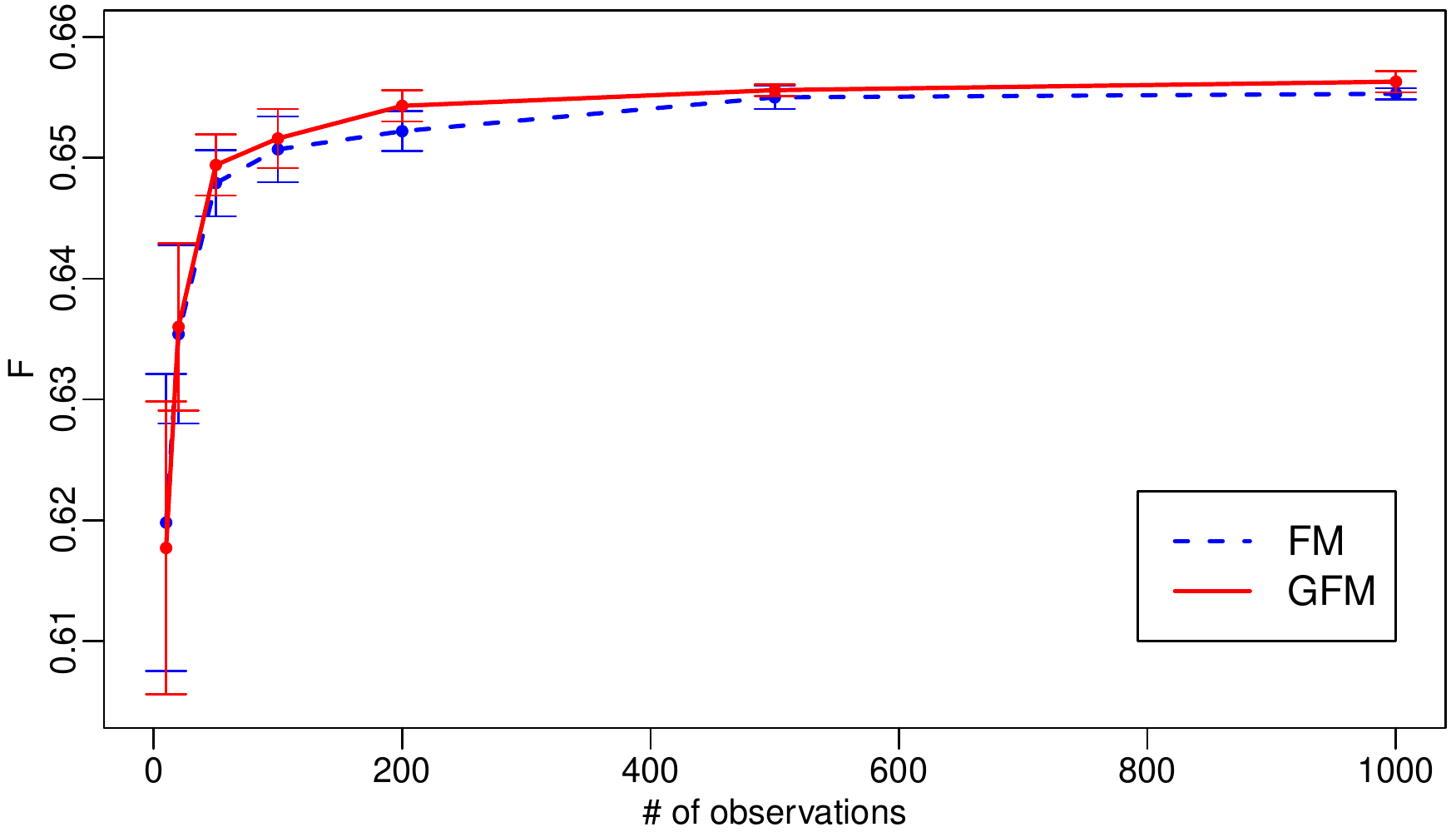} & \includegraphics[width=.45\textwidth]{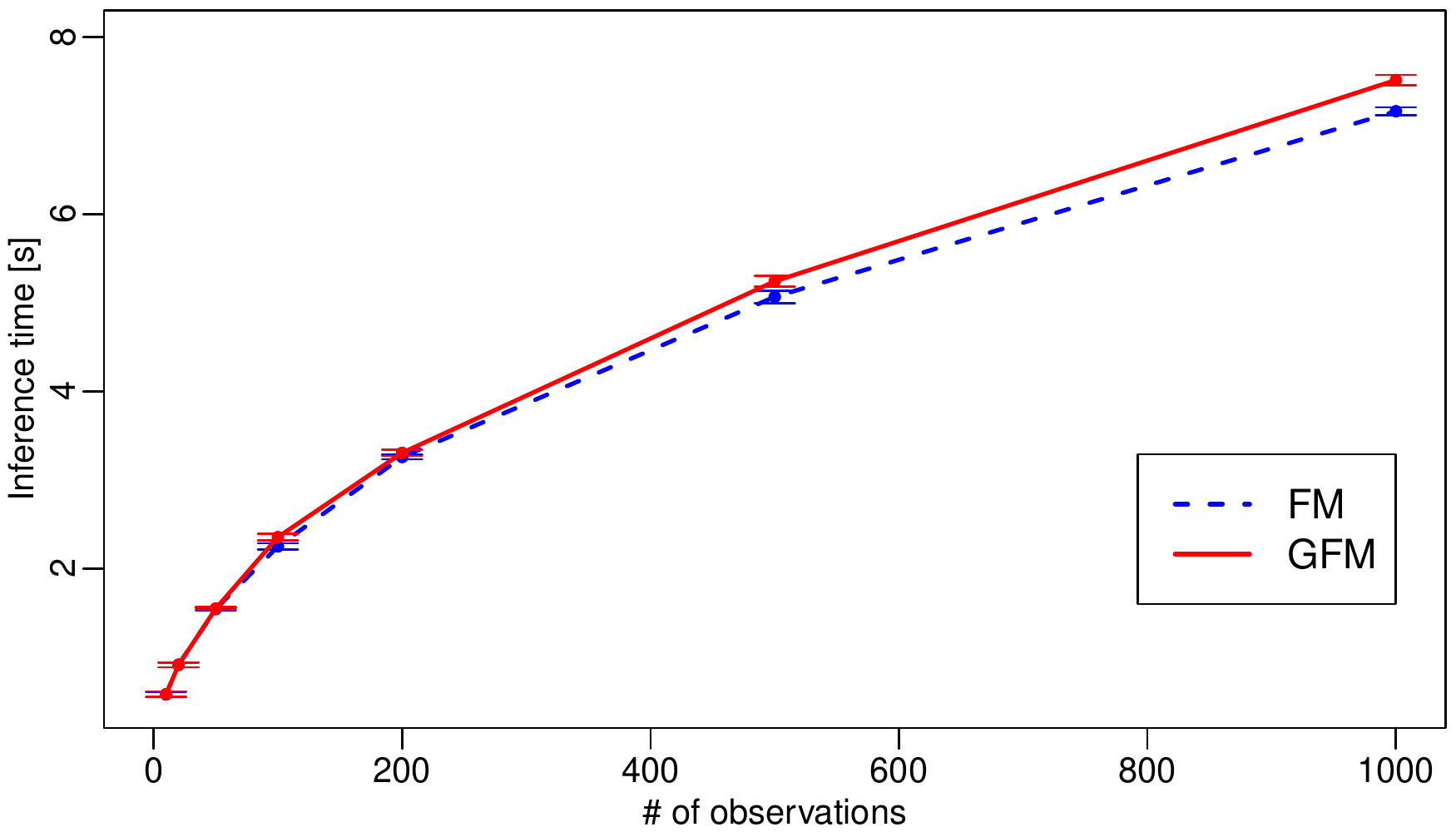}\\
\multicolumn{2}{c}{Enron} \\
\includegraphics[width=.45\textwidth]{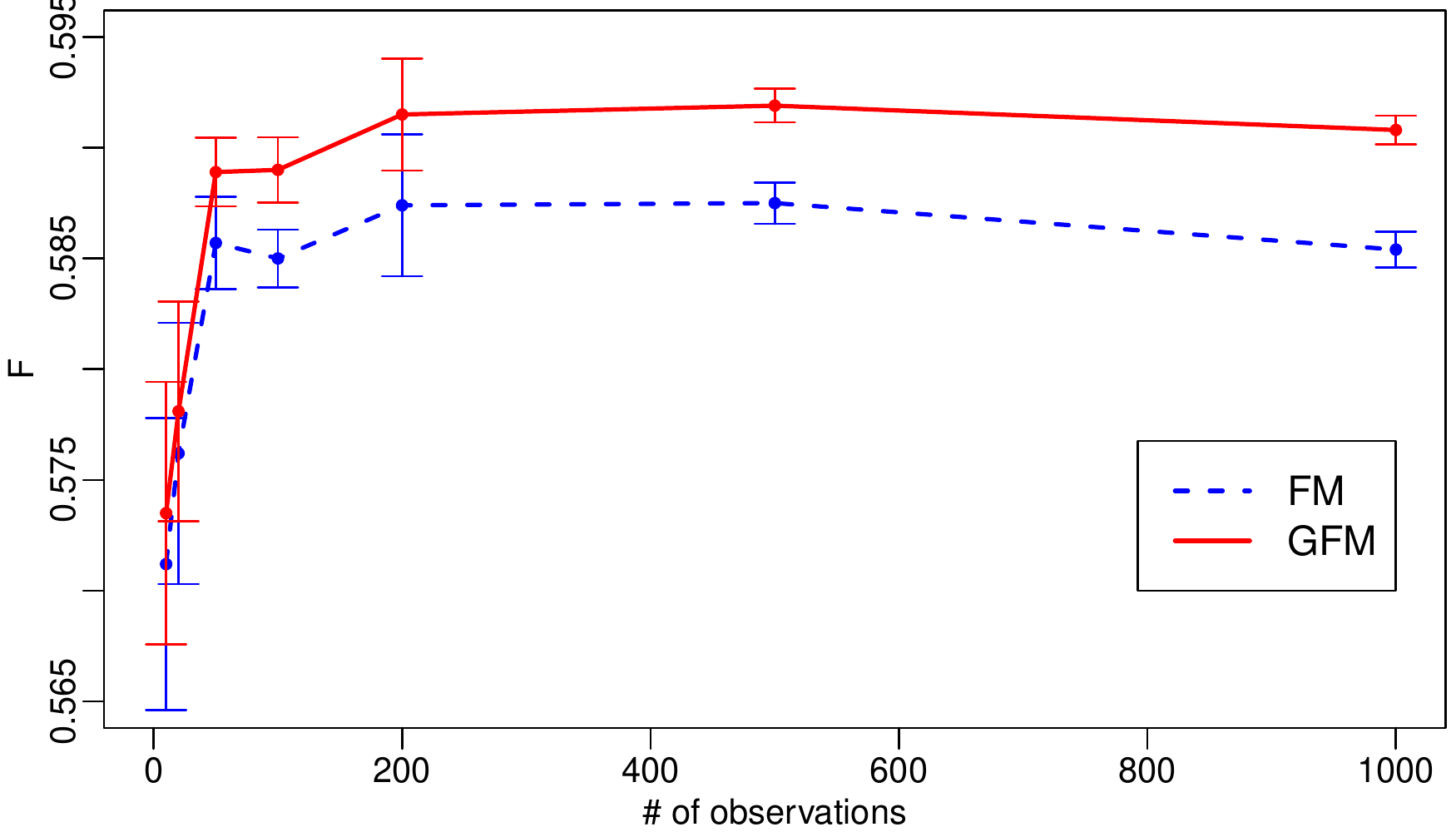} & \includegraphics[width=.45\textwidth]{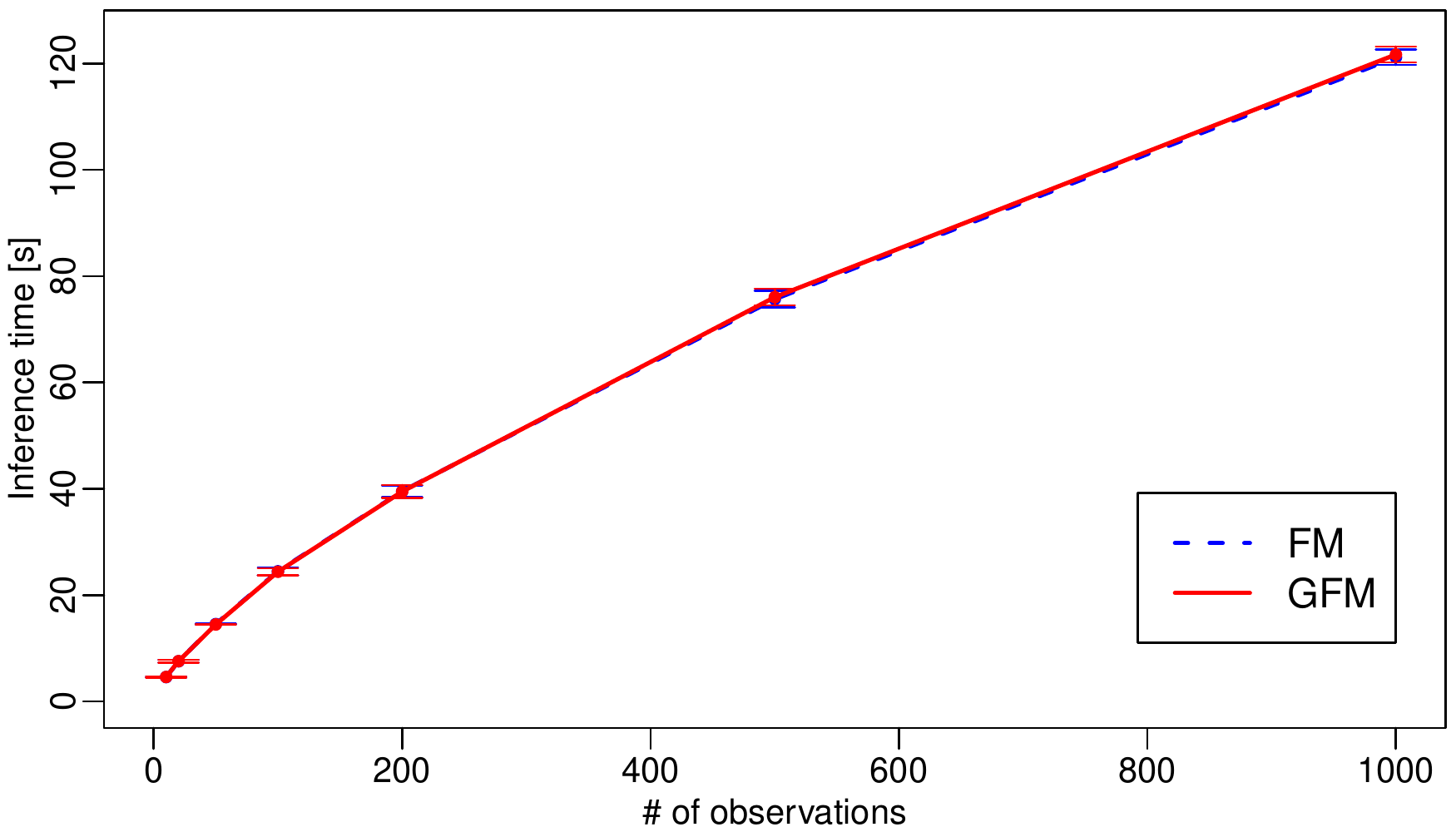}\\
\multicolumn{2}{c}{Mediamill} \\
\includegraphics[width=.45\textwidth]{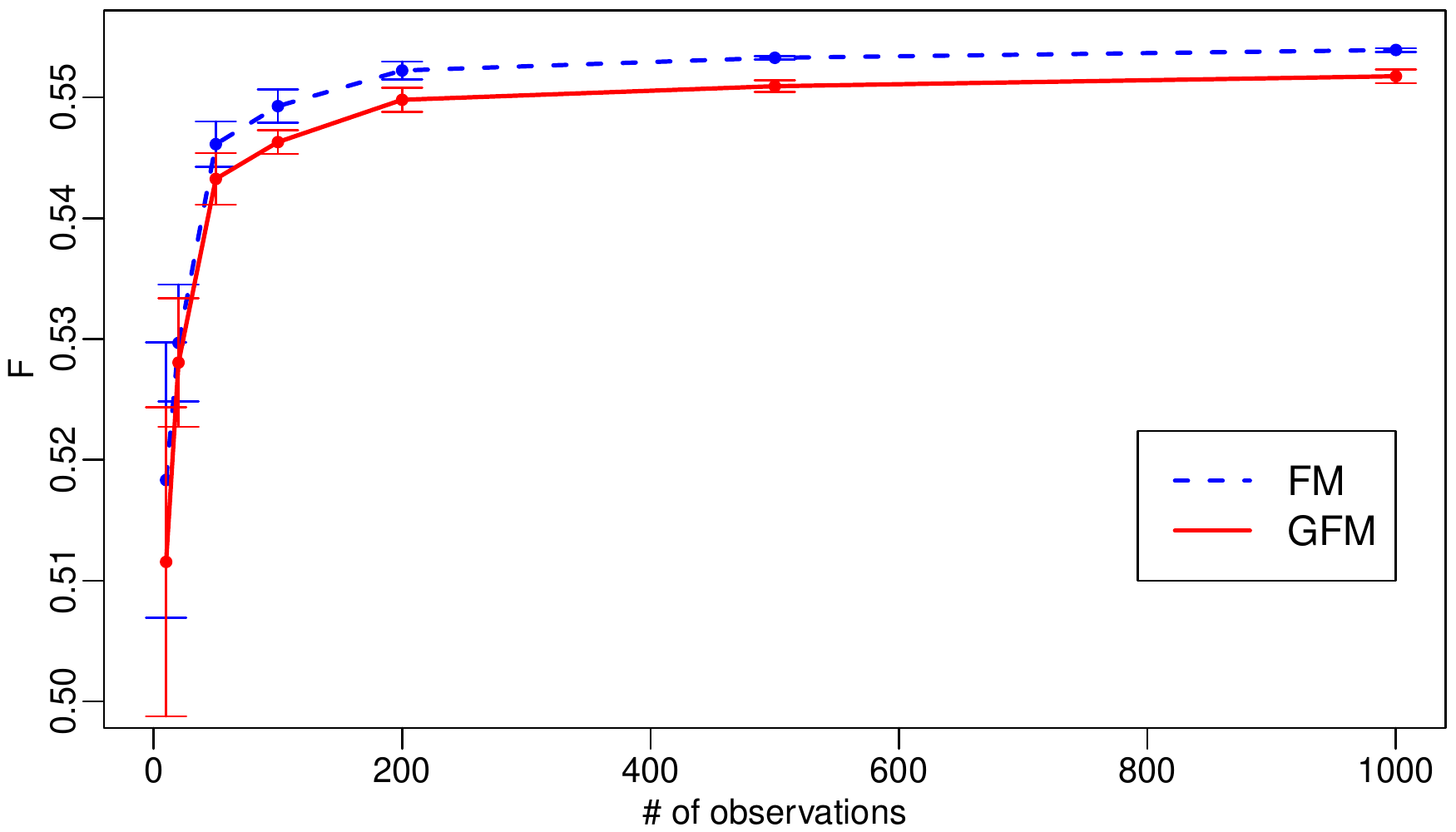} & \includegraphics[width=.45\textwidth]{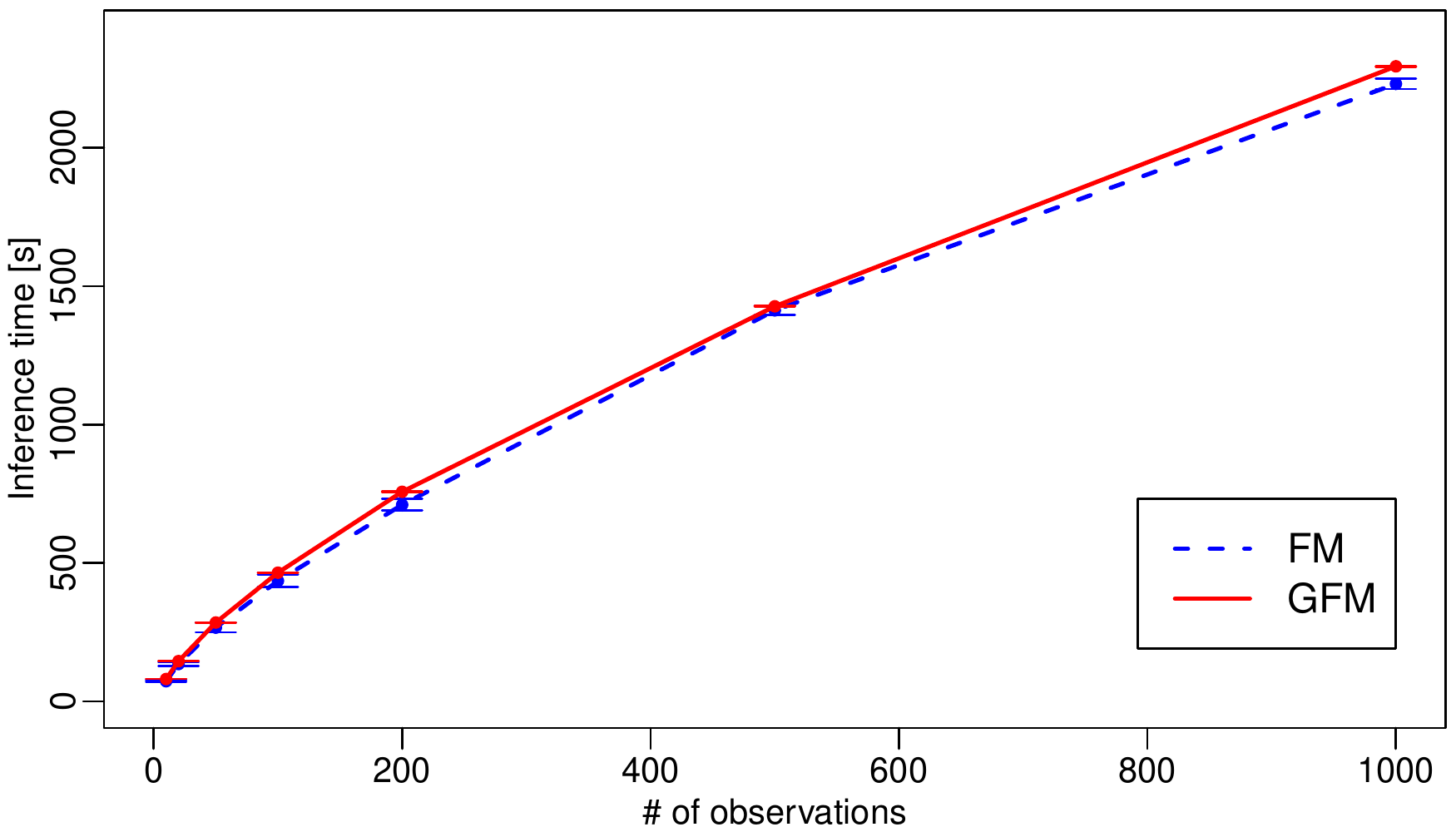}\\
\end{tabular}
\end{sc}
}
\end{center}
\caption{Performance of FM and GFM inference methods used in PCC with different number of observations obtained from ancestral sampling. The results are averaged over 5 runs of the inference methods to eliminate the randomness of sampling. Left plots show F-measure, while right plots inference times. The inference time includes sampling, estimation of parameters, and inference based on these parameters. The error bars show the standard error of the measured quantities.} 
\label{fig:pcc}
\end{figure}

Table~\ref{tbl:pcc-results} contains the results of the experiment. As before, we report the Hamming loss, subset 0/1 loss, F-measure and the Jaccard index. We also give the training and inference times. The training time concerns the entire procedure that consists of tuning of the regularization parameter in cross-validation and training of a model with the best value of the regularization parameter. The results for MM, FM, and GFM are given for sample size of 1000. From the results, we can clearly state that approaches tailored for the F-measure obtain better results on this performance criterion. It seems that GFM obtains slightly better results, but also needs a little bit more time. In Figure~\ref{fig:pcc}, we additionally present the F-measure and inference times of FM and GFM as a function of the number of observations obtained from ancestral sampling. These results are computed over 5 runs of the inference methods to decrease the impact of the randomness of the sampling method. The plots confirm our theoretical results concerning the predictive performance and time complexity. However, the inference times reported here include all three steps: sampling, estimation of parameters, and inference based on these parameters. As we can see in Table~\ref{tbl:pcc-results} and  Figure~\ref{fig:pcc} the differences between GFM, FM, and MM are not substantial here, since sampling is the most expensive step. The exact method used for joint mode estimation works much faster than the other methods based on sampling. From the other results in Table~\ref{tbl:pcc-results} we can also observe that MM is the best for Hamming loss, and JM for the subset 0/1 loss. The results for the Jaccard index show that maximization of the F-measure can be used as a proxy for this performance criterion, at least FM and GFM perform better than MM.

%
\begin{table}
\caption{Empirical results on 4 benchmark datasets. PCC is used with different inference methods: MM - estimates marginal modes, JM - the joint mode, FM - approximates the F-measure maximizer by assuming independence of labels, and GFM - estimates the exact F-measure maximizer over the conditional distribution obtained from the model. For MM, FM, and GFM, we sample 1000 observations from the conditional joint distribution for each test example. Results are reported for Hamming loss, subset 0/1 loss, F-measure and Jaccard distance. The best results for a performance measure are marked in bold.}
\label{tbl:pcc-results}
\begin{center}
{\scriptsize
\begin{sc}
\begin{tabular}{@{}l r r r r r r@{}}
\toprule 
    & Hamming   & Subset 0/1& F-Measure [\%] & Jaccard [\%] & Training & Inference  \\
    & Loss [\%] & Loss [\%] &                &              & Time [s] & Time [s]   \\
\midrule
    & \multicolumn{6}{c}{Scene} \\
\midrule
MM  &\bf   9.89 &    42.69  &    62.73 &    61.37 &  39  & 1.839 \\
JM	 &    10.40 &\bf 34.87  &    70.93 &\bf 69.48 &  39  & 0.255 \\
FM	 &    12.83 &    50.44  &\bf 72.78 &    66.71 &  39  & 1.858 \\
GFM &     12.74 &    49.86  & \bf  72.78 &    66.89 &  39  & 1.927 \\ 
\midrule
          & \multicolumn{6}{c}{Yeast} \\
\midrule
MM  &\bf 19.56 &    81.98 &    61.78 &    51.32 &  32 &  7.120 \\
JM  &    20.91 &\bf 76.34 &    63.36 &    53.56 &  32 &  0.220 \\ 
FM  &    22.31 &    84.22 &    65.53 &    54.29 &  32 &  7.161 \\
GFM &    22.54 &    84.73 &\bf 65.63 &\bf 54.32 &  32 &  7.513 \\
\midrule
          & \multicolumn{6}{c}{Enron} \\
\midrule
MM	 &\bf  4.63 &    86.88 &    52.62 &    42.51 & 81 & 120.636 \\
JM	 &     4.81 &\bf 82.56 &    55.67 &    45.80 & 81 &   1.974 \\
FM	 &     5.59 &    90.33 &    58.54 &    46.19 & 81 & 121.172 \\
GFM  &     5.53 &    89.12 &\bf 59.08 &\bf 46.89 & 81 & 121.700 \\
\midrule
          & \multicolumn{6}{c}{Mediamill} \\
\midrule
MM  &\bf  3.18 &    92.44 &    51.21 &    39.60 & 6150 & 2226.455 \\
JM  &     3.57 &\bf 90.02 &    44.99 &    35.62 & 6150 &  28.856  \\
FM  &     3.62 &    94.81 &\bf 55.39 &\bf 42.57 & 6150 & 2230.661 \\
GFM &     3.61 &    94.51 &    55.18 &    42.44 & 6150 & 2293.945 \\
\bottomrule 
\end{tabular}
\end{sc}
}
\end{center}
\end{table}

\subsubsection{Parametric Models}

To complete the picture we also present the experimental results of parametric models that have been previously published in \citep{Dembczynski_et_al_2013}.  We compare EFP and its simplified variant LFP. We also include to the comparison the binary relevance (BR) approach that learns and predicts for each label independently. Such a model should perform well under the Hamming loss.  All the methods use linear models and are trained, similarly as PCC, by logistic regression. The regularization parameter and its tunning is exactly the same as in the experiment with PCC. 

The results are summarized in~Table~\ref{tbl:parametric_models-results}. We show the results for Hamming loss, F-measure and report training and inference times.
Not surprisingly, BR achieves the best results for Hamming loss, but it is outperformed by all the other methods on the F-measure. EFP is the best method in this regard. 
 
\begin{table}[h]
\caption{Empirical results on 4 benchmark datasets of parametric models: BR, LFP, and EFP. We report the Hamming loss, F-measure and training and inference times in seconds. The best results for a performance measure are marked in bold.}
\label{tbl:parametric_models-results}
\begin{center}
{\scriptsize
\begin{sc}
\begin{tabular}{@{}l r r r r@{}}
\toprule 
    & Hamming Loss [\%]& F-Measure [\%] & Training Time [s] & Inference Time [s]  \\
\midrule
    & \multicolumn{4}{c}{Scene} \\
\midrule
BR   & 10.51 & 55.73 &  29 & 0.241 \\
LFP &12.18 & 74.38 &   29 & 0.270 \\
EFP & 12.22 & \bf 74.44 &  72	 & 0.399 \\
\midrule
          & \multicolumn{4}{c}{Yeast} \\
\midrule
 BR      & \bf 20.03 & 60.59 & 26 & 0.128 \\
 LFP     & 22.24 & 65.02 &   26 & 0.146 \\
 EFP     & 22.82 & \bf 65.47  & 101 & 0.367 \\
\midrule
          & \multicolumn{4}{c}{Enron} \\
\midrule
BR     & \bf 4.54 & 55.49 &   52 & 1.016 \\
LFP    &  6.09 & 56.86 &   52 & 1.519 \\
EFP    & 5.34 & \bf 61.04 &  214 & 2.628 \\
\midrule
          & \multicolumn{4}{c}{Mediamill} \\
\midrule
BR      &  \bf 3.19 & 51.21  &   3238 & 13 \\
LFP     & 3.67 & 55.15    &   3238 & 20 \\
EFP     &   3.63 & \bf 55.16 & 24620 & 30 \\
\bottomrule 
\end{tabular}
\end{sc}
}
\end{center}
\end{table}

BR is the most efficient in inference. Nevertheless, the inference times of LFP and EFP are quite comparable to those of BR, despite their quadratic (for LFP) and cubic (for EFP) complexity. Admittedly, however, the datasets used in the experiments only contain a small to moderate number of labels (up to 100). For datasets with thousands of labels, the difference is likely to become substantially larger.  
The training of BR and LFP (these are exactly the same procedures) is the most effective. Training of EFP leads to $m$ multinomial regression models. One should note, however, that the number of classes in each multinomial regression models can be much less than the highest possible value $m+1$. Therefore, the training of EFP is still quite effective and takes only a few times longer than the training of LFP. The training time includes here also the tunning time, similarly as in the case of PCC.

\subsubsection{Comparison with Structured Support Vector Machines}

In this subsection we gather results of all F-measure maximization methods presented so far and compare them with the results of a variant of structured SVMs~\citep{Tsochantaridis_et_al_2005} in which the effort of maximizing the F-measure is moved to the training phase. Two methods of that kind, referred to as RML and SML, have been introduced by \citet{Petterson_Caetano_2010,Petterson_Caetano_2011}. We present here only the results of RML, previously published in \citep{Dembczynski_et_al_2013}.\footnote{The results were obtained by using the software available at \url{http://users.cecs.anu.edu.au/~jpetterson/}}
Basically, this method trains one model for each label, but in a way that the margin, appropriately rescaled by the F-measure, is maximized jointly over all labels. RML uses a variant of the cutting-plane algorithm for optimization. So, in each iteration a most violating constraint for each training example is generated. This step has quadratic complexity in terms of the number of labels. In the prediction phase the models are independently applied to corresponding labels. 
Surprisingly, this method gets usually better results than the more complex SML method~\citep{Petterson_Caetano_2011}, which additionally models pairwise label dependencies.\footnote{The comparison of these two methods is given in \citep{Petterson_Caetano_2011} and \citet{Dembczynski_et_al_2013}.}
In the experiment the RML method uses a linear model. The regularization parameter is tuned in 5-fold cross-validation using a range of values corresponding to the one used for PCC, EFP, and LFP. The maximal number of iterations in the cutting-plane algorithm has been set to 1000.

Table~\ref{tbl:all_results} present the results for F-measure, training and inference times. For  instance-based methods we report the variant with the number of neighbors which gets the best results for a given dataset. Similarly for PCC, we take the number of sampled observations which leads to the best performance. Observing the results we can say that in general the methods that maximize the F-measure in the inference phase outperform the structured SVM approach. RML is only competitive on the \textsc{Scene} dataset, on which it wins against instance-based methods and PCC, but loses from EFP and LFP. On \textsc{Yeast} and \textsc{Mediamill} it gets the worst performance. On the latter dataset the difference is the most substantial. On \textsc{Enron} it wins only against the instance-based method. As we already pointed out, on this dataset all variants of nearest neighbors perform weakly, probably because of the high-dimensional feature space. 

The comparison of the running times between RML and the other methods should be interpreted with caution, due to the use of different programming languages (RML is implemented in C++, while the other algorithms in Java) and differences in the implementations (different data structures). Therefore, the evaluation times may not be fully comparable. For example, the inference times for BR (Table~\ref{tbl:parametric_models-results}) and RML should basically be very similar, as in both cases there is a single linear model for each label. Yet, the implementation of RML is much more efficient. Nevertheless, we are still able to derive several important conclusions. 

Not surprisingly RML is the most efficient in inference. However, the cutting-plane algorithm and the constraint generation step therein slow down significantly the training of RML. This also makes tuning very costly. For PCC, EFP and LFP, the tunning can be performed independently for each base classifier. In that way we hope to obtain a good probabilistic model and there is no need to perform neither a costly training nor inference. Unfortunately, this is not the case of RML. For example, tunning on the \textsc{Mediamill} dataset has not finished in reasonable time. The F-measure result reported in the table is the best one among those obtained on the test set for different values of the regularization parameter. 

Each of the methods maximizing the F-measure in the inference phase has its advantages and disadvantages. By comparing the results, we see that parametric models get the best results on \textsc{Scene} and \textsc{Enron} followed by PCC. On \textsc{Yeast} and \textsc{Mediamill} all the methods perform very similarly, but the best is the instance-based learning here. 
Learning of EFP is the most costly. PCC and LFP train a single model for each label, but in the case of the former algorithm the feature space is enhanced by the preceding labels, therefore, the training time is longer for this procedure. We do not report training time for instance-based methods, as in the simplest case there is no learning in this kind of methods. In general, however, one should consider the time needed for tuning the number of neighbors and optionally learning the metric, which we have not performed here. From the inference point of view, LFP seems to be the most efficient. As we already discussed in the previous subsection, EFP is still quite competitive in comparison with LFP, but for datasets with a large number of labels this difference shall be more substantial. Instance-based methods are also very efficient for datasets with a small number of training examples. Because of the sampling procedure applied  for each testing examples, PCC is the most time demanding procedure. However, we can see from Figure~\ref{fig:pcc} that the good performance with respect to the F-measure can be obtained with a smaller number of sampled observations. In many applications a set of 100 observations should be sufficient, resulting in a sample generation that is approximately 10 times faster. The main advantage of PCC is that once we have a trained model we can apply inference for many different performance measures without any additional training.   

\begin{table}[h]
\caption{Comparison of RML, a variant of structured SVMs for F-measure maximization, with  other F-measure maximization methods given in the previous tables: instance-based methods (IB), PCC, LFP and EFP. We present here the best variant of instance-based methods and PCC for a given dataset. The number $l$ of nearest neighbors for the best variant of IB is given in parentheses. Similarly the number of sampled observations in PCC is also given in parentheses.  The F-measure, training and inferences time in seconds are reported. The best results are marked in bold.}
\label{tbl:all_results}
\begin{center}
{\scriptsize
\begin{sc}
\begin{tabular}{@{}l r r r@{}}
\toprule 
    & F-Measure [\%] & Training Time [s] & Inference Time [s]  \\
\midrule
    & \multicolumn{3}{c}{Scene} \\
\midrule
IB FM  ($l = 10$)    & 70.80     & -      & 3.746 \\
IB GFM ($l = 10$)    & 71.42     & -      & 3.749 \\
PCC FM ($n = 1000$)  & 72.78     & 39     & 1.858 \\
PCC GFM ($n = 1000$) & 72.77     & 39     & 1.927 \\
LFP                  & 74.38     & \bf 29 & 0.270 \\
EFP                  & \bf 74.44 & 72	  & 0.399 \\[3pt]
RML                  & 73.92     & 73     & \bf 0.118 \\
\midrule
          & \multicolumn{3}{c}{Yeast} \\
\midrule
IB FM  ($l = 10$)    & 65.29     & -      & 1.518 \\
IB GFM ($l = 50$)    & \bf 65.75 & -      & 1.795 \\
PCC FM ($n = 1000$)  & 65.53     & 39     & 7.161 \\
PCC GFM ($n = 1000$) & 65.63     & 39     & 7.513 \\
LFP                  & 65.02     & \bf 26 & 0.146 \\
EFP                  & 65.47     & 101    & 0.367 \\[3pt]
RML                  & 64.78     & 206    & \bf 0.056 \\
\midrule
          & \multicolumn{3}{c}{Enron} \\
\midrule
IB FM  ($l = 10$)    & 47.50     & -      & 0.787 \\
IB GFM ($l = 10$)    & 44.43     & -      & 0.810 \\
PCC FM ($n = 500$)   & 58.75     & 81     & 75.677 \\
PCC GFM ($n = 500$)  & 59.19     & 81     & 76.056 \\
LFP                  & 56.86     & \bf 52 & 1.519 \\
EFP                  & \bf 61.04 & 214    & 2.628 \\[3pt]
RML                  & 57.69     & 3897   &  \bf 0.143 \\ 
\midrule
          & \multicolumn{3}{c}{Mediamill} \\
\midrule
IB FM  ($l = 20$)    & 55.65     & -        & 164 \\
IB GFM ($l = 20$)    & \bf 55.80 & -        & 167 \\
PCC FM ($n = 1000$)  & 55.39     & 6150     & 2230 \\
PCC GFM ($n = 1000$) & 55.18     & 6150     & 2293 \\
LFP                  & 55.15     & \bf 3238 & 20 \\
EFP                  & 55.16     & 24620    & 30 \\[3pt]
RML                  & 49.35     & -        & \bf 7 \\
\bottomrule 
\end{tabular}
\end{sc}
}
\end{center}
\end{table}

\subsection{Results in the JRS 2012 Data Mining Competition}
\label{sec:jrs_2012}

We used the algorithms maximizing the F-measure in the inference phase in our solution for the JRS 2012 Data Mining Competition~\citep{Janusz_et_al_2012}.\footnote{\url{http://tunedit.org/challenge/JRS12Contest}} This competition considered topical classification of bio-medical articles. In essence, it consisted of a multi-label learning problem, where the objective was to optimize the instance-based F-measure. We decided to participate in this competition to showcase the practical relevance of our theoretical findings regarding the F-measure maximization. Similar to many of our competitors, our final predictions in the competition were produced by a blend of several methods, and they achieved a very satisfactory result, namely the second place in the competition with more than 100 participants. In this paragraph, we briefly explain the methodology that led to this result.

Our solution was mainly based on PCC with FM and GFM inference methods and the LFP algorithm. The methods were tuned and run in a similar way as described in the previous experiments (with small differences: we used 10-fold cross validation, and considered a wider range for the regularization parameter, namely $\{10^{-5}, 10^{-4}, \ldots, 10^5\}$). At this time the EFP algorithm was not yet developed. We used neither instance-based nor decision tree methods. 
Our preprocessing on the competition data was quite straightforward. We simply deleted all the empty columns (i.e., zero vectors) in the training data, then the corresponding columns in the test data. The values of features were normalized to $[0,1]$. 

The results of the methods are presented in Table~\ref{tbl:results}. The F-measure is computed over the entire test set delivered by the organizers after the competition. This is a minor difference in comparison to the competition results, which are computed over 90\% of test examples. The remaining 10\% of test examples constitute a validation set that served for computing the scores for the leaderboard during the competition. The results of PCC we show for different sizes of samples generated from the conditional joint distribution of a given test example. In the last row in the table, we also give a result of the final method we used in the competition. It relies on averaging over all predictions we computed during the competition. These predictions were  results of different parameterization of PCC and BR. In total we gathered 16 predictions that we aggregated via voting. In this voting procedure, we tested different thresholds on the validation set and selected the best one. The solution is described in more detail in~\citep{Cheng_et_al_2012}. 

From the results we can see that there is no big difference among the methods. The voting procedure improves only slightly over LFP and PCC with the GFM inference. The results of these methods would be enough to obtain at least the third place in the competition. It shows that a quite simple model, without any blending, but with an appropriate inference method suited for a given performance measure is enough for solving complex tasks. Interestingly, LFP performs here better than PCC with GFM, which suggests independence of the labels. However, one can also observe that PCC with FM loses against other methods. This may suggest that PCC with the sampling procedure has problems with accurate estimation of marginal probabilities. Increasing the sample size improves the results (for both, FM and GFM), but it still seems that LFP is the most appropriate method in this case. It is the most cheapest one, since it does not require additional sampling in the inference step as PCC does, and gives results only slightly worse than the voting method that averages over many predictions. As we already said before, there is no clear answer which of the two inference methods, GFM or FM, will get better results on a given dataset. GFM provides an exact solution, but needs to estimate more parameters, so FM may get better results, particularly in the case of no or weak label dependencies.   


\begin{table}[t]
\caption{\small The results on the JRS 2012 Competion dataset. The number $n$ in parentheses denotes the number of sampled observations in PCC.} 
\label{tbl:results}
\begin{center}
\begin{tabular}{ll}
\toprule
Method & F-measure \\
\midrule
PCC FM ($n =50$) & 0.48650    \\
PCC FM ($n = 200$) & 0.51979   \\
PCC FM ($n = 1000$) & 0.52995 \\
PCC GFM ($n = 50$)  & 0.52286 \\
PCC GFM ($n = 200$) & 0.53005 \\
PCC GFM ($n = 1000$) & 0.53146 \\
LFP & 0.53279     \\ 
Voting (final submission) & 0.53327\\ 

\bottomrule
\end{tabular}
\end{center}
\vspace{-0.5cm}
\end{table}


\section{Discussion}

In contrast to other performance measures commonly used in experimental studies, such as error rate, squared loss, and AUC, the F-measure has been investigated less thoroughly from a theoretical point of view, and only few papers have been devoted to that kind of analysis so far  (e.g. \citet{lewis95,Chai2005,jansche07,Dembczynski_et_al_2011,Ye2012,Zhao2013}). In
this paper, we analyzed the problem of optimal predictive inference from
the joint distribution under the F-measure. While partial results were
already known from the literature, we completed the picture by presenting
the solution for the general case without any distributional assumptions and by analyzing the relations between F and other performance measures.
Our GFM algorithm requires only a polynomial number of parameters of the joint distribution and delivers the exact solution in polynomial time. From a theoretical perspective, GFM should be preferred to existing approaches, which typically perform threshold maximization on marginal probabilities, often relying on the assumption of (conditional) independence of labels. Focusing on optimizing the instance-wise F-measure, empirical results on synthetic and real-world multi-label datasets show a competitive performance for our approach.

The algorithms discussed here optimize the F-measure in the inference phase. Alternatively, one can move the effort of maximizing the F-measure to the training phase, as in 
structured  SVMs~\citep{Tsochantaridis_et_al_2005}, SEARN~\citep{Daume_et_al_2009}, or in a specific variant of CRFs~\citep{Suzuki_et_al_2006}. These algorithms, however, are usually based on additional assumptions, and their original formulation does not directly concern multi-label problems.
In the experiments, we performed a comparison to the adaptation of  structured SVMs to the multi-label setting introduced by \citet{Petterson_Caetano_2010,Petterson_Caetano_2011}. This algorithm also maximizes the F-measure, but gets worse results than the approaches based on the GFM inference. However, its prediction time is much faster, giving an interesting alternative in time-critical applications.


Let us also mention that the GFM algorithm can be easily tailored for maximizing the instance-wise F-measure in structured output prediction problems. If the structured output classifier is able to model the joint distribution, from which we can easily sample observations, then the use of the algorithm is straight-forward. An application of this kind is planned as future work. 


The GFM algorithm could also be considered for maximizing the macro F-measure, for example, in a similar setting as in~\citep{zhang10}, where a specific Bayesian on-line model is used. In order to maximize the macro F-measure, the authors sample from the graphical model to find an optimal threshold. The GFM algorithm may solve this problem optimally, since, as stated by the authors, the independence of labels is lost after integrating out the model parameters. Theoretically, one may also consider a direct maximization of the micro F-measure with GFM, but the computational burden is rather high in this case. We would also like to emphasize that maximization of instance-based F-measure leads to suboptimal results for the micro F-measure. Despite being related to each other, these two measures coincide only in a specific case when $\sum_{i=1}^m (y_i + h_i)$ is constant for all test examples. The discrepancy between these measures strongly depends on the nature of the data and the classifier used. For high variability in $\sum_{i=1}^m (y_i + h_i)$, a significant difference between the values of these two measures is to be expected. Surprisingly, experimental results are quite often reported in terms of micro F-measure, although the algorithms maximize the instance-wise F-measure on the training set.

The use of the GFM algorithm in binary classification seems to be superfluous, since in this case, the assumption of label independence is rather reasonable. The algorithm of \citet{Ye2012} seems to be an interesting alternative for probabilistic classifiers. Thresholding methods~\citep{Keerthi_et_al_2007,Ye2012,Zhao2013} or learning algorithms optimizing the F-measure directly~\citep{Musicant_et_al_2003,Joachims_2005,Jansche_2005,Ye2012} are also appropriate solutions here. 

%


\section*{Acknowledgments}
We thank Volkmar Welker for carefully checking the proofs of some theorems. Willem Waegeman was for this work supported as a postdoctoral fellow by the Research Foundation of Flanders (FWO Vlaanderen). Krzysztof Dembczy\'nski and Arkadiusz Jachnik were supported by the Foundation of Polish Science under the Homing Plus programme, co-financed by the European Regional
Development Fund. Weiwei Cheng and Eyke H\"ullermeier were supported by the German Research Foundation.

\section*{Proofs of Theorems}

\noindent {\bf Theorem 3.1}
{\it Let $\vec{h}_H$ be a vector of predictions obtained by minimizing the Hamming loss, 
Then for $m > 2$ the worst-case regret is given by:
\begin{eqnarray*}
\sup_{\Pr \in \mathcal{P}^u_{L_H}} \, (\mathbb{E} \big[ F(\brY, \vec{h}_F) -  F(\brY, \vec{h}_H) \big]) = 0.5 \,, \\
\end{eqnarray*}
where the supremum is taken over all possible distributions $\Pr$ that result in a unique F-measure maximizer and Hamming loss minimizer.}

\begin{proof}
For a fixed Hamming loss minimizer $\vec{h}_H$ it follows from (\ref{eq:hm}) that any probability distribution $\Pr \in \mathcal{P}^u_{L_H}$ should satisfy the following constraint for all $i \in \{1,...,m\}$:  
\begin{equation*}
\sum_{\vec{y} \in \{0,1\}^m : y_i \neq h_{H,i}} \Pr(\vec{y}) \leq 0.5 - \epsilon 
\end{equation*}
with $\epsilon > 0$. Practically, we will choose $\epsilon$ arbitrarily close to zero, implying that its contribution vanishes in the limit, but this construction allows to rewrite the constraint in traditional mathematical programming form. Let us also define 
$$\eta_{\vec{y}}(\vec{h},\vec{h}_H) = F(\vec{y},\vec{h}) - F(\vec{y},\vec{h}_H) $$
for all $\vec{y} \in \{0,1\}^m$.
Finding the supremum over all probability distributions then becomes equivalent to solving the following mixed integer nonlinear program:
\begin{eqnarray}
          \max_{\vec{h},\vec{h}_H,\Pr \in  \mathcal{P}^u_{L_H}} \sum_{\vec{y} \in \{0,1\}^m} \eta_{\vec{y}}(\vec{h},\vec{h}_H)  \Pr(\vec{y}) \label{eq:lpHamming}
    \end{eqnarray}
    \begin{eqnarray*}
        \textrm{subject to} \left\{ \begin{array}{ll}
        \sum_{\vec{y} \in \{0,1\}^m} \Pr(\vec{y}) = 1 \,,\\
        \forall i \in \{1,...,m\}:  \sum_{\vec{y} \in \{0,1\}^m : y_i \neq h_{H,i}} \Pr(\vec{y}) \leq 0.5 - \epsilon \,, \\
        \forall \vec{y} \in \{0,1\}^m : 0 \leq \Pr(\vec{y}) \leq 1 \,, \\
        \vec{h},\vec{h}_H \in \{0,1\}^m
        \end{array} \right.
\end{eqnarray*}
By definition, the solution $\vec{h}$ for which the maximum is obtained corresponds to the F-measure maximizer in (\ref{eq:argmax_f1}). 


To reduce the number of integer variables in the optimization problem, we introduce the following equivalence classes for the indices of labels:
\begin{eqnarray*}
A &=& \{i \in \{1,...,m\} : h_i = 1 \wedge h_{H,i} = 0 \} \,, \\
B &=& \{i \in \{1,...,m\} : h_i = 0 \wedge h_{H,i} = 1 \} \,, \\
C &=& \{i \in \{1,...,m\} : h_i = 1 \wedge h_{H,i} = 1 \} \,, \\
D &=& \{i \in \{1,...,m\} : h_i = 0 \wedge h_{H,i} = 0 \} \,.
\end{eqnarray*}
We also adopt the shorthand notation $a = |A|, b = |B|, c = |C|, d = |D|$ and
\begin{eqnarray}
\label{eq:sabcd}
s_{\vec{y}} = \sum_{i=1}^m y_i \,, \quad s_{\vec{y}}^A  = \sum_{i \in A} y_i \,, \quad s_{\vec{y}}^B = \sum_{i \in B} y_i \,, \quad s_{\vec{y}}^C = \sum_{i \in C} y_i \,, \quad s_{\vec{y}}^D = \sum_{i \in D} y_i \,.
\end{eqnarray}
The coefficients in (\ref{eq:lpHamming}) can then be simplified to:
\begin{eqnarray}
\label{eq:eta}
\eta_{\vec{y}}(\vec{h},\vec{h}_H) = \eta_{\vec{y}}(a,b,c,d) = \frac{2 s_{\vec{y}}^A (s_{\vec{y}} + b + c) - 2s_{\vec{y}}^B (s_{\vec{y}} + a+c) + 2s_{\vec{y}}^C (b-a)}{(s_{\vec{y}}+ a+c )(s_{\vec{y}} +b+c)} \,.
\end{eqnarray}
As a consequence, only four integer variables remain present in the optimization problem, which is for simplicity converted to a minimization problem in standard mixed-integer linear program form:
\begin{eqnarray*}
          \min_{a,b,c,d,\Pr} - \sum_{\vec{y} \in \{0,1\}^m} \eta_{\vec{y}}(a,b,c,d) \Pr(\vec{y})
    \end{eqnarray*}
    \begin{eqnarray*}
        \textrm{subject to} \left\{ \begin{array}{ll}
        \sum_{\vec{y} \in \{0,1\}^m} \Pr(\vec{y}) = 1 \,,\\
        \forall i \in \{1,...,m\}:  \sum_{\vec{y} \in \{0,1\}^m : y_i \neq h_{H,i}} \Pr(\vec{y}) \leq 0.5 - \epsilon \,, \\
        \forall \vec{y} \in \{0,1\}^m : 0 \leq \Pr(\vec{y}) \leq 1 \,, \\
        a+b+c+d = m \,, \\
        a,b,c,d \in \mathbb{N}\,.
        \end{array} \right.
\end{eqnarray*}
This new optimization problem is a relaxation of (\ref{eq:lpHamming}), since the F-maximizer of the probability distribution found as solution will not necessarily comply with the definition of the sets $A$, $B$, $C$ and $D$. However, this will not cause any trouble, because the oracle solution that is derived below will obey this additional constraint.  

One arrives at a standard linear program formulation by keeping the four integer variables fixed. As the key element of our proof, we show that for every allowed value of $(a,b,c,d)$ a solution of the linear program is given by the following probability distribution:
\begin{eqnarray*}
\Pr_A(\vec{y}) & = & \left\{ \begin{array}{cl} 0.5 - \epsilon  & \quad
\textrm{if $\vec{y} = \vec{y}^A \,$}\\
0.5 - (2d-1) \epsilon  & \quad \textrm{if $\vec{y} = \vec{y}^{BCD} \,$}\\
2 \epsilon  & \quad \textrm{if $\vec{y} \in \Omega^D_m \,$} \\
0  & \quad \textrm{otherwise $\,$}
\end{array} \right. \enspace ,
\end{eqnarray*}
where $\vec{y}^A$ is defined as a vector containing ones at positions $i \in A$ and zeros at all other positions. Similarly, $\vec{y}^{BCD}$ contains zeros at positions $i \in A$ and ones at all other positions:
\begin{eqnarray*}
y^A_i = 1 + y^{BCD}_i = 1, &&\forall i \in A \,, \\
y^A_i = 1 - y^{BCD}_i = 0, &&\forall i \in B \cup C \cup D\,. \\
\end{eqnarray*}
The set $\Omega_m^D$ is defined as:
$$\Omega_m^D = \{\vec{y} \in \{0,1\}^m \mid \sum_{i \in A} y_i = 0 \wedge \sum_{i \in B \cup C} y_i = b + c \wedge \sum_{i \in D} y_i = d-1  \} \,,$$
so this set contains $d$ vectors, which differ only in one position with $\vec{y}^{BCD}$.

We verify the Karush-Kuhn-Tucker (KKT) conditions to prove that the above probability distribution yields the optimum of the linear program for every $(a,b,c,d)$. For linear programs, which represent a specific case of optimizing an invex function, the KKT-conditions are not only necessary but also sufficient for optimality \citep{Hanson1981}. Let us define the primal Lagrangian as:
\begin{eqnarray*}
\mathfrak{L}_p &=& - \sum_{\vec{y} \in \{0,1\}^m} \eta_{\vec{y}}(a,b,c,d) \Pr(\vec{y}) + \nu
        \sum_{\vec{y} \in \{0,1\}^m} \big( \Pr(\vec{y}) - 1 \big) \\
        && \qquad + \sum_{i=1}^m \mu_i \big[ \sum_{\vec{y} \in \{0,1\}^m : y_i \neq h_{H,i}} \Pr(\vec{y}) - 0.5 + \epsilon \big] \\
        && \qquad - \sum_{\vec{y} \in \{0,1\}^m} \lambda^-_{\vec{}y} \Pr(\vec{y})
        + \sum_{\vec{y} \in \{0,1\}^m} \lambda^+_{\vec{y}} (\Pr(\vec{y}) - 1) \,.
\end{eqnarray*}
with $\nu, \mu_i, \lambda^+_{\vec{y}}$ and $\lambda^-_{\vec{y}}$ Lagrange multipliers. For the above-mentioned probability distribution, the complementary slackness conditions imply that $\lambda^+_{\vec{y}} = 0$ for all $\vec{y} \in \{0,1\}^m$ and $\lambda^-_{\vec{y}} = 0$ for all $\vec{y}$ contained in $\Omega^D_m \cup \{\vec{y}^A,\vec{y}^{BCD}\}$. Hence, the zero-gradient condition results in the following system of equations:
\begin{eqnarray*}
\eta_{\vec{y}}(a,b,c,d) &=&  \nu + \sum_{i: y_i \neq h_{H,i}} \mu_i \,, \qquad \qquad \forall  \vec{y} \in \Omega^D_m \cup \{\vec{y}^A, \vec{y}^{ABC}\} \,, \\
\eta_{\vec{y}}(a,b,c,d) &=&  \nu  + \sum_{i: y_i \neq h_{H,i}} \mu_i - \lambda_{\vec{y}}^- \,, \qquad \forall  \vec{y} \notin \Omega^D_m \cup \{\vec{y}^A, \vec{y}^{ABC}\} \,.
\end{eqnarray*}
First, we show that a solution always exists for this system, and
additionally, we check the dual feasibility conditions $\mu_i \geq 0, \lambda^+_{\vec{y}} \geq 0$ and $\lambda^-_{\vec{y}} \geq 0$. For all $\vec{y} \notin \Omega^D_m \cup \{\vec{y}^A, \vec{y}^{ABC}\}$, the equations have an individual variable $\lambda_{\vec{y}}^-$, so these equations do not impose any further restrictions, apart from the non-negativity constraint on the respective Lagrange multipliers. For the other equations, one arrives at a new system of equations by solving for $\nu$:
\begin{eqnarray*}
\sum_{i: y_i \neq h_{H,i}} \mu_i - \sum_{j: y'_j \neq h_{H,j}} \mu_j &=& \eta_{\vec{y}}(a,b,c,d) - \eta_{\vec{y}'}(a,b,c,d)  \,, \\
&& \qquad \qquad  \forall  \vec{y},\vec{y}' \in \Omega^D_m \cup \{\vec{y}^A, \vec{y}^{ABC}\} \,.
\end{eqnarray*}
Writing out these equations for all pairs explicitly yields:
\begin{eqnarray}
\label{eq:line1}
\sum_{i \in A \cup B \cup C} \mu_i - \sum_{i \in D} \mu_i = \eta_{\vec{y}}(a,b,c,d) - \eta_{\vec{y}'}(a,b,c,d) \,,
\end{eqnarray}
for the pair corresponding to $\vec{y} = \vec{y}^A$ and $\vec{y}' = \vec{y}^{BCD}$,
\begin{eqnarray}
\label{eq:line2}
\sum_{i \in A \cup B \cup C} \mu_i - \sum_{i \in D: i \neq j} \mu_i = \eta_{\vec{y}}(a,b,c,d) - \eta_{\vec{y}'}(a,b,c,d) \,,
\end{eqnarray}
for all pairs having $\vec{y} = \vec{y}^{A}$ and $\vec{y}' \in \Omega^D_m$, so $d$ pairs in total with $j \in D$,
\begin{eqnarray}
\label{eq:line3}
\mu_i = \eta_{\vec{y}}(a,b,c,d) - \eta_{\vec{y}'}(a,b,c,d) \,,
\end{eqnarray}
for all pairs having $\vec{y} = \vec{y}^{BCD}$ and  $\vec{y}' \in \Omega^D_m$, so again $d$ pairs with  $i \in D$,
\begin{eqnarray}
\label{eq:line4}
\mu_i = \mu_j \,,
\end{eqnarray}
for all pairs having $\vec{y}, \vec{y}' \in \Omega^D_m$, so $d(d-1)$ pairs with $i,j \in D$.

Let us observe that the $d$ equations (\ref{eq:line2}) and the $d$ equations (\ref{eq:line3}) are equivalent due to (\ref{eq:line1}). Moreover, the $d(d-1)$ equations in (\ref{eq:line4}) are trivially satisfied, resulting in a system that can be reduced to $d +1$ equations and $m$ variables. Hence, a solution always exists if the dual feasibility conditions are satisfied. Plugging (\ref{eq:eta}) into (\ref{eq:line3}) yields for $i \in D$:
\begin{eqnarray*}
\mu_i &=&  \frac{2c(b-a)}{(m+c)(2b+2c+d)} - \frac{2b}{2b+2c+d} \\ && \qquad -\frac{2c(b-a)}{(m+c-1)(2b+2c+d-1)} + \frac{2b}{2b+2c+d-1}  \\
&\geq&  \frac{2 cb(m-2)}{(m+c)(m-1+c)(2b+2c+d)} \,.
\end{eqnarray*}
So, $\mu_i \geq 0$ as soon as $m>1$. Due to the denominator it is also required that $2b + 2c + d > 1$. For the $\mu_i$ having $i \notin D$ we find:
\begin{eqnarray*}
\sum_{i \in A \cup B \cup C} \mu_i  &\geq& \eta_{\vec{y}}(a,b,c,d) - \eta_{\vec{y}'}(a,b,c,d) \\
&=& \frac{2a}{2a+c} + \frac{2b}{2b+2c+d} \\
&& \qquad - \frac{2c(b-a)}{(m+c)(2b+2c+d)} \\
&\geq& \frac{2mb}{(m+c)(2b+2c+d)} \geq 0 \,.
\end{eqnarray*}
Thus, the individual $\mu_i$ can be made greater than zero in this case, implying that also the active Lagrange multipliers $\lambda_{\vec{y}}^-$ can be chosen in such a way that they are greater than zero.

Consequently, when $m>1$, all KKT conditions are satisfied for the oracle solution that we provide. Let us compare the value of the objective function for all allowed values of the four integer variables $a,b,c$ and $d$. By omitting $\epsilon$-dependent terms, which vanish when $\epsilon$ approaches zero, we obtain:
\begin{eqnarray*}
\sum_{\vec{y} \in \Omega^D_m \cup \{\vec{y}^A, \vec{y}^{BCD}\}} \eta_{\vec{y}}(a,b,c,d) \Pr(\vec{y}) &=& \frac{a}{(2a+c)} - \frac{ca}{(m+c)(2b+2c+d)} \\
&&- \quad \frac{mb}{(m+c)(2b+2c+d)} \,.
\end{eqnarray*}
This function is decreasing in $b$ and $c$, it is constant in $a$ as soon as $c=0$ and $a>0$. It is constant in $d$ when $b+c =0$, thus its maximum of 0.5 is not unique. This maximum is for example obtained for $(a=1,b=0,c=0,d=m-1)$ and it corresponds to the worst-case regret mentioned in the theorem. Remark as well that the solution imposes as additional constraint $m>2$, as a result of the previous constraint $2b + 2c + d > 1$. Two cases were excluded from our analysis: $(a=m,b=0,c=0,d=0)$ and $(a=m-1,b=0,c=0,d=1)$. These cases do not deserve further attention, since they lead to a worst-case regret that is always upper bounded by 0.5. 
\end{proof}

\noindent {\bf Theorem 3.2}
{\it Let $\vec{h}_s$ be a vector of predictions obtained by minimizing the subset 0/1 loss, then for $m > 2$ the worst-case regret is given by:
\begin{eqnarray*}
\sup_{\Pr \in \mathcal{P}^u_{L_s}} \, (\mathbb{E} \big[ F(\brY, \vec{h}_F) -  F(\brY, \vec{h}_s) \big]) = \frac{ (-m-2+2m^2)m}{(2m-1)(4+m+m^2)} \,, \\
\end{eqnarray*}
where the supremum is taken over all possible distributions $\Pr$.}

\begin{proof}
It follows from (\ref{eqn:zero-one_loss}) that optimization problem (\ref{eq:argmin_L}) has a unique solution for the subset zero-one loss if and only if the underlying probability distribution has a unique mode. Translating this requirement into a mathematical programming formulation implies that any probability distribution $\Pr \in \mathcal{P}^u_{L_s}$ should satisfy the following constraint:  
\begin{equation*}
\Pr(\vec{y}) +\epsilon \leq \Pr(\vec{h}_s) \,,
\end{equation*}
for all $\vec{y} \in \{0,1\}^m \setminus \vec{h}_s$ and any $\epsilon > 0$. Practically, the contribution of $\epsilon$ will again vanish by choosing it arbitrarily close to zero. 

As a result, the supremum can be interpreted as the solution of a mixed integer nonlinear program:
\begin{eqnarray}
    \label{opt:excesssubset01}
          \max_{\vec{h},\vec{h}_s,\Pr} \sum_{\vec{y} \in \{0,1\}^m} (F(\vec{y},\vec{h}) - F(\vec{y},\vec{h}_s)) \Pr(\vec{y})
    \end{eqnarray}
    \begin{eqnarray*}
        \textrm{subject to} \left\{ \begin{array}{ll}
        \sum_{\vec{y} \in \{0,1\}^m} \Pr(\vec{y}) = 1 \,,\\
        \forall \vec{y} \in \{0,1\}^m \setminus \vec{h}_s: \Pr(\vec{y}) +\epsilon \leq \Pr(\vec{h}_s)  \,, \\
        \forall \vec{y} \in \{0,1\}^m : 0 \leq \Pr(\vec{y}) \leq 1 \,, \\
        \vec{h}_s,\vec{h}_F \in \{0,1\}^m  \quad \epsilon \geq 0\,.
        \end{array} \right.
\end{eqnarray*}
Recall that by construction the solution for $\vec{h}$ again coincides with the F-measure maximizer in (\ref{eq:argmax_f1}). The mixed integer nonlinear program contains $2^{2m}$ integer variables and $2^m$ real variables. In what follows, we assume that $\epsilon$ is arbitrarily close to zero, so that all $\epsilon$-dependent terms cancel out, while guaranteeing unique risk minimizers. The only consequence of this decision is that the presented solution acts as a supremum (the maximum is not reached). 

Despite a similar formulation as the previous theorem, our proof technique will be quite different. The first part of the proof is a bit similar to a proof given for the regret of the subset 0/1 loss minimizer w.r.t.\ the Hamming loss \citep{Demb2012a}. Unlike the previous theorem, it is impossible to derive an oracle solution for the entire mixed integer nonlinear program. Alternatively, we start by providing an oracle solution for the linear program that is obtained by fixing the integer variables to $\vec{h}_F = \vec{1}_m$ and $\vec{h}_s = \vec{0}_m$. Optimization problem (\ref{opt:excesssubset01}) can then be reformulated in standard linear program form as: 
\begin{eqnarray*}
          \min_{\Pr} \sum_{\vec{y} \in \{0,1\}^m} - \eta(\vec{y}) \Pr(\vec{y})
    \end{eqnarray*}
    \begin{eqnarray*}
        \textrm{subject to} \left\{ \begin{array}{ll}
        \sum_{\vec{y} \in \{0,1\}^m} \Pr(\vec{y}) - 1 = 0 \,,\\
        \forall \vec{y} \in \{0,1\}^m \setminus \vec{0}_m: \Pr(\vec{y}) - \Pr(\vec{0}_m) \leq 0  \,, \\
        \forall \vec{y} \in \{0,1\}^m :
- \Pr(\vec{y}) \leq 0 \,, \\
        \forall \vec{y} \in \{0,1\}^m : \Pr(\vec{y}) - 1\leq 0 \,,
        \end{array} \right.
\end{eqnarray*}  
with 
\begin{eqnarray*}
\eta(\vec{y}) & = & \left\{ \begin{array}{cl} \frac{2s_{\vec{y}}}{s_{\vec{y}} + m}  & \quad
\textrm{if $\vec{y} \neq \vec{0}_m \,,$} \\
-1  & \quad \textrm{if $\vec{y} = \vec{0}_m \,.$}
\end{array} \right. 
\end{eqnarray*}
We verify the KKT conditions to show that the following probability distribution corresponds to the solution of the linear program:
\begin{eqnarray*}
\Pr_A(\vec{y}) & = & \left\{ \begin{array}{cl} \frac{2}{m^2 + m + 4} & \quad
\textrm{if $d(\vec{y},\vec{0}_m) \geq m-2 \vee \vec{y} = \vec{0}_m \,,$} \\
0  & \quad \textrm{otherwise $\,,$}
\end{array} \right. 
\end{eqnarray*}
where $d_H(\vec{y},\vec{y}') = \sum_{i=1}^m |y_i - {y'}_i |$ denotes the Hamming distance. Recall that for simplicity the above probability distribution represents a case with non-unique F-measure maximizer and subset 0/1 loss minimizer. However, unique risk minimizers can be easily obtained by introducing $\epsilon$-dependent terms that vanish when $\epsilon$ approaches zero. The primal Lagrangian of the linear program can be defined as:
\begin{eqnarray*}
\mathfrak{L}_p &=& - \sum_{\vec{y} \in \{0,1\}^m} \eta(\vec{y}) \Pr(\vec{y}) + \nu
        \sum_{\vec{y} \in \{0,1\}^m} \big( \Pr(\vec{y}) - 1 \big) + \sum_{ \vec{y} \neq \vec{0}_m} \lambda^2_{\vec{y}} \big( \Pr(\vec{y}) - \Pr(\vec{0}_m) \big) \\
        && \qquad - \sum_{\vec{y} \in \{0,1\}^m} \lambda^0_{\vec{y}} \Pr(\vec{y})
        + \sum_{\vec{y} \in \{0,1\}^m} \lambda^1_{\vec{y}} \Pr(\vec{y}) \,,
\end{eqnarray*}
with $\nu, \lambda^0_{\vec{y}}, \lambda^1_{\vec{y}}$ and $\lambda^2_{\vec{y}}$ Lagrange multipliers.
The stationarity condition for optimality leads to the following system of linear equations:
\begin{eqnarray}
\label{eq:stat1}
-\eta(\vec{y}) + \nu
+ \lambda^1_{\vec{y}} - \lambda^0_{\vec{y}} + \lambda^2_{\vec{y}} &=& 0 \quad \quad \forall \vec{y} \neq \vec{0}_m \,,\\
\label{eq:stat2}  1 + \nu
+ \lambda^1_{\vec{y}} - \lambda^0_{\vec{y}} - \sum_{ \vec{y} \neq \vec{0}_m} \lambda^2_{\vec{y}} &=& 0 \quad \quad \vec{y} = \vec{0}_m 
\end{eqnarray}
Other conditions that need to be satisfied are dual feasibility
\begin{eqnarray}
\label{eq:lam0} \forall \vec{y}: &&\lambda^0_{\vec{y}} \geq 0 \,,\\
\label{eq:lam1} \forall \vec{y}: &&\lambda^1_{\vec{y}} \geq 0 \,,\\
\label{eq:lam2} \forall \vec{y}: &&\lambda^2_{\vec{y}} \geq 0 \,,
\end{eqnarray}
and the complementary slackness conditions of our oracle solution $\Pr_A(\vec{y})$:
\begin{eqnarray*}
 \forall \vec{y} \in \Omega^u \cup \{\vec{0}_m\}: &&\lambda^0_{\vec{y}} = 0 \,,\\
\forall \vec{y}: &&\lambda^1_{\vec{y}} = 0 \,,\\
\forall \vec{y} \notin \Omega^u: &&\lambda^2_{\vec{y}} = 0 \,,\\
\end{eqnarray*}
where $\Omega^u = \Omega(m) \cup \Omega(m-1) \cup \Omega(m-2)$ and $\Omega(t) = \{\vec{y} \in \{0,1\}^m \mid d_H(\vec{0}_m,\vec{y}) = t\}$. Plugging the latter three conditions into (\ref{eq:stat1}) and (\ref{eq:stat2}) yields
$$
\begin{array}{ll}
-\eta(\vec{y}) + v - \lambda^0_{\vec{y}} = 0\,, &  \quad \forall \vec{y} \notin \Omega^u \cup \{\vec{0}_m\}\,,\\
-\eta(\vec{y}) + v + \lambda^2_{\vec{y}} = 0\,, &  \quad \forall \vec{y} \in \Omega^u \,,\\
v = \sum_{ \vec{y} \neq \vec{0}_m} \lambda^2_{\vec{y}} - 1 \,. & 
\end{array}
$$
Solving the last equation for $v$ results in
\begin{eqnarray*}
v = \frac{(-m-2+2m^2)m}{(2m-1)(4+m+m^2)} \,.
\end{eqnarray*}
Subsequently, one can verify that this solution for $v$ obeys the non-negativity conditions for all $\lambda_{\vec{y}}$. The non-negativity of $\lambda^2_{\vec{y}}$ turns out to be most restrictive for the equivalence class $\vec{y} \in \Omega(m-2)$. In this case we obtain:
\begin{eqnarray}
\label{eq:lambda2}
\lambda^2_{\vec{y}} = \frac{2m-4}{2m-2} -v = \frac{2(3m^2-10m+4)}{(m-1)(2m-1)(4+m+m^2)} \,.
\end{eqnarray} 
Analyzing this function more thoroughly reveals that it is strictly positive in the interval $[+3,+\infty[$.
Similarly, we find that the most restrictive condition on $\lambda^0_{\vec{y}}$ is obtained for the elements in the equivalence class $\vec{y} \in \Omega(m-3)$, leading to the following equality:
\begin{eqnarray}
\label{eq:lambda0}
\lambda^0_{\vec{y}} = v - \frac{2m-6}{2m-3} = \frac{-9m^2+2m^3+56m-24}{(2m-3)(2m-1)(4+m+m^2)} \,. 
\end{eqnarray}
This function is also strictly positive in the interval $[+3,+\infty[$, so all non-negativity conditions are satisfied. Plots of the functions are shown in Figure~\ref{fig:lambda}.
\begin{figure}
\begin{center}
\begin{tabular}{cc}
$\lambda^2_{\vec{y}}$ & $\lambda^0_{\vec{y}}$ \\[3pt]
\includegraphics[scale=0.4]{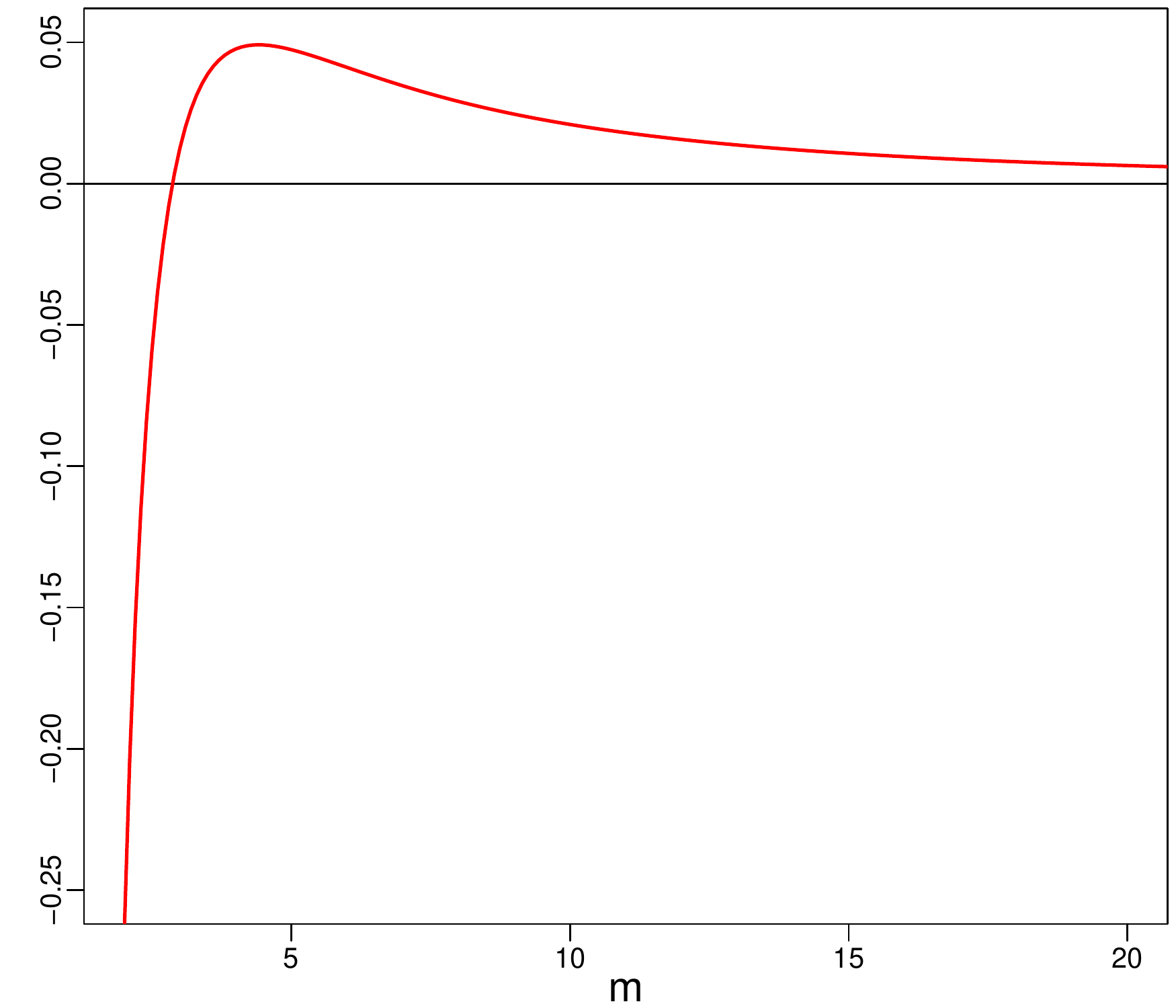} & \includegraphics[scale=0.4]{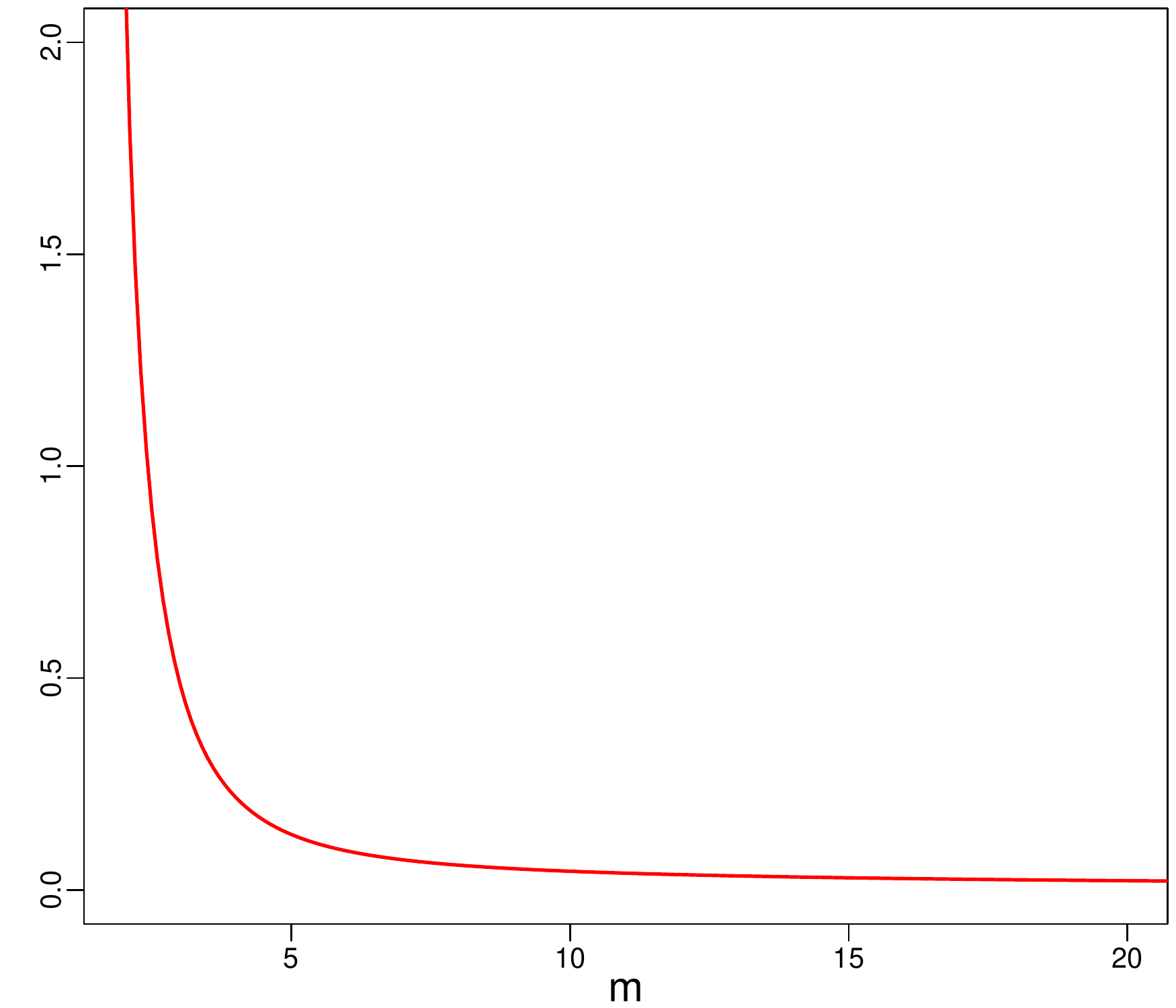} \\
\end{tabular}
\caption{Plots of the functions $\lambda^2_{\vec{y}}$ and $\lambda^0_{\vec{y}}$ as defined by equations (\ref{eq:lambda2}) and (\ref{eq:lambda0}).} 
\label{fig:lambda}
\end{center}
\end{figure}

One can observe that $\Pr_A(\vec{y})$ yields the regret mentioned in the theorem. Thus, what we found so far is a lower bound on the worst-case regret. The tightness of the bound is further proven by showing that the supremum is always obtained by $\vec{h}_s = \vec{0}_m$ and $\vec{h}_F = \vec{1}_m$ as soon as $m>2$. Since it is impossible to enumerate all solutions for the $2^{2m}$ possible values of the integer variables, we analyse the properties of the objective function to prove that the optimum is obtained for $\vec{h}_s = \vec{0}_m$ and $\vec{h}_F = \vec{1}_m$. Similar to the previous theorem, let us introduce
\begin{eqnarray*}
A &=& \{i \in \{1,...,m\} : h_i = 1 \wedge h_{s,i} = 0 \} \,, \\
B &=& \{i \in \{1,...,m\} : h_i = 0 \wedge h_{s,i} = 1 \} \,, \\
C &=& \{i \in \{1,...,m\} : h_i = 1 \wedge h_{s,i} = 1 \} \,, \\
D &=& \{i \in \{1,...,m\} : h_i = 0 \wedge h_{s,i} = 0 \} \,.
\end{eqnarray*}
and the shorthand notations $a = |A|, b = |B|, c = |C|, d = |D|, 
s_{\vec{y}}, s_{\vec{y}}^A, s_{\vec{y}}^B, s_{\vec{y}}^C, s_{\vec{y}}^D$, as defined in (\ref{eq:sabcd}).

Optimization problem (\ref{opt:excesssubset01}) can then be relaxed to the following standard mixed-integer nonlinear program form:
\begin{eqnarray*}
          \min_{a,b,c,d,\Pr} - \sum_{\vec{y} \in \{0,1\}^m} \eta_{\vec{y}}(a,b,c,d) \Pr(\vec{y})
    \end{eqnarray*}
    \begin{eqnarray*}
        \textrm{subject to} \left\{ \begin{array}{ll}
        \sum_{\vec{y} \in \{0,1\}^m} \Pr(\vec{y}) = 1 \,,\\
        \forall \vec{y} \in \{0,1\}^m \setminus \vec{h}_s: \Pr(\vec{y}) - \Pr(\vec{h}_s) \leq 0  \,, \\
        \forall \vec{y} \in \{0,1\}^m : 0 \leq \Pr(\vec{y}) \leq 1 \,, \\
        a+b+c+d = m \,, \\
        a,b,c,d \in \mathbb{N}\,,
        \end{array} \right.
\end{eqnarray*}
writing down the coefficients of the objective function as:
\begin{eqnarray*}
\eta_{\vec{y}}(\vec{h},\vec{h}_s) = \eta_{\vec{y}}(a,b,c,d) =  \frac{s_{\vec{y}}^A + s_{\vec{y}}^C}{s_{\vec{y}}+ a+c} - \frac{s_{\vec{y}}^B + s_{\vec{y}}^C}{s_{\vec{y}}+ b+c} \,.
\end{eqnarray*}
Recall that the relaxation again originates from the fact that the solution not necessarily complies with the definitions of the sets $A$, $B$, $C$ and $D$.

Now observe that the objective function of the mixed-integer nonlinear program is strictly decreasing in $b$, independent of the other variables, so we can fix $b=0$. Subsequently, observe that for $b=0$ the objective function is also strictly decreasing in $c$, independent of the other variables, so we also fix $c=0$. As a result, the coefficients can be further simplified to:  
\begin{eqnarray}
\eta_{\vec{y}}(\vec{h},\vec{h}_s) = \eta_{\vec{y}}(a,b,c,d) = \left\{ \begin{array}{cl} \frac{ s_{\vec{y}}^A}{s_{\vec{y}}+ a}  & \quad
\textrm{if $\vec{y} \neq \vec{0}_m \,,$} \\
-1  & \quad \textrm{if $\vec{y} = \vec{0}_m \,.$}
\end{array} \right. 
\end{eqnarray}
To complete the proof we show by contradiction that the optimum is obtained for $a=m$. In order to construct the recurrence equations below let us introduce $q \in \{1,...,m-1\}$ and let
\begin{eqnarray*}
\Omega^0 &=& \{\vec{y} \in \{0,1\}^m \mid y_{q+1} = 0 \wedge \vec{y} \neq \vec{0}_m\} \,, \\
\Omega^1 &=& \{\vec{y} \in \{0,1\}^{m} \mid y_{q+1} = 1 \wedge \vec{y} \neq \vec{0}_{q+1}^1\} \,,
\end{eqnarray*}
where $\vec{0}_{q+1}^1$ denotes a vector of $m-1$ zeros apart from a single one at position $q+1$. Furthermore, let us introduce the mapping $\Psi: \{0,1\}^m \rightarrow \{0,1\}^m$, which, in any binary vector of length $m$, shifts the bit at position $q+1$: a zero at that position becomes a one and vice versa. The mapping $\Psi$ hence defines a unique correspondence between any element in $\Omega^0$ and its sister element in $\Omega^1$. For $a=q$, the objective function can then be written as:
\begin{eqnarray*}
\delta^q(\Pr) =  \sum_{\vec{y} \in \{0,1\}^m} \eta_{\vec{y}}(a,b,c,d) \Pr(\vec{y}) &=& - \Pr(\vec{0}_m) \\ && + \sum_{\vec{y} \in \Omega^0} \frac{ 2s_{\vec{y}}^A}{s_{\vec{y}}+ q} \Pr(\vec{y}) +
\sum_{\vec{y} \in \Omega^1} +\frac{2s_{\vec{\Psi(y)}}^A}{s_{\vec{\Psi(y)}}+ q + 1} \Pr(\vec{y})
\end{eqnarray*}
while for $a=q+1$, it can be written as:
\begin{eqnarray*}
\delta^{q+1}(\Pr) = \sum_{\vec{y} \in \{0,1\}^m} \eta_{\vec{y}}(a,b,c,d) \Pr(\vec{y}) &=& - \Pr(\vec{0}_m) + \frac{2}{a+1} \Pr(\vec{0}_{q+1}^1) \\ && + \sum_{\vec{y} \in \Omega^0} \frac{ 2s_{\vec{y}}^A}{s_{\vec{y}}+ q + 1} \Pr(\vec{y}) +
\sum_{\vec{y} \in \Omega^1} +\frac{2 s_{\vec{\Psi(y)}}^A + 2}{s_{\vec{\Psi(y)}}+ q + 2} \Pr(\vec{y})
\end{eqnarray*}
Let us assume that the global optimum is obtained for $a < m$. Furthermore, let $\Pr^q(\vec{y})$ be the probability distribution that delivers this optimum for $(a=q,d=m-q)$. We construct a new probability distribution $\Pr^{q+1}(\vec{y})$ as follows:
\begin{eqnarray*}
\Pr^{q+1}(\vec{y}) & = & \left\{ \begin{array}{cl} \Pr^q(\Psi(\vec{y})) & \quad
\textrm{if $(\vec{y} \in \Omega^1 \wedge \Pr^q(\vec{y}) > \Pr^q(\Psi(\vec{y}))) \vee (\vec{y} \in \Omega^0 \wedge \Pr^q(\vec{y}) < \Pr^q(\Psi(\vec{y}))) \,,$} \\
\Pr(\vec{y})  & \quad \textrm{otherwise $\,,$}
\end{array} \right. 
\end{eqnarray*}
If $\Pr^q(\vec{y})$ is feasible, then $\Pr^{q+1}(\vec{y})$ is feasible, too. It follows from 
$$ \frac{2 s_{ \vec{\Psi(y)}}^A + 2}{s_{\vec{\Psi(y)}}+ q + 2} \geq \frac{2 s_{\vec{\Psi(y)}}^A}{s_{\vec{\Psi(y)}}+ q} \,, \quad \forall \vec{y} \in \{0,1\}^m $$
that $\delta^{q+1}(\Pr^{q+1}) \geq \delta^q(\Pr^q)$ for all $q$. This is a contradiction. Consequently, the global optimum of the optimization problem is given by $\Pr^A$. 
\end{proof}

\noindent {\bf Theorem 3.3}
{\it Let $\vec{h}_J$ and $\vec{h}_F$ be vectors of predictions obtained by maximizing the Jaccard index and the F-measure, respectively. Let the utility of the F-measure maximizer be given by
\begin{equation*}
\delta(\Pr) = \max_{\vec{h} \in \lbrace 0,1 \rbrace^m} \mathbb{E} \left[F(\brY,\vec{h})\right]
          = \max_{\vec{h} \in \lbrace 0,1 \rbrace^m} \sum_{\vec{y} \in \lbrace 0,1 \rbrace^m}\Pr(\vec{y})\,F(\vec{y},\vec{h})  .
\end{equation*}
The regret of the F-measure maximizer with respect to the Jaccard index is then upper bounded by 
\begin{eqnarray*}
 \mathbb{E} \big[J(\brY, \vec{h}_J) -  J(\brY, \vec{h}_F) \big] \leq 1 - \delta(\Pr) / 2  \, \\
\end{eqnarray*}
for all possible distributions $\Pr$}

\begin{proof}
The proof follows immediately from the following double inequality:
\begin{eqnarray*}
\frac{\sum_{i=1}^m y_i h_i(\vec{x})}{\sum_{i=1}^m y_i + \sum_{i=1}^m h_i(\vec{x})}
\leq \\
\frac{\sum_{i=1}^m y_i h_i(\vec{x})}{\sum_{i=1}^m y_i + \sum_{i=1}^m h_i(\vec{x}) - \sum_{i=1}^m y_i h_i(\vec{x})} \leq \\
\frac{2 \sum_{i=1}^m y_i h_i(\vec{x})}{\sum_{i=1}^m y_i + \sum_{i=1}^m h_i(\vec{x})} \,,
\end{eqnarray*}
which results in 
$$\frac{F(\vec{y}, \vec{h}_F)}{2} \leq  J(\vec{y},\vec{h}_F) \leq F(\vec{y}, \vec{h}_F) \,,$$
for all $\vec{y}, \vec{h} \in \{0,1\}^m$. 
 \end{proof}
 
\noindent
{\bf Theorem 4.2}
{\it Let $\vec{h}_I$ be a vector of predictions obtained by assuming label independence as defined in (\ref{eq:label_independence}), then the worst-case regret is lower-bounded by:
\begin{eqnarray*}
\sup_{\Pr} \, (\mathbb{E} \big[ F(\brY, \vec{h}_F) -  F(\brY, \vec{h}_I) \big]) \geq 2q -1 , \\
\end{eqnarray*}
for all $q \in [1/2,1]$ satisfying
$\sum_{s=1}^m  \big( \frac{2m!}{(m-s)! (s-1)! (m + s)} q^{m-s} (1-q)^s \big)  - q^m > 0 $ and the supremum taken over all possible distributions $\Pr$.} \\

\begin{proof}
To analyze the potential regret of methods that assume independence, it is sufficient to compare the F-maximizers and their corresponding F-measures for a joint distribution defined on independent random variables and a second joint distribution having the same marginal distributions, but no independence. Below, we analyze two families of probability distributions that are parameterized by a single parameter $q$, which is defined as $q= \Pr(Y_i = 0)$ for all $i=1,...,m$. The first family resembles the case of independent random variables, for which the joint distribution is defined as the product of marginal probabilities:
\begin{eqnarray}
\Pr_A(\vec{y}) = q^{m-s} (1-q)^s \quad \textrm{where } s = \sum_{i=1}^m y_i \,.
\end{eqnarray}
The second family of distributions captures one particular case of a very strong stochastic dependence:
\begin{eqnarray*}
\Pr_B(\vec{y}) & = & \left\{ \begin{array}{cl} q & \quad
\textrm{if $\vec{y} = \vec{0}_m \,$}\\
1-q  & \quad \textrm{if $\vec{y} = \vec{1}_m \,$}\\
0  & \quad \textrm{otherwise $\,$}
\end{array} \right. \enspace ,
\end{eqnarray*}
If $q > 0.5$, then the F-measure maximizer of $\Pr_B$ is given by $\vec{0}_m$ and its corresponding F-measure is $q$. It is less straightforward to find the F-measure maximizer of $\Pr_A$. Let us first introduce the equivalence classes
\begin{eqnarray*}
\Omega_m(s) = \{ \vec{y} \in \{0,1\}^m \mid \sum_{i=1}^m y_i = s \} \,, \\
\Omega_m(s,l) = \{ \vec{y} \in \{0,1\}^m \mid \sum_{i=1}^m y_i = s \wedge y_l = 1\} \,,
\end{eqnarray*}
with $s,l \in \{1,...,m\}$. The cardinality of these equivalence classes is given by
\begin{eqnarray*}
|\Omega_m(s)| &=& \Big( \, \begin{matrix} m \\ s \end{matrix} \, \Big) = \frac{m!}{(m-s)! s!} \,, \quad \\
|\Omega_m(s,l)| &=& \Big( \, \begin{matrix} m-1 \\ s-1 \end{matrix} \, \Big) = \frac{(m-1)!}{(m-s)! (s-1)!} \,.
\end{eqnarray*}
Let $\vec{h}_k$ be a series of predictions such that $\sum_{i=1}^m h_i = k$. Without loss of generality, we can fix $\vec{h}_k$ to a vector of $k$ ones that are followed by $m-k$ zeros, because the distributions that we analyze are fully symmetric (i.e., all index permutations of $\{1,...,m\}$ in the label vectors yield the same values in probability mass). As a result, we can write the expected F-measure of $\vec{h}_k$ with $k>0$ as:
\begin{eqnarray}
\mathbb{E}_{\brY \sim \Pr_A} \left[ F(\brY, \vec{h}_k) \right] &=&  \sum_{\vec{y} \in \{0,1\}^m } F(\vec{y}, \vec{h}_k) \Pr_A(\vec{y}) \label{eq:expected}\\
&=& \sum_{\vec{y} \in \{0,1\}^m } \frac{\sum_{i=1}^k 2 y_i}{s_{\vec{y}}+k} \Pr_A(\vec{y}) \nonumber \\
&=& \sum_{s=1}^m \sum_{\vec{y} \in \Omega_m(s)} \frac{\sum_{i=1}^k 2 y_i}{s + k} \Pr_A(\vec{y}) \nonumber \\
&=& \sum_{s=1}^m \sum_{i=1}^k \sum_{\vec{y} \in \Omega_m(s)} \frac{2 y_i}{s + k} q^{m-s} (1-q)^s \nonumber \\
&=& \sum_{s=1}^m \sum_{i=1}^k \sum_{\vec{y} \in \Omega_m(s,i)} \frac{2 }{s + k} q^{m-s} (1-q)^s \nonumber \\
&=& \sum_{s=1}^m  \frac{(m-1)!}{(m-s)! (s-1)!} \frac{2k }{s + k} q^{m-s} (1-q)^s \nonumber
\end{eqnarray}
This is an increasing function of $k$, which implies that the F-measure maximizer consists of a vector of solely ones (or solely zeros) for $\Pr_A$. In addition, the expected F-measure for a prediction vector of zeros is given by
$$\mathbb{E}_{\brY \sim \Pr_A} \left[ F(\brY, \vec{0}_m) \right] = q^m \,.$$
Let us define $\delta_m$ as
\begin{eqnarray*}
\delta_m &=& \mathbb{E}_{\brY \sim \Pr_A} \left[ F(\brY, \vec{1}_m) - F(\brY, \vec{0}_m) \right] \\
&=& \sum_{s=1}^m  \Big( \frac{(m-1)!}{(m-s)! (s-1)!} \frac{2m }{m + s} q^{m-s} (1-q)^s \Big) - q^m \,.
\end{eqnarray*}
Then, assuming independence delivers the wrong maximizer for $\Pr_B$ as soon as $\delta_m>0$. 
\end{proof}

\noindent
{\bf Corollary 4.3}
{\it Let $\vec{h}_I$ be a vector of predictions obtained by assuming independence, then the worst-case regret converges to 1 in the limit of $m$, i.e.,
\begin{eqnarray*}
\lim_{m \rightarrow \infty} \sup_{\Pr} \, (\mathbb{E} \big[ F(\brY, \vec{h}_F) -  F(\brY, \vec{h}_I) \big]) = 1 , \\
\end{eqnarray*}
where the supremum is taken over all possible distributions $\Pr$.} \\

\begin{proof}
For increasing $m$, the condition is satisfied for $q$ close to one. The easiest way to observe this is computing the limit
\begin{eqnarray*}
\lim_{m \rightarrow \infty} q^m &=& 0 \,,
\end{eqnarray*}
which implies
\begin{eqnarray*}
\lim_{m \rightarrow \infty} \sum_{s=1}^m  \frac{m!}{(m-s)! s!}  q^{m-s} (1-q)^s &=& 1 \,.
\end{eqnarray*}   
From this last limit and $$\frac{2m }{m + s} \geq \frac{m}{s}$$ it follows that:
$$\lim_{m \rightarrow \infty} \sum_{s=1}^m  \frac{(m-1)!}{(m-s)! (s-1)!} \frac{2m }{m + s} q^{m-s} (1-q)^s \geq 1 \,.$$
By definition, (\ref{eq:expected}) cannot exceed the upper bound of one, so this inequality must hold as an equality.  
In such a scenario, the worst-case regret is lower bounded by $R_q = 2 q - 1$, so that $\lim_{q \rightarrow 1, m \rightarrow \infty} R_q = 1$.
As a consequence, the lower bound becomes tight in the limit of $m$ going to infinity. \\
\end{proof}

\noindent {\bf Theorem 4.5}
{\it Let $\vec{h}_T$ be a vector of predictions obtained by putting a threshold on sorted marginal probabilities, then the worst-case regret is lower bounded by
\begin{eqnarray*}
\sup_{\Pr} \, (\mathbb{E} \big[ F(\brY, \vec{h}_F) -  F(\brY, \vec{h}_T) \big]) \geq \max \left(0,\frac{1}{6} - \frac{2}{m+4} \right) , 
\end{eqnarray*}
where the supremum is taken over all possible distributions $\Pr$.}

\begin{proof}
To analyze the regret of thresholding approaches, we have to construct a counterexample for which the F-measure is not consistent with the order of the marginal probabilities. The following family of distributions is such a counterexample:
\begin{eqnarray}
\label{eq:prob_dist}
\Pr(\vec{y}) & = & \left\{ \begin{array}{cl} 1/2 - \epsilon & \quad
\textrm{if $y_1 = 1 \wedge \sum_{i=1}^m y_i = 1 \,$}\\
(1/2+\epsilon)/(2m-4) & \quad \textrm{if $y_2 = 1 \wedge \sum_{i=3}^{m/2 +1} y_i = \sum_{i=3}^{m} y_i = m/2 \,$}\\
(1/2+\epsilon)/(2m-4)  & \quad \textrm{if $y_2 = 1 \wedge \sum_{i=m/2+2}^{m} y_i = \sum_{i=3}^{m} y_i = m/2 \,$}\\
0  & \quad \textrm{otherwise $\,$}
\end{array} \right. \enspace
\end{eqnarray}
where we consider for simplicity that $m$ is even and $\epsilon \in [0,1/2]$ represents a positive constant. For $\epsilon$ close to zero, one can easily show that the F-measure maximizer is given by a vector of predictions consisting of only zeros, apart from a single one at position one. The F-measure of this prediction vector is $1/2 - \epsilon$. However, this prediction vector can never be returned by a method that relies on thresholding over marginal probabilities, because $\Pr(Y_2 = 1) > \Pr(Y_1 = 1)$ in this particular case. By enumerating all candidate solutions examined by thresholding, one will find instead a prediction vector $\vec{h}_{11}$ consisting of zeros, apart from a one at the first two positions. The expected F-measure of this prediction vector is
$$\mathbb{E} \left[ F(\brY, \vec{h}_{11}) \right] = (1/2 - \epsilon) (2/3) + (1/2 + \epsilon) (2/(2+(m/2)))\,.$$
As a consequence, this results in the above-mentioned regret when $\epsilon$ approaches zero.
\end{proof}
 
\bibliography{references}

\end{document}